\documentclass[10pt,twocolumn,letterpaper]{article}

\usepackage[footnotesize]{subfigure}
\usepackage[pagenumbers]{cvpr} 

\usepackage{booktabs}
\usepackage{graphicx}
\usepackage{amsmath}
\usepackage{amssymb}
\usepackage{booktabs}
\usepackage{multirow}
\usepackage[ruled]{algorithm2e}
\usepackage[accsupp]{axessibility}  

\usepackage[pagebackref,breaklinks,colorlinks]{hyperref}

\usepackage[capitalize]{cleveref}
\crefname{section}{Sec.}{Secs.}
\Crefname{section}{Section}{Sections}
\Crefname{table}{Table}{Tables}
\crefname{table}{Tab.}{Tabs.}

\def\cvprPaperID{3727} 
\def\confName{CVPR}
\def\confYear{2023}

\begin{document}

\title{Multi-View Stereo Representation Revisit: Region-Aware MVSNet}

\author{Yisu Zhang$^1$, \ \ \ Jianke Zhu$^{1,2}\thanks{Corresponding author is Jianke Zhu.}$\ \ \ Lixiang Lin$^1$, \\
 $^1$Zhejiang University \\
 $^2$Alibaba-Zhejiang University Joint Research Institute of Frontier Technologies \\ 
 {\tt\small \{zhyisu, jkzhu, lxlin\}@zju.edu.cn}
}
\maketitle

\begin{abstract}

Deep learning-based multi-view stereo has emerged as a powerful paradigm for reconstructing the complete geometrically-detailed objects from multi-views. Most of the existing approaches only estimate the pixel-wise depth value by minimizing the gap between the predicted point and the intersection of ray and surface, which usually ignore the surface topology. It is essential to the textureless regions and surface boundary that cannot be properly reconstructed.
To address this issue, we suggest to take advantage of point-to-surface distance so that the model is able to perceive a wider range of surfaces. To this end, we predict the distance volume from cost volume to estimate the signed distance of points around the surface. Our proposed RA-MVSNet is patch-awared, since the perception range is enhanced by associating hypothetical planes with a patch of surface. Therefore, it could increase the completion of textureless regions and reduce the outliers at the boundary.
Moreover, the mesh topologies with fine details can be generated by the introduced distance volume. Comparing to the conventional deep learning-based multi-view stereo methods, our proposed RA-MVSNet approach obtains more complete reconstruction results by taking advantage of signed distance supervision. The experiments on both the DTU and Tanks \& Temples datasets demonstrate that our proposed approach achieves the state-of-the-art results.

\end{abstract}

\section{Introduction}
\label{sec:intro}

Multi-view stereo (MVS) is able to efficiently recover geometry from multiple images, which makes use of the matching relationship and stereo correspondences of overlapping images. 

To achieve the promising reconstruction results, the conventional patch-based and PatchMatch-based methods~\cite{barnes2009patchmatch, furukawa2009accurate, schonberger2016pixelwise} require rich textures and restricted lighting conditions. Alternatively, the deep learning-based approaches~\cite{chen2019point, yao2018mvsnet, huang2018deepmvs, ji2017surfacenet} try to take advantage of global scene semantic information, including environmental illumination and object materials, to maintain high performance in complex lighting. The key of these methods is to warp deep image features into the reference camera frustum so that the 3D cost volume can be built via differentiable homographies. Then, the depth map is predicted by regularizing cost volume with 3D CNNs.

\begin{figure}
  \begin{center}
     \includegraphics[trim={0cm 1.5cm 2cm 0cm},clip,width=1.0\linewidth]{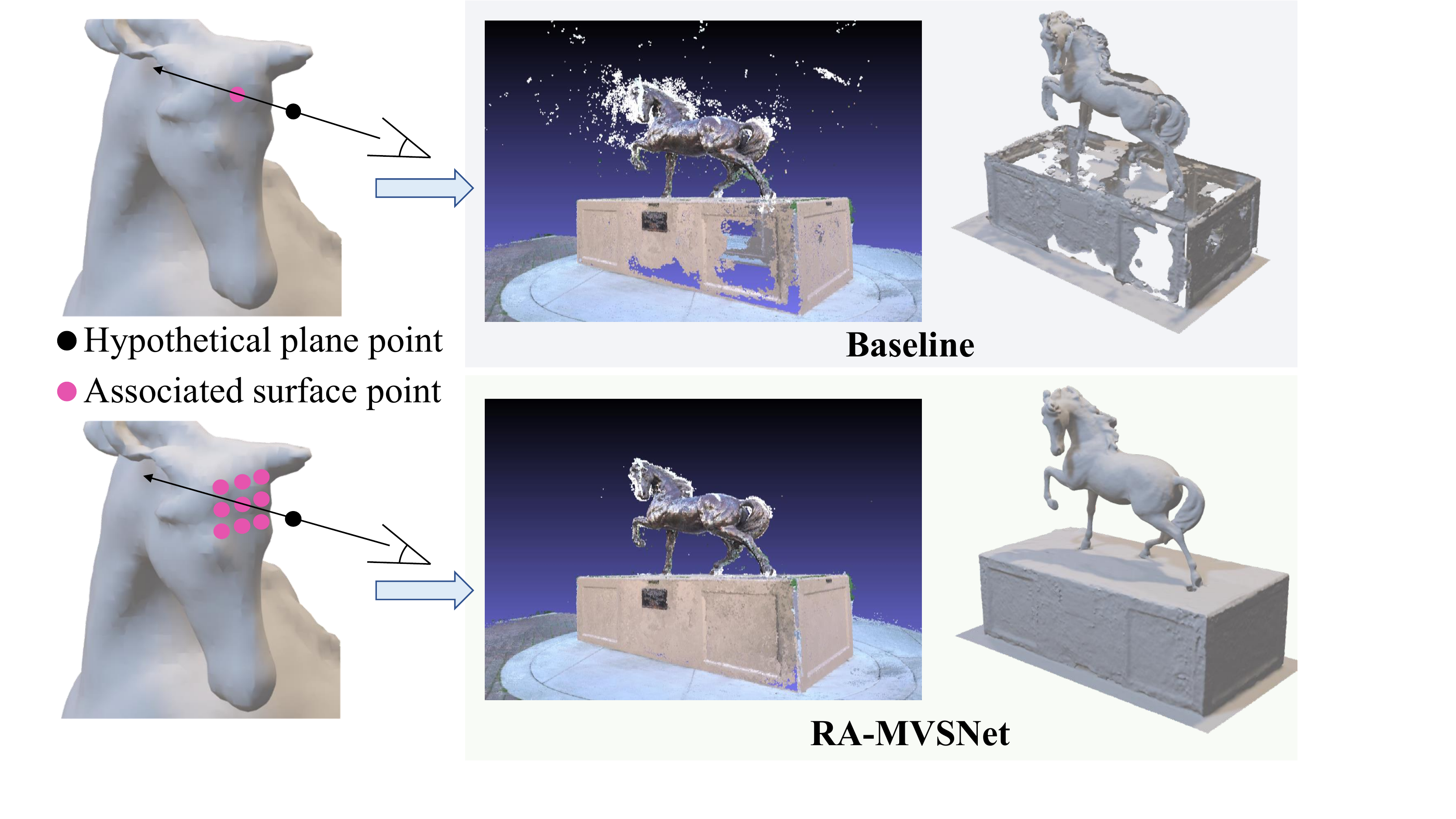}
  \end{center}
  \vspace{-0.4cm}
  \caption{{\bf Comparison on reconstruction results between baseline and RA-MVSNet.} Our RA-MVSNet enables the model to perceive a wider range of surfaces so as to achieve the promising performance in complementing textureless regions and removing outliers at boundaries. Furthermore, our model is able to generate correct mesh topologies with fine details.} 
  \label{fig:structure}
  \vspace{-0.4cm}
\end{figure}

Despite the encouraging results, the pixel-wise depth estimation suffers from two intractable flaws. One is the low estimation confidence in the textureless area. The other is many outliers near the boundary of the object. This is mainly because the surface is usually treated as a set of uncorrelated sample points rather than the one with topology. 
As each ray is only associated with a single surface sampling point, it is impossible to pay attention to the adjacent area of the surface. As shown in~\cref{fig:structure}, the estimation of each depth value is constrained by only one surface sample point, which makes it unable to use the surrounding surface for inference. Unfortunately, it is difficult to infer without broader surface information in textureless regions and object boundaries. Therefore, too small perception range limits the existing learning-based MVS methods.

To tackle this issue, we present a novel RA-MVSNet framework that is able to make each hypothetical plane associated with a wider surface area through point-to-surface distance. Thus, our presented method is capable of inferring the surrounding surface information at textureless areas and object boundaries. To this end, our network not only estimates the probability volume but also predicts the point-to-surface distance of each hypothetical plane. Specifically, RA-MVSNet makes use of the cost volume to generate the probability and distance volumes, which are further combined to estimate the final depth map. The introduction of point-to-surface distance supervision uses the model patch-aware in estimating the depth value corresponding to a particular pixel. This leads to the improved performance in textureless or boundary areas. Since the distance volume estimates the length of the sample points near the surface, we are able to predict a SDF-based implicit representation with the correct topology and fine details.

In summary, our contribution is three-fold:
{\begin{itemize}
    \item We introduce point-to-surface distance supervision of sampled points to expand the perception range predicted by the model, which achieves complete estimation in textureless areas and reduce outliers in object boundary regions.
    \item To tackle the challenge of lacking the ground-truth mesh, we compute the signed distance between point sets based on the triangulated mesh, which trades off between accuracy and speed.
    \item Experimental results on the challenging MVS datasets show that our proposed approach performs the best both on indoor dataset \emph{DTU}~\cite{aanaes2016large} and large-scale outdoor dataset \emph{Tanks and Temples}~\cite{knapitsch2017tanks}. 
\end{itemize}}

\begin{figure*}
  \begin{center}
     \includegraphics[trim={0.1cm 0.1cm 0.1cm 0cm},clip,width=1.0\linewidth]{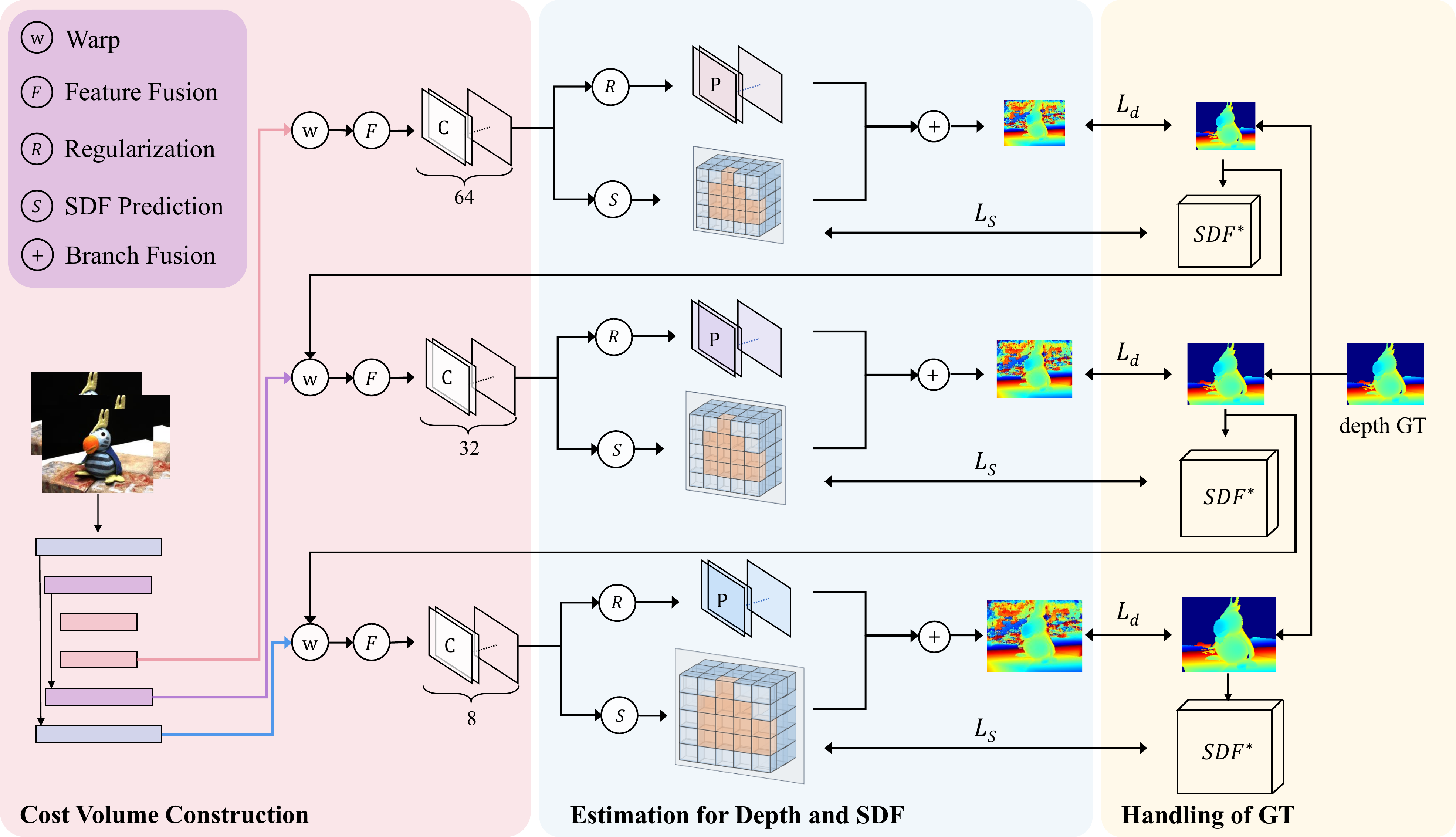}
  \end{center}
  \vspace{-0.4cm}
  \caption{{\bf Illustration of RA-MVSNet.} Our RA-MVSNet framework consists of two branches. The first branch predicts probability volume, and the second one estimates the signed distance volume. Fusing two branches can get the filtered depth maps while SDF branch can generate implicit representation.
  } 
  \label{fig:framework}
  \vspace{-0.4cm}
\end{figure*}

\vspace{-0.13cm}
\section{Related Work}
We review the multi-view stereo studies from two aspects, including conventional methods and learning-based approaches.

\textbf{Conventional Multi-View Stereo.} The conventional MVS methods make use of various 3D representations, such as mesh~\cite{fua1995object}, point cloud~\cite{furukawa2009accurate, lhuillier2005quasi}, voxel~\cite{kutulakos2000theory, seitz1999photorealistic} and depth map~\cite{campbell2008using, galliani2015massively, schonberger2016pixelwise}. Among these different representations, the depth map-based methods can obtain more complete surface reconstruction with higher robustness. They avoided solving the intractable topology problem by formulating the multi-view reconstruction into a depth estimation problem, which fuses all depth map into single 3D point cloud. Among them, COLMAP~\cite{schonberger2016pixelwise} and ACMM~\cite{xu2019multi} can obtain the stable results. Specifically, ACMM employs multi-scale geometric consistency to reconstruct features at different scales. COLMAP estimates the pixelwise depth and normal using photometric and geometric priors. In the cases of complicated scenario, large matching noise and poor correspondences, the results of traditional MVS may have the obvious artifacts.

\textbf{Learning-based Multi-View Stereo.} To address the limitations of traditional methods, deep learning-based approaches are proposed to robustly estimate depth map. MVSNet~\cite{yao2018mvsnet} firstly builds 3D cost volume to aggregate the warped features from the reference and source images, and then regresses the depth map by a 3D CNN. 
Later, some works~\cite{zhang2020visibility, ChenVAPMVSNet2020TPAMI, ding2022transmvsnet} take into consideration of the attention mechanism to focus on the areas to be reconstructed.
~\cite{ Wang_2022_CVPR, mi2022generalized} replace the cost volume representation to reduce the unnecessary computation. Ding et al. ~\cite{ding2022enhancing} try to optimize the depth map representation. Meanwhile, some studies intend to learn how to regularize cost volume better by hybrid 3D U-Net~\cite{luo2019p, sormann2020bp} and epipolar attention~\cite{ma2021epp, yang2022mvs2d}. 
Although the vanilla MVSNet is able to obtain the pixel-wise depth prediction, dense hypothetical planes and 3D cost volume consume a large amount of memory. Recurrent MVSNet architectures~\cite{yao2019recurrent, Xi_2022_CVPR, nonlocal}, coarse-to-fine manner~\cite{gu2020cascade} and multi-stage binary search~\cite{mi2022generalized} are proposed to further excavate the potential capacity of this pipeline. 
Similar ideas are later explored to reduce the memory consumption of 3D convolutions and increase the depth quality, such as coarse-to-fine depth optimization ~\cite{cheng2020deep, xu2020learning, yan2020dense, yu2020fast}, attention-based feature aggregation ~\cite{luo2020attention, wei2021aa, ding2022transmvsnet, yu2021attention, zhang2021long}, and patch matching-based method ~\cite{ wang2021patchmatchnet}. 
Meanwhile, Uni-MVSNet~\cite{unimvsnet} analyzes the impact of regression and classification operations in the pipeline, and combines these two approaches to achieve more accurate predictions. 

In general, the traditional methods based on patch matching cannot cope with the complicated lighting conditions and textureless areas while the deep learning methods based on cost volume predict many outliers in the object's boundary regions. In this paper, our proposed RA-MVSNet makes learning-based MVS patch-aware to increase the prediction performance for textureless regions and reduce erroneous outlier points near the object's boundary.

\section{Method}
In this section, we introduce the detailed structure of the proposed RA-MVSNet. 
As shown in~\Cref{fig:framework}, the overall framework mainly consists of three parts, including cost volume construction, the multi-scale depth map and signed distance prediction, and handling of ground truth.
Since our proposed point-to-surface distance supervision employs an additional branch to calculate the signed distance of the sampling points around the surface through cost volume, it is easy to be added into the existing learning-based MVSNet scheme with slight changes. To investigate the effectiveness of our method, we mainly employ the cascade MVSNet as the baseline and use two branches on the basis of Cas-MVSNet~\cite{gu2020cascade} to predict the depth and signed distance, respectively.

\subsection{Cost Volume Construction} \label{3.1}

The construction of cost volume mainly relies on MVSNet framework~\cite{yao2018mvsnet}, which utilizes the warped frustum features to predict the depth map $\{\mathbf{D} \in \mathbb{R}^{H^{\prime} \times W^{\prime}}\}$ corresponding to the reference image $\{\mathbf{I_0} \in \mathbb{R}^{H^{\prime} \times W^{\prime}}\}$. The feature volume is aggregated by warping source image features into the reference view, where all image features are extracted by 2D FPN-based network with the shared weights. 
To pay more attention to the object to be reconstructed, we employ Recursive Feature Pyramid~(RFP) structure~\cite{Qiao_2021_CVPR} as image encoder in order to obtain a pyramid of feature maps $\{\mathbf{F}_{i} \in \mathbb{R}^{{C^{\prime}}\times{H^{\prime}}\times{W^{\prime}}} \}_{i=1}^{N}$ with three different scales.

By regularizing 3D cost volume in the whole known depth range, the estimated depth map consists of the depth hypothesis of $M$ layers. The key to learning-based MVS lies in the way of dealing with 3D cost volume. Specifically, the feature volume can be aggregated by differentiable homography as below
\vspace{-0.1cm}
\begin{equation}
\mathbf{H}_{i}(d)=d \mathbf{K}_{i} \mathbf{R}_{i} \mathbf{R}_{0}^{-1} \mathbf{K}_{0}^{-1}, 
\end{equation}
where ${d}$ refers to the hypothetical depth of source image $\mathbf{R}_{i}$. $\mathbf{K}_{i} \mathbf{R}_{i}$ and $\mathbf{K}_{0} \mathbf{T}_{0}$ represent camera intrinsic and extrinsic parameters of source and reference images, respectively. Therefore, the warped pixel ${p^\prime}$ in source image $\mathbf{I}_{i}$ of reference pixel ${p}$ can be computed by
\begin{equation}
\mathbf{p}^{\prime}=\mathbf{H}_{i}(d) \cdot p +  \mathbf{t}_{\mathbf{I}_{0} \rightarrow \mathbf{I}_{i}},
\end{equation}
where $\mathbf{t}_{\mathbf{I}_{0} \rightarrow \mathbf{I}_{i}}$ is the relative camera translation from reference to the source image. To handle the arbitrary number of source images, we aggregate all feature volumes $\{V_i \in \mathbb{R}^{D \times C^{\prime} \times H^{\prime} \times W^{\prime}}\}_{i=0}^{N-1}$ into single cost volume $\{\mathbf{C} \in \mathbb{R}^{D \times C^{\prime} \times H^{\prime} \times W^{\prime}}\}$ using an adaptive strategy so that several 3D CNN layers can be employed to predict the pixel-wise weighting matrices $\{W_i\}_{i=1}^{N-1}$. Thus, the final cost volume can be computed as follows
\vspace{-0.1cm}
\begin{equation}
\mathbf{C} = \sum_{i=1}^{N-1} \frac{1}{N-1} \mathcal{W}_{i} \odot\left(\mathbf{V}_{i}-\mathbf{V}_{0}\right)^{2},
\end{equation}
where ${\mathbf{C}}$ is the cost volume of reference view. $\odot$ denotes the element-wise multiplication. ${\mathbf{V}_i}$ and ${\mathbf{V}_0}$ are the features extracted from source images and reference view using image encoder.

\subsection{Signed Distance Supervision} \label{3.2}

The point-to-surface distance is usually represented as SDF (signed distance field) in a recent study~\cite{Park_2019_CVPR}. The core of this implicit representation is to calculate the distance from the sampled point near the surface to the object. Therefore, we follow the idea of SDF to construct a distance volume to predict the point-to-surface distances so as to take advantage of implicit representation.

\begin{figure}[htb]
    \centering
    \includegraphics[trim={0cm 0cm 0cm 0cm},clip,width=0.45\textwidth]{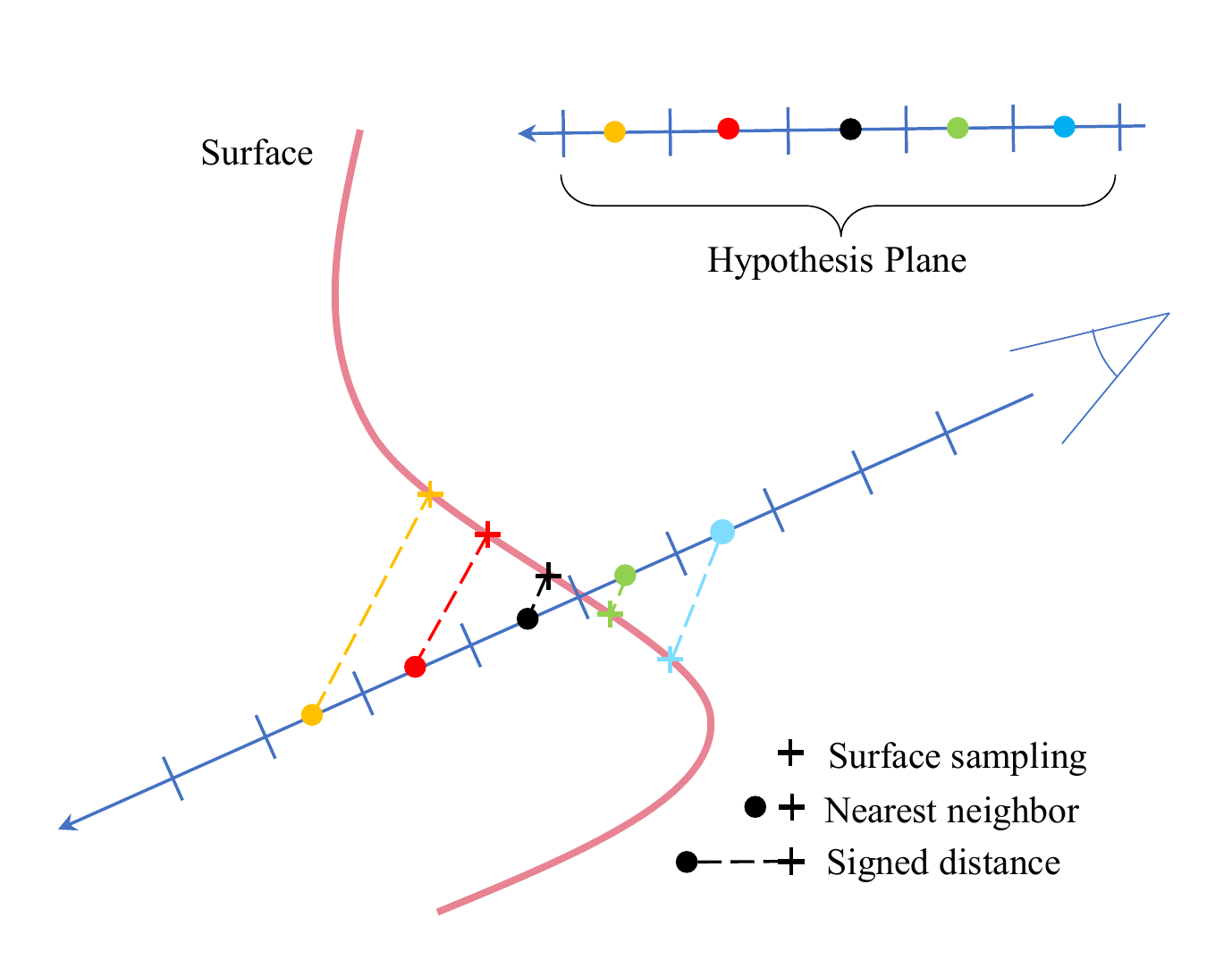}
    
    \caption{{\bf Ground truth of signed distance.} The ground truth of signed distance is represented by the sided distance between the two point sets. We treat each hypothetical plane as a sampled point around the surface and find its corresponding nearest-neighbor surface sampled point to get the ground truth signed distance.}
    \label{fig1}
    \vspace{-0.2cm}
\end{figure}

Given the 3D cost volume aggregating the feature of reference view and source views, the regularization networks are typically employed to obtain the probability volume $\mathbf{P}$ that is treated as the weight of hypothetical planes at different depths
\begin{equation}
\mathbf{P} = F_{softmax}(\mathbf{C}),
\end{equation}
where $F_{SoftMax}$ is softmax-based 3D CNN regularization networks. Distance volume $\mathbf{S}$ represents the signed distance of these hypothetical planes
\vspace{-0.1cm}
\begin{equation}
\mathbf{S} = F_{tanh}(\mathbf{C}),
\end{equation}
where $F_{tanh}$ denotes the tanh-based 3D CNN regularization networks. As the points far away from the surface are usually unhelpful for reconstruction, we employ $tanh$ as the activation layer for distance volume. Thus, we can focus on nearby sampled points.

Since the predictions of distance are introduced, we need to extend the ground truth from depth map to the signed distance field. Therefore, the depth map only contains the sampled points with the signed distance of 0, which lacks the ground truth of points around the surface.

For an exact query point ${\mathbf{p}_{i}}$ from each hypothetical plane of cost volume ${\{\mathbf{C}\}}$, we compute the shortest distance from ${\mathbf{p}_{i}}$ to the surface sampling points ${\mathbf{p}^\prime}$ as the ground truth of the signed distance. As shown in \cref{fig1}, we employ the two-point distance ${d(\mathbf{p}_{i}, \mathbf{p}_{j}^\prime)}$ that is calculated by Kaolin~\cite{KaolinLibrary} as ground truth signed distance. 

\begin{figure}[htb]
  \begin{center}
     \includegraphics[trim={2cm 3cm 5cm 1.5cm},clip,width=1.0\linewidth]{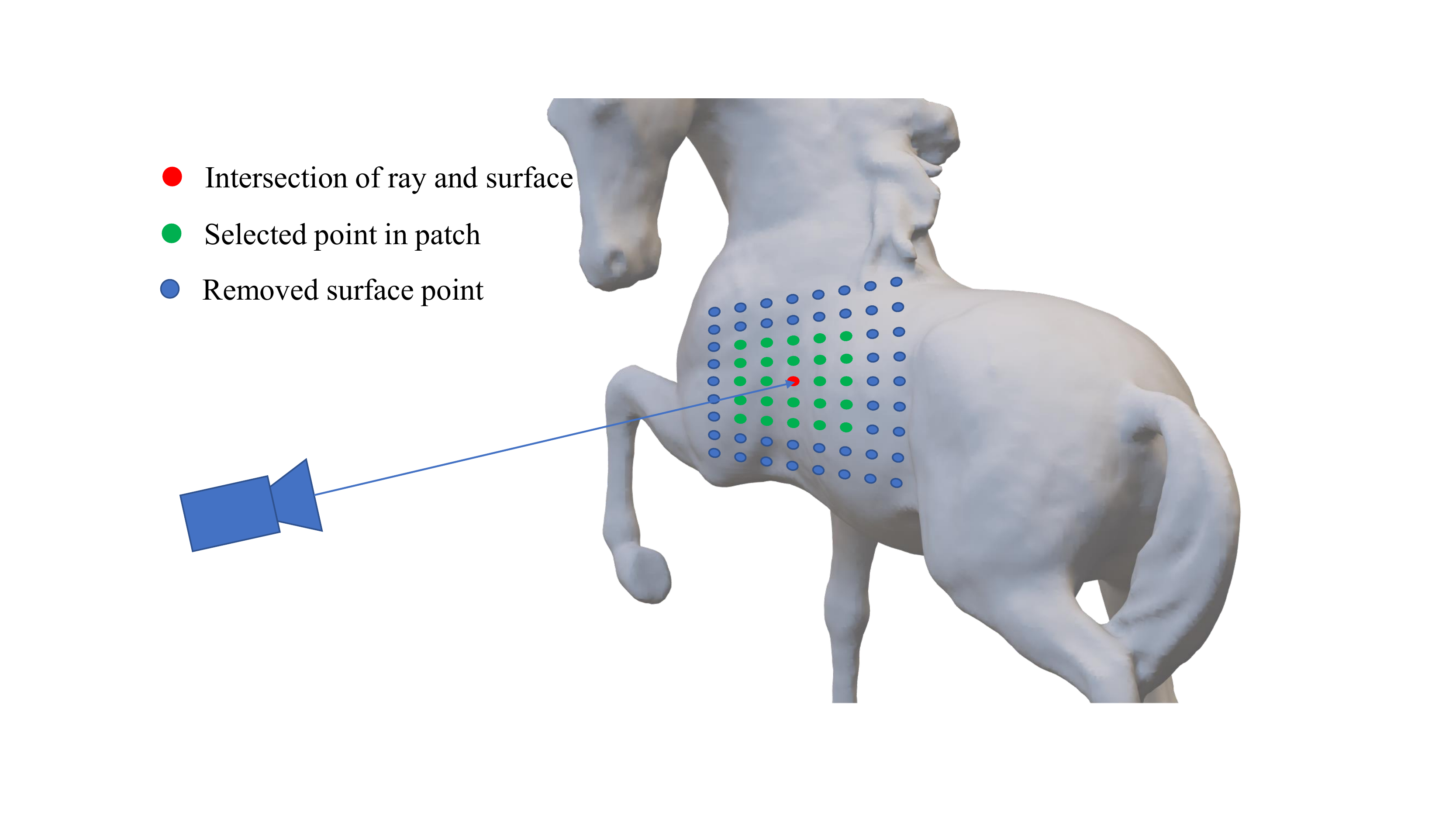}
  \end{center}
  \vspace{-0.4cm}
  \caption{{\bf Patch-based nearest neighbor search.} The nearest neighbor is usually located near the query point so that a large number of useless surface sampled points are removed while only the sampled points within the local patch located at the intersection are retained.}
  \label{fig:patch_compute}
  \vspace{-0.2cm}
\end{figure}

To speed up the process, finding the nearest neighbor from all surface sampling points is replaced by local search, as shown in \cref{fig:patch_compute}.

This patch-based local search method keeps the points that need to be calculated as few as possible within a reasonable range, thereby reducing the time complexity on search. We assume that the resolution of the depth map is $H \times W$ and the number of query points is $n$. Then, the time complexity of naive calculation is $O(n \times H \times W)$, which is proportional to the resolution of the depth map. In contrast, the time complexity of patch-based local search is simplified to $O(n \times k \times k)$, where $k$ is the patch size and usually set to 5. Therefore, the time complexity of patch-based local search can be simplified to $O(n)$. That is to say, it is only proportional to the number of query points $n$, and the search time for each query point is constant.

\subsection{Volume Fusion} \label{3.3}

Once the probability volume $\{\mathbf{P} \in \mathbb{R}^{D \times H^{\prime} \times W^{\prime}}\}$ and distance volume $\{\mathbf{S} \in \mathbb{R}^{D \times H^{\prime} \times W^{\prime}}\}$ are obtained, we fuse these two volumes to get the final depth map $D \in \mathbb{R}^{ H^{\prime} \times W^{\prime}}$. In general, a softmax-based regularization network is typically employed to predict the depth map from $\mathbf{P}$ that is treated as the weight of hypothetical planes at different depths. Therefore, the depth map can be calculated as follows
\begin{equation}
{D}^{U, V}=\sum_{i=\mathbf{d}_{min}^{U, V}}^{\mathbf{d}_{max}^{U, V}} i \mathbf{P}(i)^{U, V},
\end{equation}
where $\mathbf{d}_{min}$ and $\mathbf{d}_{max}$ refer to the distance of the nearest and farthest hypothetical plane, respectively. However, this method has the accuracy problems due to involving with multiple invalid planes in the calculation. The depth value of a pixel $(U, V)$ is only related to several hypothetical planes corresponding to this pixel, which cannot be associated with other sampled points on surface. 

\begin{figure}[htb]
  \begin{center}
     \includegraphics[trim={0cm 6.2cm 1cm 0cm},clip,width=0.95\linewidth]{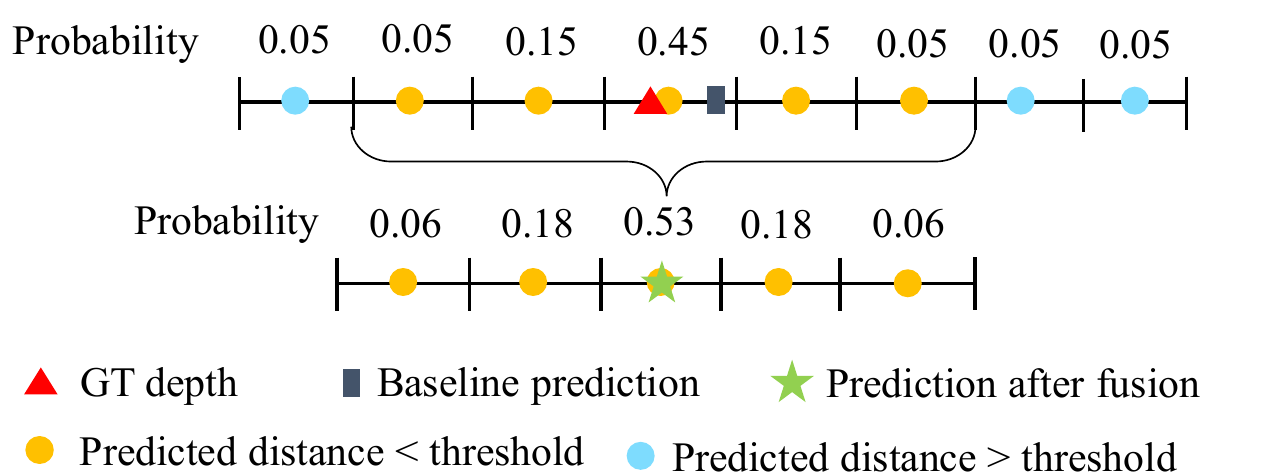}
  \end{center}
  \vspace{-0.4cm}
  \caption{{\bf Illustration of volume fusion.} RA-MVSNet can remove the invalid hypothesis planes through distance volume, which makes the results more accurate.}
  \label{fig:fusion}
  \vspace{-0.2cm}
\end{figure}

\SetKwInOut{KwIni}{Initialization}
\SetKw{KwAnd}{and}
\begin{algorithm}[t]
  \caption{Branches Fusion}
  \label{ag:fusion}
  \LinesNumbered
  \KwIn{Probability volume $\mathbf{P} \in \mathbb{R}^{D' \times H' \times W'}$; 
  Distance volume $\mathbf{S} \in \mathbb{R}^{D' \times H' \times W'}$.}
  \KwOut{Depth map $\mathbf{D} \in \mathbb{R}^{H' \times W'}$.}
  \KwIni{Depth map $\mathbf{D}=0$.}

    \For{$(u,v)=(1,1)$ \KwTo $(H',W')$}{
      \For{$i = \mathbf{d}_{min}^{u,v}$ \KwTo $\mathbf{d}_{max}^{u,v}$}{
        \eIf{$\mathbf{S}_i^{u,v} \le threshold$}{
          $\mathbf{D}_i^{u,v} = \mathbf{D}_i^{u,v} + Softmax(\mathbf{P}_i^{u,v}) \times \mathbf{d}_{i}^{u,v}$ 
        }{
          $\mathbf{D}_i^{u,v} = \mathbf{D}_i^{u,v}$ \;
        }
      }
    }
  \KwRet{$\mathbf{D}$}.
  \vspace{-0.1cm}
\end{algorithm} 

As shown in \cref{fig:fusion}, we fuse probability volume $\mathbf{P}$ and the introduced distance volume $\mathbf{S}$ to calculate the depth map so that each pixel is related to the surrounding surface patch. Specifically, $\mathbf{S}$ can be regarded as a filter of probability values by a threshold. The fusion process of these two volumes is illustrated in \cref{ag:fusion}. 
Finally, we use depth map ground truth and generated signed distance ground truth for supervision of two volumes $P$ and $S$. We employ $L_1$ loss for depth map and signed distance as follows
\vspace{-0.1cm}
\begin{equation}
{ L_d }=\sum_{i=1}^3 \|{{D_i^*}} -{{D_i}}\|,
\end{equation}
\begin{equation}
{ L_S }=\sum_{i=1}^3  \|{{S_i^*}} -{{S_i}}\|,
\end{equation}
where ${D_i^*}$ and ${S_i^*}$ are ground truth depth map and point-to-surface distance at stage $i$, respectively. ${D_i}$ and ${S_i}$ are the predicted value for two branches. Therefore, the total loss $L$ of our model is the weighted sum of two branches:
\begin{equation}
L = L_d + \lambda \cdot L_S,
\end{equation}
$\lambda$ is a weight to balance two terms, which is set to 0.1 in all experiments.

\begin{table*}[htb]
  \begin{center}
  
  \resizebox{1.0\linewidth}{!}{
  \begin{tabular}{l|ccccccccc|ccccccc}
  \toprule
  \multirow{2}*{Method} & \multicolumn{9}{c|}{Intermediate} & \multicolumn{7}{c}{Advanced} \\
  \cline{2-17}
  & Mean & Fam. & Fra. & Hor. & Lig. & M60 & Pan. & Pla. & Tra. & Mean & Aud. & Bal. & Cou. & Mus. & Pal. & Tem. \\
  \hline
  MVSNet~\cite{yao2018mvsnet} & 43.48 & 55.99 & 28.55 & 25.07 & 50.79 & 53.96 & 50.86 & 47.90 & 34.69 \\
  Point-MVSNet & 48.27 & 61.79 & 41.15 & 34.20 & 50.79 & 51.97 & 50.85 & 52.38 & 43.06 & - & - & - & - & - & - & - \\
  CVP-MVSNet~\cite{yang2020cost} & 54.03 & 76.50 & 47.74 & 36.34 & 55.12 & 57.28 & 54.28 & 57.43 & 47.54 & - & - & - & - & - & - & - \\
  P-MVSNet~\cite{ChenVAPMVSNet2020TPAMI} & 55.62 & 70.04 & 44.64 & 40.22 & {\bf 65.20} & 55.08 & 55.17 & 60.37 & 54.29 & - & - & - & - & - & - & - \\
  $D^2$HC-RMVSNet~\cite{yan2020dense} & 59.20 & 74.69 & 56.04 & 49.42 & 60.08 & 59.81 & 59.61 & 60.04 & 53.92 & - & - & - & - & - & - & - \\
  RayMVSNet~\cite{Xi_2022_CVPR} & 59.48 & 78.55 & 61.93 & 45.48 & 57.59 & 61.00 & 59.78 & 59.19 & 52.32 & - & - & - & - & - & - & - \\
  \hline
  PatchmatchNet~\cite{wang2021patchmatchnet} & 53.15 & 66.99 & 52.64 & 43.24 & 54.87 & 52.87 & 49.54 & 54.21 & 50.81 & 32.31 & 23.69 & 37.73 & 30.04 & 41.80 & 28.31 & 32.29 \\
  CasMVSNet~\cite{gu2020cascade} & 56.84 & 76.37 & 58.45 & 46.26 & 55.81 & 56.11 & 54.06 & 58.18 & 49.51 & 31.12 & 19.81 & 38.46 & 29.10 & 43.87 & 27.36 & 28.11 \\
  AA-RMVSNet~\cite{wei2021aa} & 61.51 & 77.77 & 59.53 & 51.53 & 64.02 & 64.05 & 59.47 & 60.85 & 55.50 & 33.53 & 20.96 & 40.15 & 32.05 & 46.01 & 29.28 & 32.71 \\
  GBi-Net~\cite{mi2022generalized} & 61.42 & 79.77 & {\bf 67.69} & 51.81 & 61.25 & 60.37 & 55.87 & 60.67 & 53.89 & 37.32 & {\bf 29.77} & 42.41 & 36.30 & 47.69 & 31.11 & 36.93 \\
  EPP-MVSNet~\cite{ma2021epp} & 61.68 & 77.86 & 60.54 & 52.96 & 62.33 & 61.69 & 60.34 & 62.44 & 55.30 & 35.72 & 21.28 & 39.74 & 35.34 & 49.21 & 30.00 & {\bf 38.75} \\
  TransMVSNet~\cite{ding2022transmvsnet} & 63.52 & 80.92 & 65.83 & 56.94 & 62.54 & 63.06 & 60.00 & 60.20 & {\bf 58.67} & 37.00 & 24.84 & 44.59 & 34.77 & 46.49 & {\bf 34.69} & 36.62 \\
  Uni-MVSNet~\cite{unimvsnet} & 64.36 & 81.20 & 66.43 & 53.11 & 63.46 & 66.09 & 64.84 & 62.23 & 57.53 & 38.96 & 28.33 & 44.36 & 39.74 & 52.89 & 33.80 & 34.63 \\

  \hline
  \hline
  RA-MVSNet (ours) & {\bf 65.72} & {\bf 82.44} & 66.61 & {\bf 58.40} & 64.78 & {\bf 67.14} & {\bf 65.60} & {\bf 62.74} & 58.08 & {\bf 39.93} & 29.17 & {\bf 46.05} & {\bf 40.23} & {\bf 53.22} & 34.62 & 36.30 \\
  \bottomrule
  \end{tabular}
  }
  \end{center}
  \vspace{-0.4cm}
  \caption{{\bf Quantitative results of F-score on Tanks and Temples benchmark.} 
  The best results in each category are in {\bf bold}. ``Mean'' refers to the mean F-score of all scenes (higher is better). 
  Our RA-MVSNet achieves competitive results on both intermediate and advanced set.}
  \label{tb:tank_compare}
  \vspace{-0.4cm}
\end{table*}

\subsection{Supervision of SDF Branch} \label{3.4}
\begin{figure}[htb]
	\centering
	\subfigure[The tangent point of the ball coincides with the surface sampled point]{
		\begin{minipage}[t]{0.22\textwidth} 
		\centering
        \includegraphics[width=\textwidth]{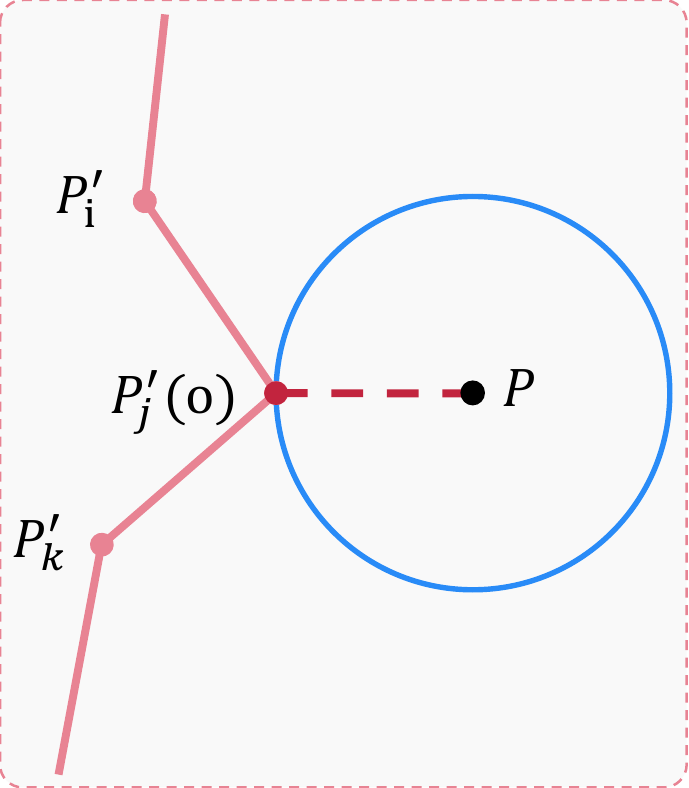} \\
		\end{minipage}
	}
	\subfigure[The tangent point of the ball falls on one of the sides of the triangle]{
	\begin{minipage}[t]{0.22\textwidth}
	\centering
	\includegraphics[width=\textwidth]{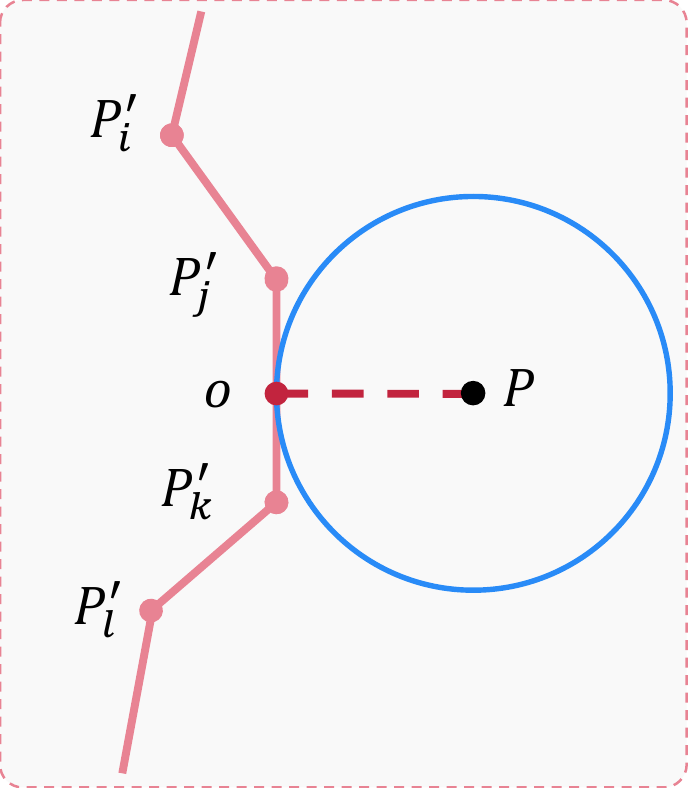} \\
	\end{minipage}
	}
	
	\subfigure[The tangent point of the ball is inside the triangle]{
	\begin{minipage}[t]{0.45\textwidth}
	\centering
	\includegraphics[width=\textwidth]{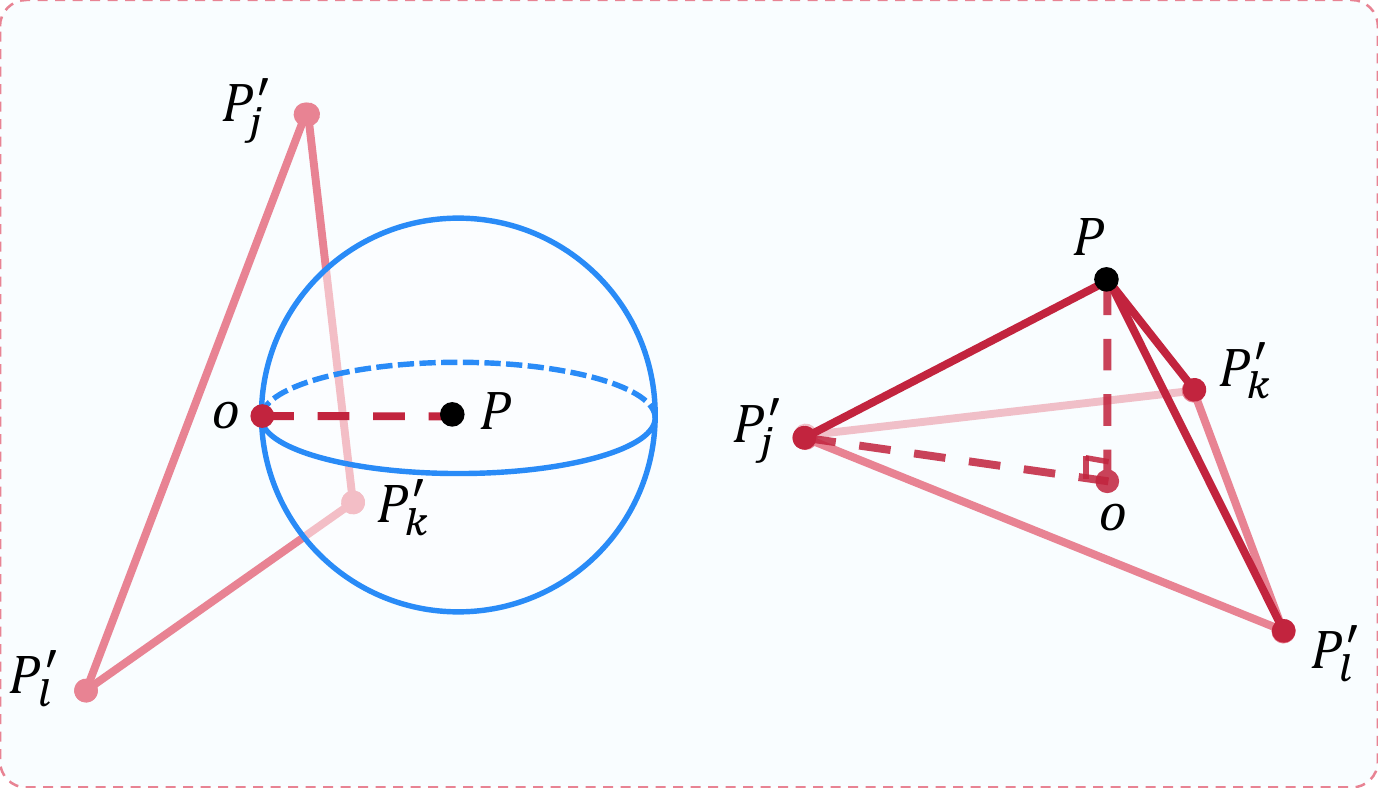} \\
	\end{minipage}
	}
	
	\caption{{\bf Three cases for approximation.} The basic assumption is that the surface is represented by a triangular patch, which can only appear either straight or flat, rather curved.} 
	\label{fig:approximate}
	\vspace{-0.2cm}
\end{figure}

Since we generate the point-to-surface distance ground truth from corresponding depth map, the error bound analysis is necessary. A reasonable assumption is to employ the triangulated mesh to represent the surface. There are three difference cases, as shown in \cref{fig:approximate}.

In case (a), the largest sphere centering at the query point $\mathbf{p}$ is tangent to the surface of the object at point $\mathbf{o}$.
Then, the ground truth of signed distance at $\mathbf{p}$ is ${d(\mathbf{p}, \mathbf{o})}$. The sided distance from the query point $\mathbf{p}$ to the sampled point set $\mathbf{\{p^\prime\}}$ is ${d(\mathbf{p}, \mathbf{p_j^\prime})}$. Since $\mathbf{p_j^\prime}$ coincides with the tangent point $\mathbf{o}$, the error for case $(a)$ is  ${e_a^2}$ as below
\vspace{-0.1cm}
\begin{equation}
\mathbf{e}_a^{2}=\left(d(\mathbf{p}, \mathbf{o}) - d\left(\mathbf{p}, \mathbf{p}_{\mathbf{j}}^{\prime}\right)\right)^{2}=0,
\vspace{-0.1cm}
\end{equation}
where $d(\mathbf{p}, \mathbf{o})$ and $d\left(\mathbf{p}, \mathbf{p}_{\mathbf{j}}^{\prime}\right)$ represent ground truth and the approximate value of signed distance, respectively. 

In case $(b)$ and $(c)$, we use the similar analytical method. 
Suppose $\mathbf{o^\prime}$ and $\mathbf{o^{\prime\prime}}$ are the tangent points of the surface and the sphere centering at $\mathbf{p}$.
The ground truth of signed distance in case $(b)$ is ${d(\mathbf{p}, \mathbf{o^\prime})}$, which is ${d(\mathbf{p}, \mathbf{o^{\prime\prime}})}$ in case $(c)$. Thus, the error range of case $(b)$ and $(c)$ can be expressed by the following equation
\vspace{-0.2cm}

$$
\mathbf{e}_b^{2} \leq \min \left( d\left(\mathbf{p_j^\prime}, \mathbf{o^\prime}\right)^2, d\left(\mathbf{p_k^\prime}, \mathbf{o^\prime}\right)^2 \right) \leq \frac{d\left(\mathbf{p_j^\prime}, \mathbf{\mathbf{p_k^\prime}}\right)^2}{4},
$$

\begin{equation}
\begin{aligned}
\mathbf{e}_c^{2} &\leq \min \left( d\left(\mathbf{p_j^\prime}, \mathbf{o^{\prime\prime}}\right)^2, d\left(\mathbf{p_k^\prime}, \mathbf{o^{\prime\prime}}\right)^2, d\left(\mathbf{p_l^\prime}, \mathbf{o^{\prime\prime}}\right)^2 \right) \\
  &\leq \frac{\min \left( d(\mathbf{p_j^\prime}, \mathbf{\mathbf{p_k^\prime}})^2, d(\mathbf{p_k^\prime}, \mathbf{\mathbf{p_l^\prime}})^2, d(\mathbf{p_j^\prime}, \mathbf{\mathbf{p_l^\prime}})^2 \right)}{3} \\
\end{aligned}
\end{equation}
where ${e_b^2}$ and ${e_c^2}$ are the square of error in case $(b), (c)$, respectively. By combining these three cases covering all possible situations, we obtain the final error bound for the query point ${\mathbf{p}}$ as follows

\begin{equation}
0 \leq \mathbf{e}^{2} = \max (\mathbf{e}_a^{2}, \mathbf{e}_b^{2}, \mathbf{e}_c^{2}) < d(\mathbf{p_j^\prime}, \mathbf{\mathbf{p_{j+1}^\prime}})^2,
\end{equation}
where $\mathbf{e}$ is the general error of the query point $\mathbf{p}$. $\mathbf{p_{j}^\prime}$ and $\mathbf{p_{j+1}^\prime}$ are the two adjacent surface points. This inequality shows that the square of error $\mathbf{e}^2$ does not exceed the square of distance between the two points that are reprojected from two adjacent pixels.

\section{Experiments}
In this section, we conduct the experiments and ablation studies on MVS benchmark datasets. The experimental results results show that our proposed RA-MVSNet approach achieves the start-of-the-art performance.

\subsection{Implementation Setup}
\subsubsection{Training.} 

\hspace{1pc} Like the previous methods~\cite{yao2018mvsnet, unimvsnet}, our proposed RA-MVSNet is trained on DTU dataset for DTU evaluation, which is finetuned on BlendedMVS dataset for Tanks and Temples benchmark. As for DTU dataset, we use 79 scenes for training, 18 scenes for validation and the rest of data for evaluation. The original image size is 1200 $\times$ 1600, and each scene have 7 different lighting conditions. We crop the rectified images into 512 $\times$ 640.  Meanwhile, we utilize the finer DTU ground truth as~\cite{wei2021aa}. Similar to~\cite{gu2020cascade}, we implement our RA-MVSNet in three stages with $\frac{1}{4}$, $\frac{1}{2}$ and original input images, respectively. From low-resolution to high-resolution stages, the number of depth hypothesis is 64, 32 and 8. Their corresponding depth intervals are set to 4, 2 and 1. In the training, the number of input images $N$ is set to 5. Thus, there are single reference image and four source images. Our model is trained for 16 epochs with Adam optimizer~\cite{kingma2014adam}. The initial learning rate is 0.001, which is multiplied by 0.5 after 10, 12 and 14 epochs. Since the premature introduction of 3D CNN for SDF prediction may lead to slow convergence, we start the training of this branch at 10 epoch. The fusion parameter $\theta$ is set to 0.1. As for BlendedMVS dataset, we train for 10 epochs with an initial learning rate of 0.0002, which is down-scaled by a factor of 2 after 6 and 8 epochs. During finetuning, the number of input images is 10 with the original size of 576 $\times$ 768. The batch size is 2 on two NVIDIA RTX 2080Ti for DTU dataset, which is set to one on single NVIDIA Tesla P40 for BlendedMVS dataset.

\begin{figure}[htb]
	\centering
	\subfigure{
		\begin{minipage}[t]{0.14\textwidth} 
		\centering
        \includegraphics[width=\textwidth]{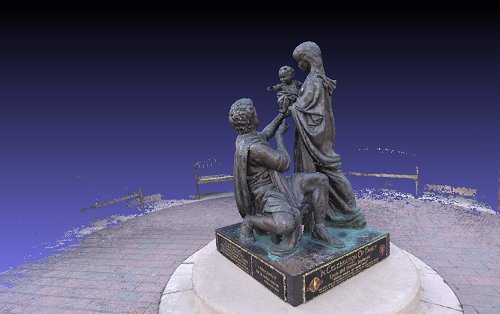} \\
        {Family}
		\end{minipage}
	}
	\subfigure{
    	\begin{minipage}[t]{0.14\textwidth}
    	\centering
    	\includegraphics[width=\textwidth]{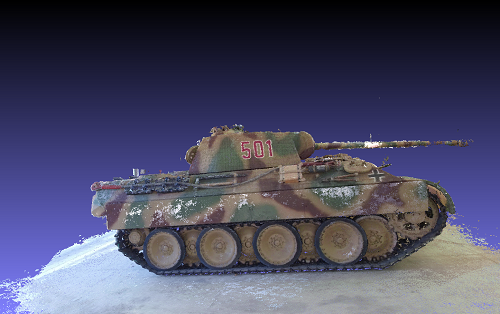} \\
    	{Panther}
	    \end{minipage}
	}
	\subfigure{
    	\begin{minipage}[t]{0.14\textwidth}
    	\centering
    	\includegraphics[width=\textwidth]{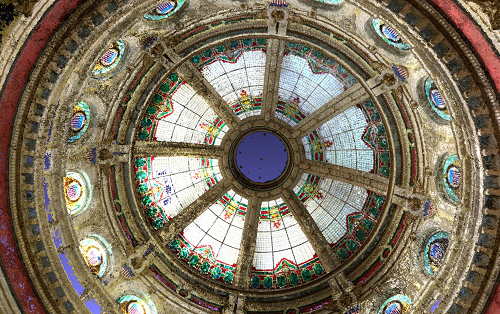} \\
    	{Museum}
	    \end{minipage}
	}
	\subfigure{
		\begin{minipage}[t]{0.14\textwidth} 
		\centering
        \includegraphics[width=\textwidth]{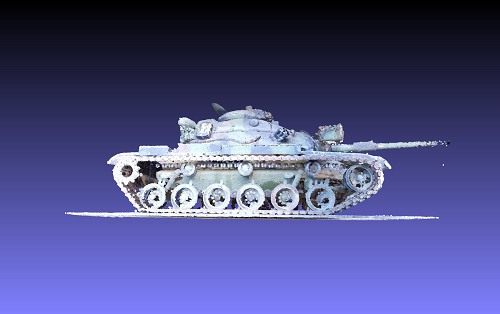} \\
        {M60}
		\end{minipage}
	}
	\subfigure{
    	\begin{minipage}[t]{0.14\textwidth}
    	\centering
    	\includegraphics[width=\textwidth]{figs/train00.png} \\
    	{Train}
	    \end{minipage}
	}
	\subfigure{
    	\begin{minipage}[t]{0.14\textwidth}
    	\centering
    	\includegraphics[width=\textwidth]{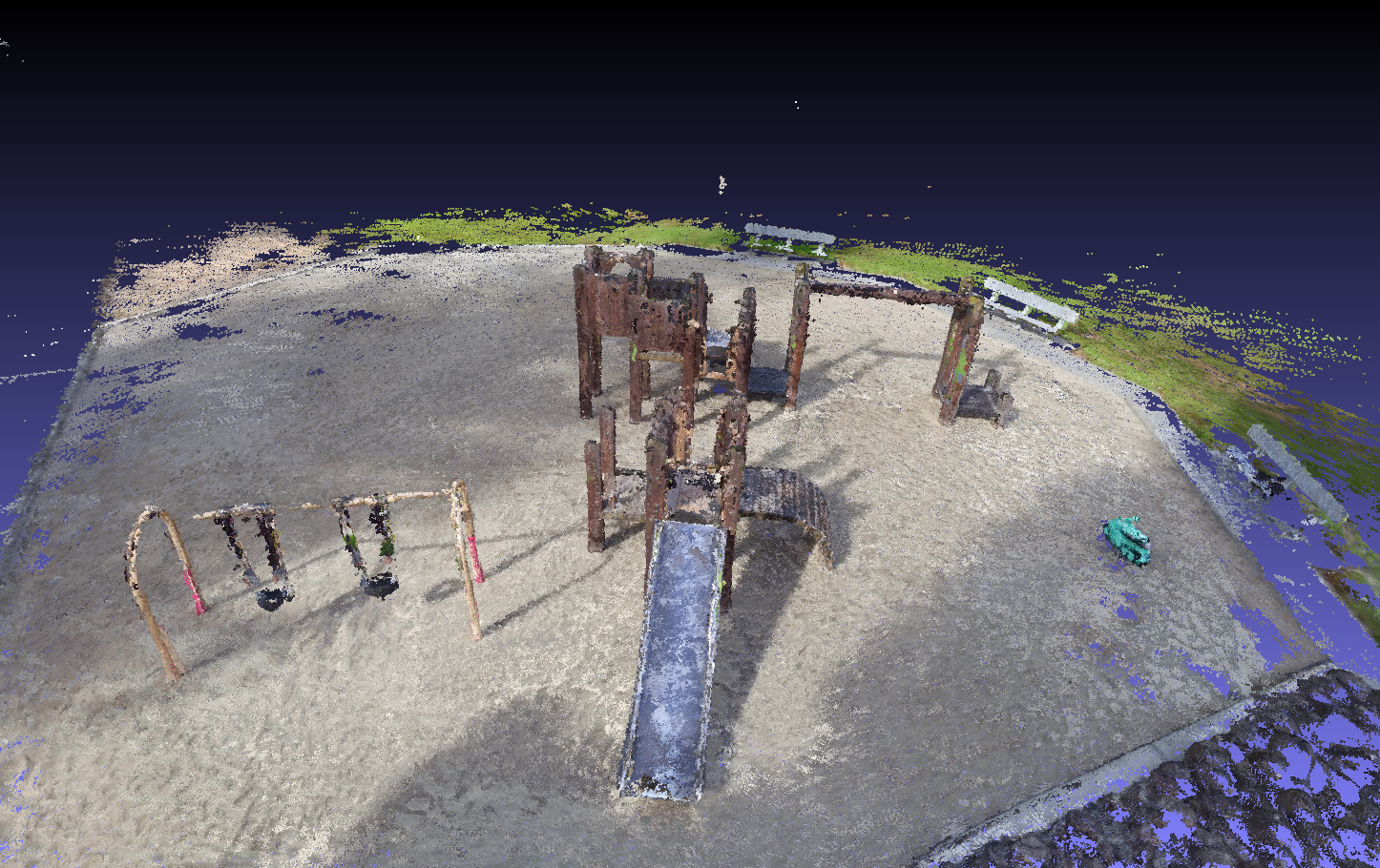} \\
    	{Playground}
	    \end{minipage}
	}
	\vspace{-0.4cm}
	\caption{{\bf Qualitative results on T\&T dataset.} Our RA-MVSNet still performs well in large outdoor scenes with complex lighting.} 
	\label{Tanks}
	\vspace{-0.4cm}
\end{figure}

\vspace{-0.3cm}
\subsubsection{Testing.} 
\hspace{1pc} When testing on the DTU dataset, the resolution is 864 $\times$ 1152, and the number of input images $N$ is set to 5. Besides, we set the number of hypothetical planes for the three stages to 64, 32, and 8, which are the same as training. As for Tanks and Temples dataset, the resolution of input images is either 1024 $\times$ 1920 or 1024$\times$ 2048. The number of input images is 11 like~\cite{yao2018mvsnet}. To evaluate on the DTU dataset and the Tanks and Temples dataset, we use NVIDIA Tesla P40 GPU with 24G RAM. For the results on DTU, we report the evaluation metrics (accuracy, completeness and overall) described in~\cite{aanaes2016large}. For the benchmark results on Tanks and Temples, we report the F-score metric.

\begin{figure*}[htb]
    \centering
    
    \subfigure{
		\begin{minipage}[t]{0.23\textwidth} 
		\centering
		\includegraphics[trim={0cm 2cm 0cm 0cm}, clip,  width=\textwidth]{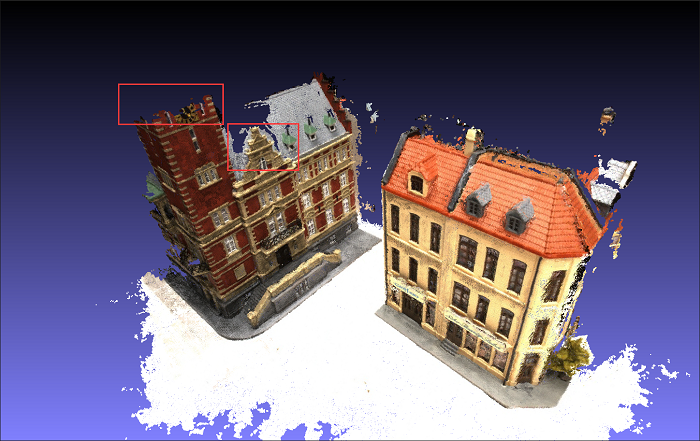} \\
		\end{minipage}
	}
    \subfigure{
		\begin{minipage}[t]{0.23\textwidth} 
		\centering
		\includegraphics[trim={0cm 2cm 0cm 0cm}, clip,  width=\textwidth]{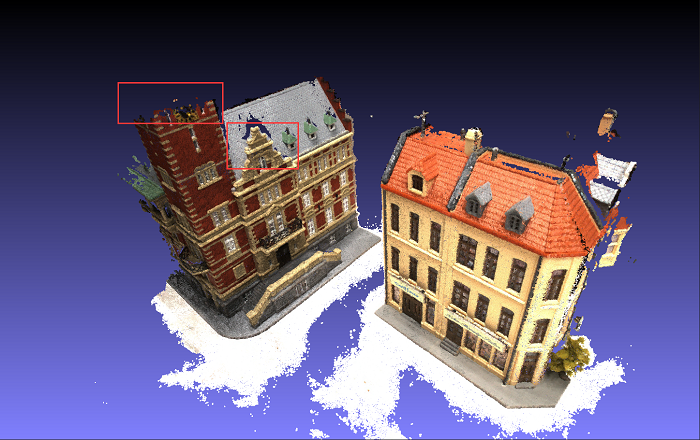} \\
		\end{minipage}
	}
	\subfigure{
		\begin{minipage}[t]{0.23\textwidth} 
		\centering
        \includegraphics[trim={0cm 2cm 0cm 0cm}, clip, width=\textwidth]{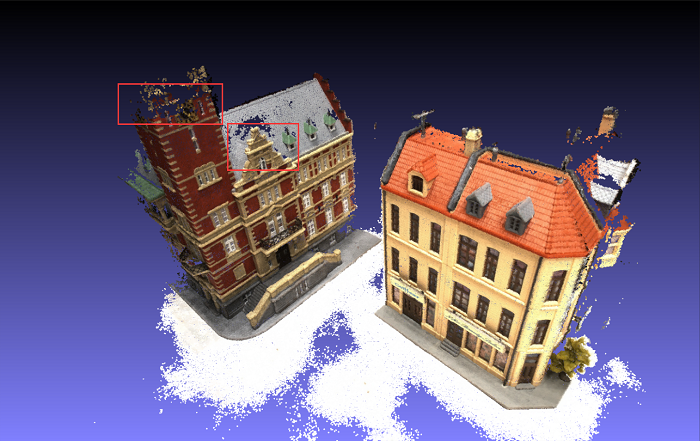} \\
		\end{minipage}
	}
	\subfigure{
		\begin{minipage}[t]{0.23\textwidth} 
		\centering
        \includegraphics[trim={0cm 2cm 0cm 0cm}, clip, width=\textwidth]{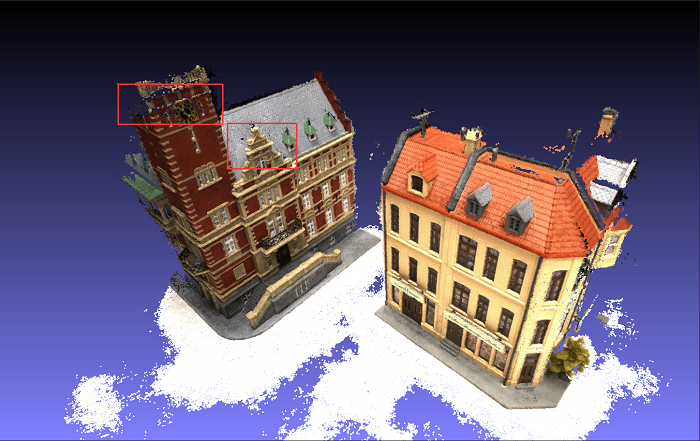} \\
		\end{minipage}
	}
	
	\subfigure{
		\begin{minipage}[t]{0.23\textwidth} 
		\centering
        \includegraphics[trim={0cm 1cm 0cm 0cm}, clip, width=\textwidth]{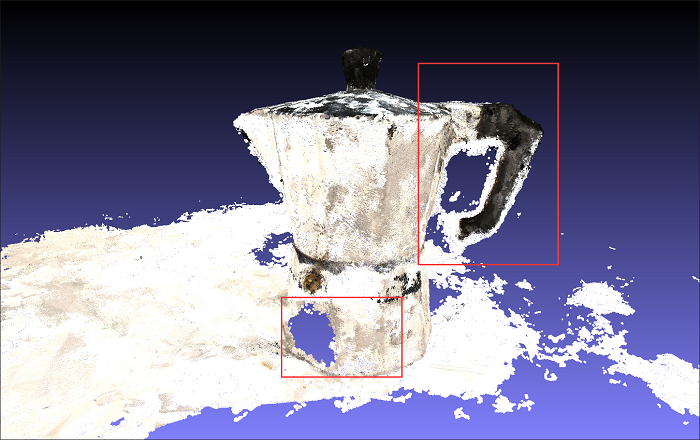} \\
        {CasMVSNet}
		\end{minipage}
	}
	\subfigure{
		\begin{minipage}[t]{0.23\textwidth} 
		\centering
        \includegraphics[trim={0cm 1cm 0cm 0cm}, clip, width=\textwidth]{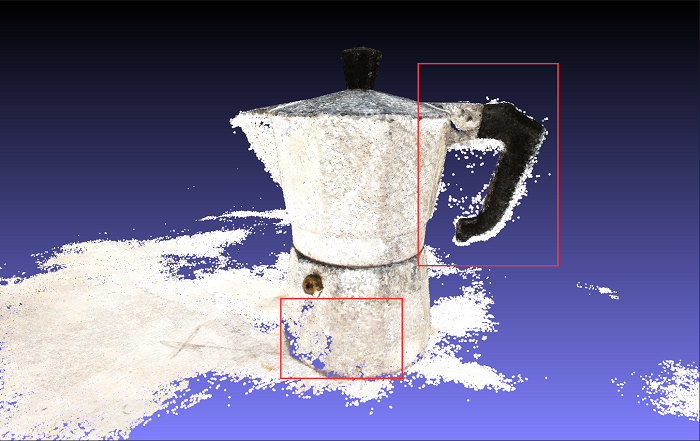} \\
        {TransMVSNet}
		\end{minipage}
	}
	\subfigure{
		\begin{minipage}[t]{0.23\textwidth} 
		\centering
        \includegraphics[trim={0cm 1cm 0cm 0cm}, clip, width=\textwidth]{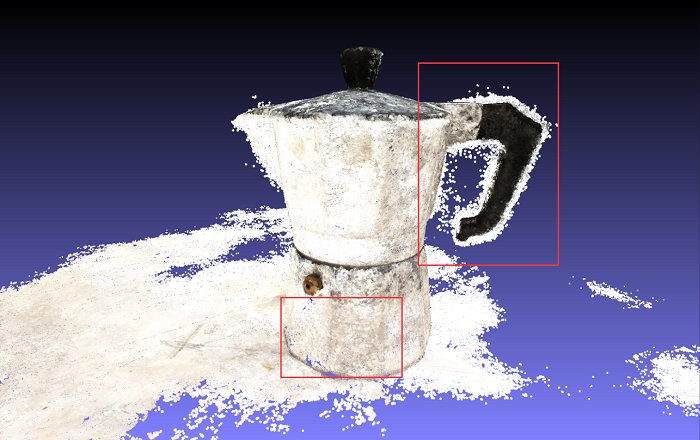} \\
        {UniMVSNet}
		\end{minipage}
	}
	\subfigure{
		\begin{minipage}[t]{0.23\textwidth} 
		\centering
        \includegraphics[trim={0cm 1cm 0cm 0cm}, clip, width=\textwidth]{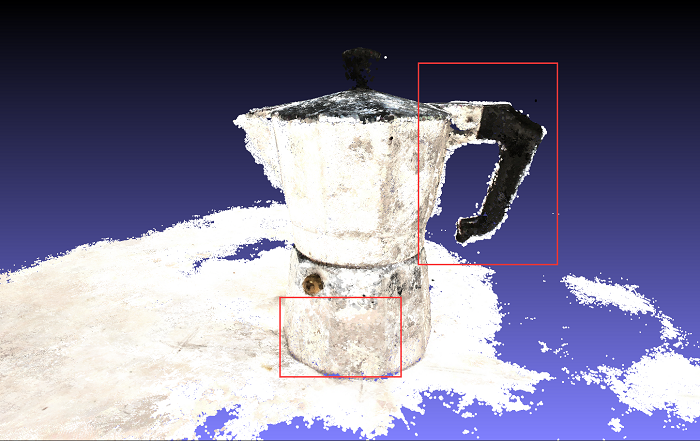} \\
        {Ours}
		\end{minipage}
	}
    \vspace{-0.3cm}
    \caption{{\bf Qualitative results on DTU dataset.} Our RA-MVSNet produces more complete and less outlier results than the previous methods like Uni-MVSNet~\cite{unimvsnet} and TransMVSNet~\cite{ding2022transmvsnet}.} 
    \label{DTU}
\end{figure*}

\subsection{Results on Tanks and Temples}
Our RA-MVSNet maintains the promising performance in large-scale, complex lighting scenes, which achieves the best score on Tanks and Temples dataset. Similar to the previous method~\cite{yan2020dense}, we employ the dynamic geometric consistency strategies. 

For fair evaluation, we compare our method against current excellent work. The corresponding quantitative results on intermediate and advanced sets are reported in \Cref{tb:tank_compare}. It can be clearly seen that our method achieves the state-of-the-art performance on both intermediate and advanced sets. Specifically, RA-MVSNet obtains the best F-score of $\mathbf{65.72}$ and $\mathbf{39.93}$ (higher is better) on intermediate and advanced subset, respectively. Moreover, it performs the best on 5 scenes and 3 scenes of two subset, respectively. Notably, intermediate subset mostly contains one object to be reconstructed, and the advanced subset has the large-scale outdoor scenes. Our method achieves the best performance on both subsets. This shows that our model is effective in various scenarios. Moreover, \cref{Tanks} gives some qualitative results on two subsets, which demonstrate that our model exhibits the strongest generalization and robustness in textureless and object boundary regions.

\begin{table}[htb]
  \begin{center}
  \resizebox{1.0\linewidth}{!}{
  \begin{tabular}{l|ccc}

  \toprule
  Method & ACC.(mm) $\downarrow$ & Comp.(mm) $\downarrow$ & Overall(mm) $\downarrow$ \\
  \hline
  Furu~\cite{furukawa2009accurate} & 0.613 & 0.941 & 0.777 \\
  Gipuma~\cite{galliani2015massively} & {\bf 0.283} & 0.873 & 0.578 \\
  COLMAP~\cite{schonberger2016pixelwise} & 0.400 & 0.664 & 0.532 \\
  \hline
  SurfaceNet~\cite{ji2017surfacenet} & 0.450 & 1.040 & 0.745 \\
  MVSNet~\cite{yao2018mvsnet} & 0.396 & 0.527 & 0.462 \\
  Point-MVSNet~\cite{chen2019point} & 0.342 & 0.411 & 0.376 \\
  AA-RMVSNet~\cite{wei2021aa} & 0.376 & 0.339 & 0.357 \\
  CasMVSNet~\cite{gu2020cascade} & 0.325 & 0.385 & 0.355 \\
  UCS-Net~\cite{cheng2020deep} & 0.338 & 0.349 & 0.344 \\
  Uni-MVSNet~\cite{unimvsnet} & 0.352 & 0.278 & 0.315 \\
  TransMVSNet~\cite{ding2022transmvsnet}  & 0.321 & 0.289 & 0.305 \\
  GBi-Net~\cite{mi2022generalized} & 0.327 & { \bf 0.268} & 0.298 \\
  \hline
  \hline
  RA-MVSNet (ours) & 0.326 & { \bf 0.268} & { \bf 0.297}  \\
  \bottomrule
  \end{tabular}
  }
  \end{center}
  \vspace{-0.4cm}
  \caption{{\bf Quantitative results on DTU evaluation set.} The best results in each category are in {\bf bold}. Our model ranks the first in terms of Completeness and Overall metrics.}
  \label{tb:dtu_compare}
  \vspace{-0.6cm}
\end{table}

\begin{table*}
  \begin{center}
  \resizebox{1.0\linewidth}{!}{
  \begin{tabular}{l|cc|cc|ccc|ccc}
  \toprule
  \multirow{2}*{Method} & \multicolumn{2}{c|}{Branch} & \multicolumn{2}{c|}{Representation} & \multicolumn{3}{c|}{DTU} & \multicolumn{3}{c}{Tanks and Temples} \\
  \cline{2-11}
  & depth & SDF & point clouds & mesh & ACC.(mm) & Comp.(mm) & Overall(mm) & Prec. & Rec. & F-score \\
  \hline
  Baseline & \checkmark & & \checkmark & & 0.348 & 0.290 & 0.319 & 56.62 & 75.35 & 64.02 \\
  \hline
  Two-branch(W/O fusion) & \checkmark & \checkmark & \checkmark & \checkmark &  0.357 & {\bf 0.262} & 0.310 & 56.17 & {\bf 77.65} & 64.62 \\
  Two-branch(With fusion) & \checkmark & \checkmark & \checkmark & \checkmark & 0.330 & 0.274 & 0.302 & 57.58 & 77.21 & 65.39 \\
  Two-branch(With fusion) + RFP & \checkmark & \checkmark & \checkmark & \checkmark & {\bf 0.326} & 0.268 & {\bf 0.297} & {\bf 58.68} & 75.23 & {\bf 65.72} \\
  \hline
  \end{tabular}
  }
  \end{center}
  \vspace{-0.5cm}
 
    \caption{{\bf Ablation study on DTU and T\&T evaluation set.} "RFP" refer to Recursive Feature Pyramid for feature extraction. The Baseline is the original CasMVSNet \cite{gu2020cascade}. Our RA-MVSNet with fusion of two branches outperforms in the overall metric. }
  \label{tb:ab}
  \vspace{-0.45cm}
\end{table*}

\vspace{-0.1cm}
\subsection{Results on DTU}
As in~\cite{gu2020cascade, yao2018mvsnet, yao2020blendedmvs}, we make use of geometric and photometric constraints for filtering. Moreover, we employ the fusion method in Gipuma~\cite{galliani2015massively} similar to~\cite{gu2020cascade, yao2018mvsnet, yao2020blendedmvs}. The final results are evaluated on DTU testing set by two metrics, accuracy and completeness. We compare our RA-MVSNet with previous methods. The quantitative results are summarized in \Cref{tb:dtu_compare}. It can be observed that our RA-MVSNet outperforms both traditional methods and learning-based approaches. For the accuracy, the traditional method \cite{galliani2015massively} achieves the best results. For the completeness metric, our method achieves the state-of-the-art performance. Overall, our RA-MVSNet method ranks the first, which achieves $\mathbf{0.297}$ score. \cref{DTU} shows some qualitative results on DTU testing dataset compared against other methods. It can be seen that RA-MVSNet obtains more complete reconstruction results with less outlier.

\subsection{Ablation Studies}
As metioned above, we introduce the signed distance prediction branch in MVS network, which can not only improve the completeness of reconstruction results but also generate the explicit mesh surfaces. In the following, we show the effectiveness of the introduced branch. CasMVSNet~\cite{gu2020cascade} is treated as baseline, which only has the depth prediction branch at each stage. Besides, we explore the effect of fusion parameter $\theta$ on performance.

\vspace{-0.3cm}

\subsubsection{SDF Branch}
\vspace{-0.1cm}

\hspace{1pc} As shown in \Cref{tb:ab}, we evaluate the performance of baseline and the model with SDF supervised branch. From the results, it can be obviously seen that the SDF branches are able to improve the performance, especially on the integrity of reconstruction results. This is because the approximate signed distance introduces the extra supervision on depth map prediction. The comparison experiment of whether to fuse these two branches shows that the introduction of SDF branch fusion can effectively improve the accuracy. This is because the SDF branch can remove the outliers whose sign distance exceeds the threshold. Meanwhile, our RA-MVSNet is able to generate both point clouds and mesh using two branches, respectively. However, the baseline method only employs point clouds as representation. 

\begin{figure}[htb]
	\centering
	\subfigure[Point cloud metric on DTU]{
		\begin{minipage}[t]{0.225\textwidth} 
		\centering
        \includegraphics[trim={1.5cm 8.5cm 1.5cm 10.5cm}, clip,width=\textwidth]{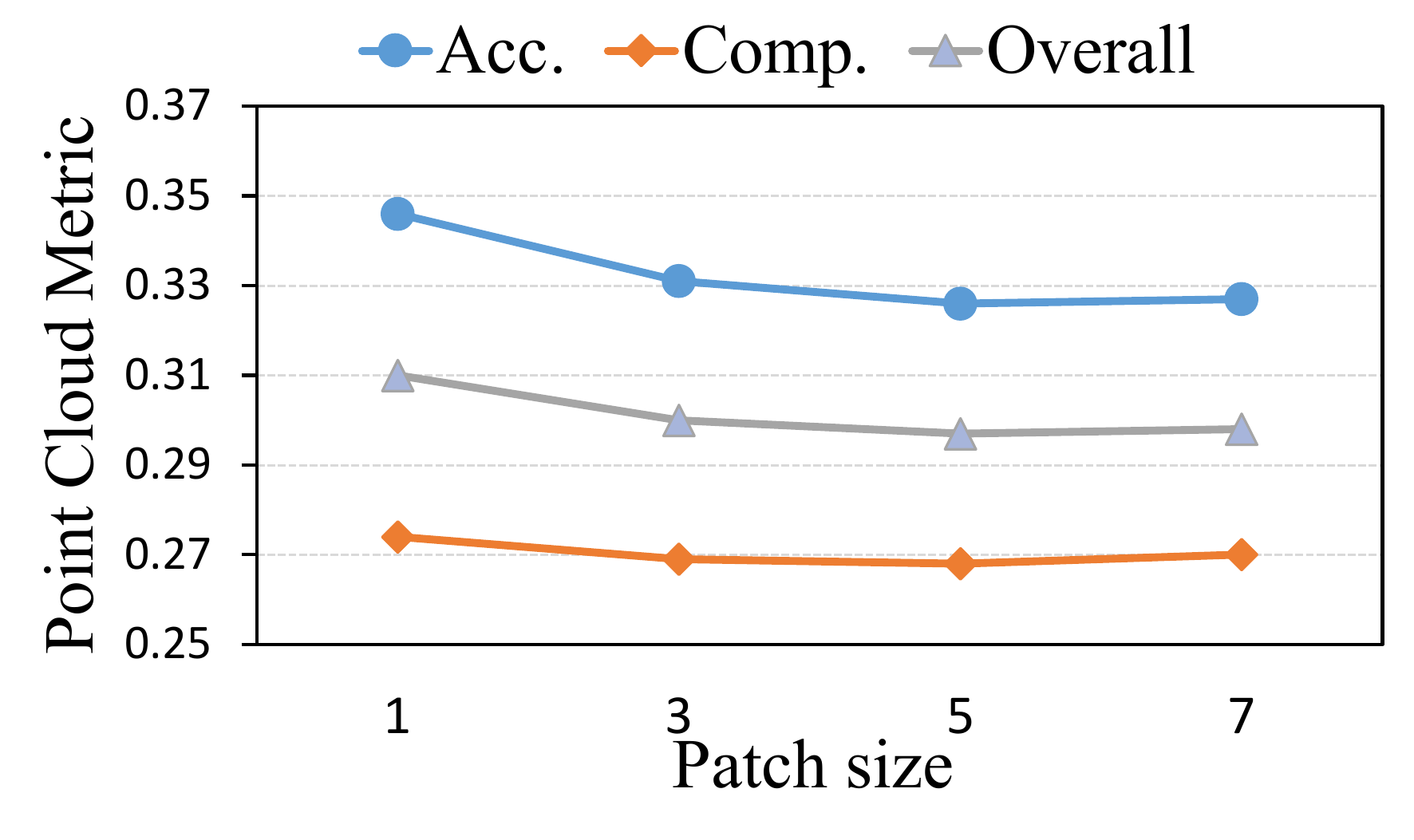} \\
		\end{minipage}
	}
	\subfigure[Percentage of $<n$ of the depth maps on BlendedMVS]{
	\begin{minipage}[t]{0.225\textwidth}
	\centering
	\includegraphics[trim={1.5cm 2cm 1.5cm 3cm}, clip,width=\textwidth]{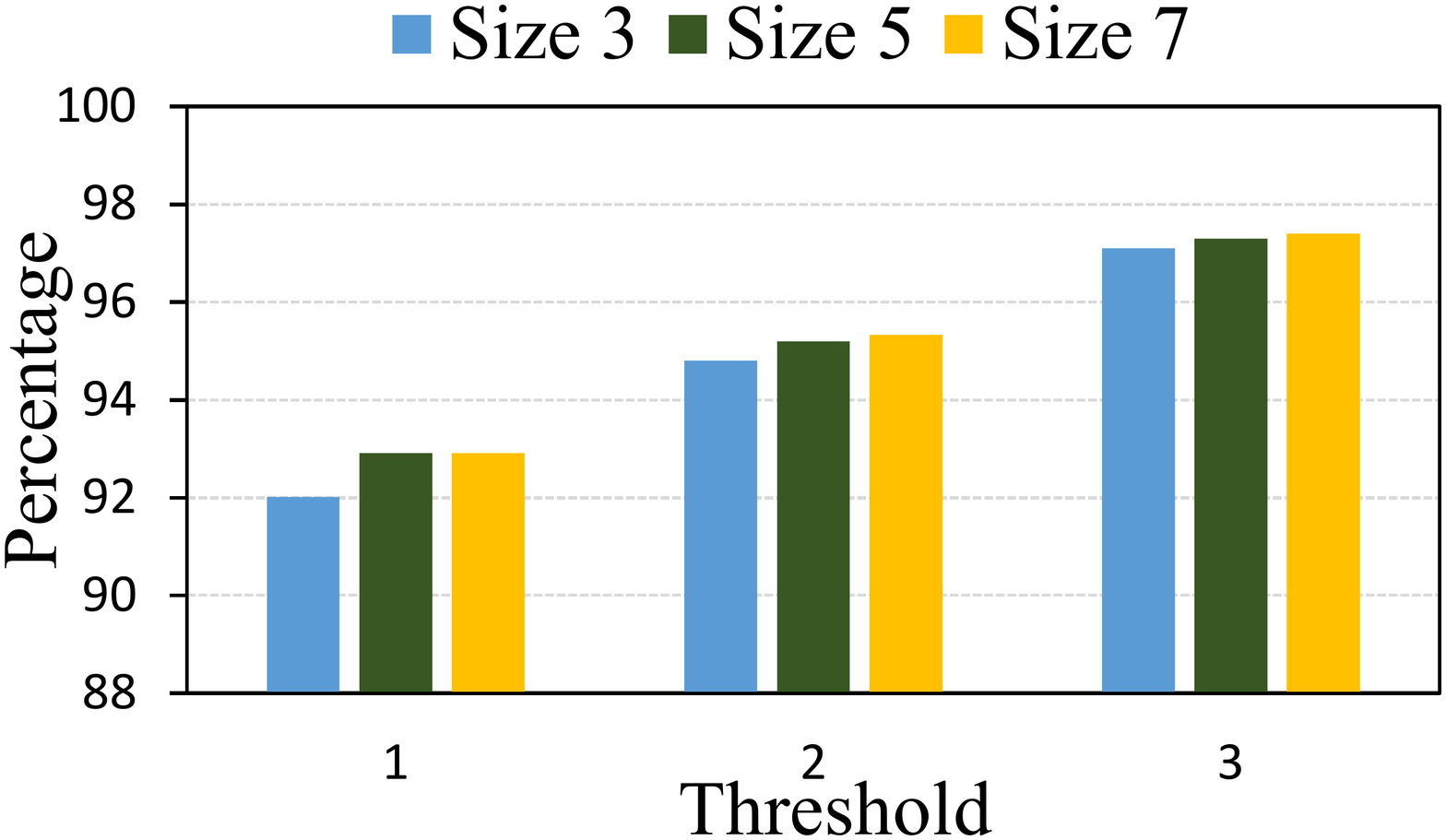} \\
	\end{minipage}
	}
	\caption{{\bf Ablation studies on patch size.} We use depth metrics and point cloud metrics to evaluate on BlendedMVS and DTU.} 
	\label{fig:patch}
    \vspace{-0.4cm}
\end{figure}

\vspace{-0.3cm}
\subsubsection{Local Patch Size}

\hspace{1pc} Instead of using global search, it is very efficient to compute the ground truth of distance volume in training by local search. Obviously, the size of patch affects the accuracy which affects the performance of RA-MVSNet. Therefore, we examine the performance of the model with different patch sizes, as shown in \cref{fig:patch}. When $k$ becomes large, the performance of patch-based local search gradually increases. Moreover, the improvement is saturated after $k$ exceeds 5. Therefore, we finally choose $k=5$ as the patch size.

\vspace{-0.3cm}
\subsubsection{Fusion Threshold $\theta$}
\hspace{1pc} The threshold $\theta$ is the parameter that trades off the fusion of two branches. From the essence of signed distance, the point closer to the surface has a smaller signed distance value. Therefore, we finally set $\theta$ to 0.1 as the threshold in this paper. As shown in \Cref{tb:ab_threshold}, we compare the performance of different models with various thresholds. The model without fusing two branches is treated as the baseline. The experimental results show that the reconstruction accuracy gradually decreases with the larger threshold $\theta$ while the completeness gradually increases.

\begin{table}[htb]
  \vspace{-0.1cm}
  \begin{center}
  \resizebox{1.0\linewidth}{!}{
  \begin{tabular}{l|ccc}
  \hline
  Method & ACC.(mm) & Comp.(mm) & Overall(mm) \\
  \hline
  Baseline(W/O fusion) & 0.357 & {\bf 0.262} & 0.310 \\
  \hline
  RA-MVSNet($\theta=0.1$) & {\bf 0.326} & 0.268 & {\bf 0.297} \\

  RA-MVSNet($\theta=0.2$) & 0.338 & 0.266 & 0.302 \\

  RA-MVSNet($\theta=0.5$) & 0.355 & 0.265 & 0.310 \\

  RA-MVSNet($\theta=1.0$) & 0.357 & {\bf 0.262} & 0.310 \\
  \hline
  \end{tabular}
  }
  \end{center}
  \vspace{-0.4cm}

  \caption{{\bf Ablation study on fusion threshold.}}
  \label{tb:ab_threshold}
 \vspace{-0.5cm}
\end{table}

\section{Conclusion}
In this paper, we proposed a novel RA-MVSNet approach to recover the detailed 3D scenes by taking advantage of cost volume using both depth and SDF branches. The SDF supervision enabled more hypothetical planes for the depth prediction, especially in textureless and boundary regions. Furthermore, the sided distance was employed to represent the ground truth signed distance for training, which can be computed efficiently. Our proposed RA-MVSNet approach achieves the promising results on several challenging datasets, which outperforms the state-of-the-art methods.

In the future, we plan to only use the SDF branch for MVS, which can effectively reduce memory consumption.

\section*{Acknowledgments} This work is supported by National Natural Science Foundation of China under Grants (61831015). It is also supported by Information Technology Center and State Key Lab of CAD\&CG, Zhejiang University.

{\small
\bibliographystyle{ieee_fullname}
\bibliography{egbib}
}

\clearpage
\appendix
\section*{Appendix}

\section{More Ablation Studies}

\subsection{Image Resolution}

The point cloud is generated by pixel-wise depth map reprojection. Therefore, the image resolution affects the precision and recall of 3D metrics. Besides, the scaled image inevitably uses the interpolation methods, which may lead to the unstable results. We explored the performance and memory usage of the same model using different image resolutions in inference on the \textit{Tanks and Temples datasets}~\cite{knapitsch2017tanks}, as shown in \cref{tb:reso}. 
As the resolution increases, the accuracy and recall rate of the model will increase along with the memory. Our model achieves the competitive results at half the resolution by using {\bf 3.6 GB} memory.

\begin{table}[htb]
  \vspace{-0.1cm}
  \begin{center}
  \resizebox{1.0\linewidth}{!}{
  \begin{tabular}{l|cccc}
  \hline
  Resolution & Prec. & Rec. & F-Score & Mem. \\
  \hline
  540 $\times$ 1024 (n=11) & 50.21 & 68.55 & 57.69 & {\bf 3702M} \\
  810 $\times$ 1536 (n=11) & 56.38 & 73.70 & 63.62 & 6799M \\
  \hline
  1080 $\times$ 2048 (n=11) & {\bf 58.68} & {\bf 75.23} & {\bf 65.72} & 10527M \\
  \hline
  \end{tabular}
  }
  \end{center}
  \vspace{-0.2cm}

  \caption{{\bf Ablation study on image resolution.}}
  \label{tb:reso}
 \vspace{-0.2cm}
\end{table}

\subsection{Image Encoder}
The baseline image encoder employs a 6-layer FPN structure. Instead, we employ Recursive Feature Pyramid (RFP)\cite{detectors} structure as image encoder in order to pay more attention to the object to be reconstructed. To examine the performance of this module and the cost of increased memory, we compare the results of two image encoders, as shown in \cref{tb:encoder}. It can be seen that RFP structure is able to improve the both accuracy and recall performance.

\begin{table}[htb]
  \vspace{-0.1cm}
  \begin{center}
  \resizebox{1.0\linewidth}{!}{
  \begin{tabular}{l|ccc|c}
  \hline
  Encoder & Acc. & Comp. & Overall & Mem.(on DTU~\cite{aanaes2016large}) \\
  \hline
  FPN (n=5) & 0.330 & 0.274 & 0.302 & 8624M \\
  RFP (n=5) & 0.326 & 0.268 & 0.297 & 9459M \\
  \hline
  \end{tabular}
  }
  \end{center}
  \vspace{-0.2cm}

  \caption{{\bf Ablation study on image encoder.}}
  \label{tb:encoder}
 \vspace{-0.1cm}
\end{table}

In addition, we compare the memory usage of two branches using different image encoders, as shown in \cref{tb:branch}. The memory consumption from two branches is far less than the increment by different image encoders. Therefore, our model can significantly improve model performance with a small overhead on memory usage.

\begin{table}[htb]
  \vspace{-0.1cm}
  \begin{center}
  \resizebox{1.0\linewidth}{!}{
  \begin{tabular}{l|cc}
  \hline
  Method & Mem.(with FPN) & Mem.(with RFP) \\
  \hline
  Baseline (n=5) & 8487M & 9276M \\
  Two-brance (n=5) & 8624M & 9459M \\
  \hline
  \end{tabular}
  }
  \end{center}
  \vspace{-0.2cm}

  \caption{{\bf Ablation study on memory.}}
  \label{tb:branch}
 \vspace{-0.2cm}
\end{table}

\subsection{Input view}
\begin{figure}[htb]
  \begin{center}
     \includegraphics[trim={2cm 3cm 1.5cm  4cm},clip,width=0.95\linewidth]{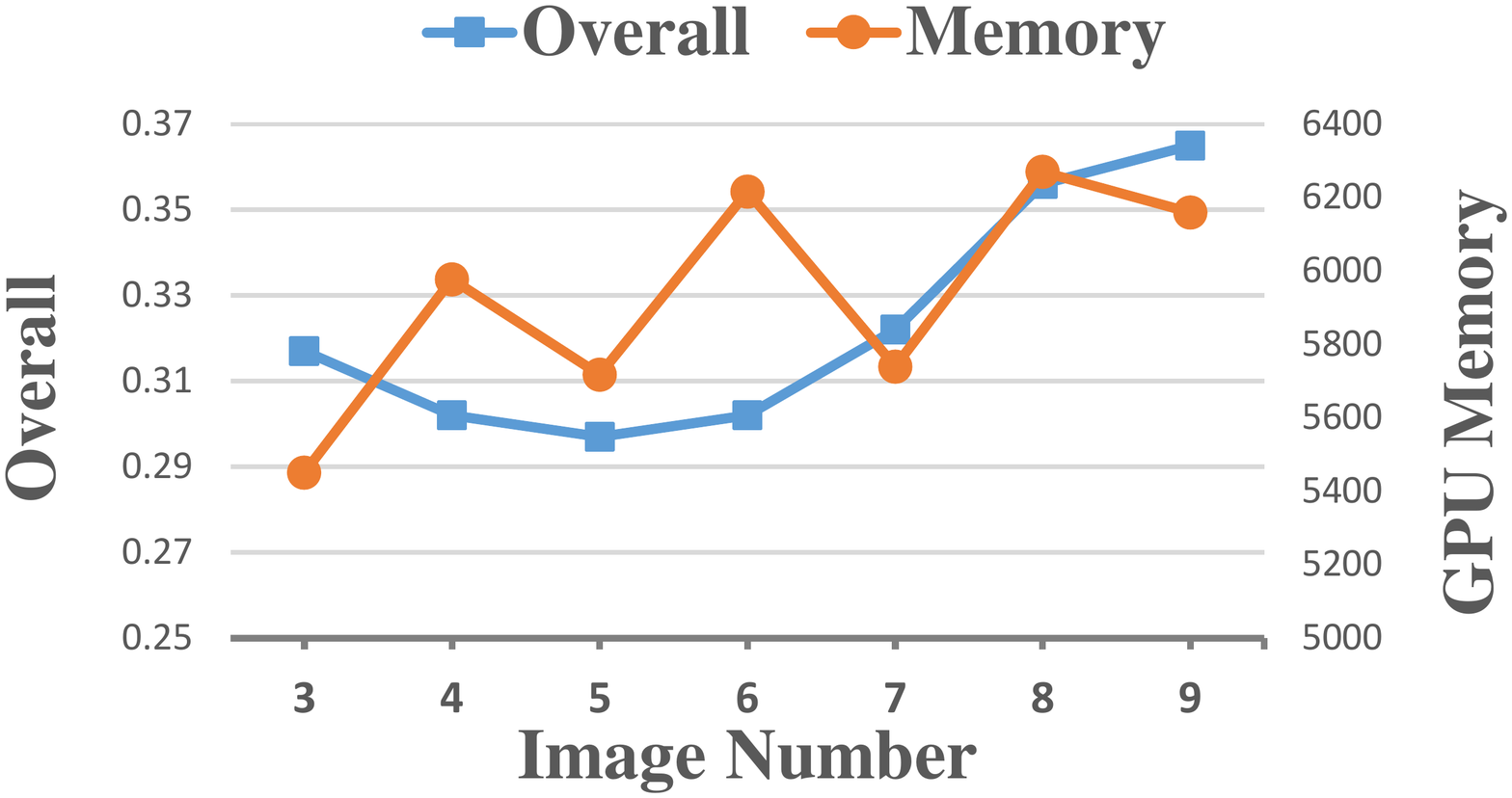}
  \end{center}
  \vspace{-0.2cm}
  \caption{{\bf Ablation study on input number $N$.} The accuracy varies with the number of input images $N$.}
  \label{fig:input}
  \vspace{-0.2cm}
\end{figure}

According to the setting, the input $N$ images include a reference view and several source images. We use a differentiable homography warping from the features of the source image to the reference view. In addition, we follow the source image selection method as MVSNet~\cite{yao2018mvsnet}, and a reference image matches up to 10 source images. Therefore, the number of input image $N$ will affect the performance of the model. We evaluated on the DTU dataset and plot the relationship between the performance and the number of input images $N$, as shown in \cref{fig:input}.

As the number of input source images $N$ increases, the performance on the DTU dataset does not increase monotonically. The performance is optimal when $N=5$. This is due to the large interval between adjacent images on the DTU dataset. As the number of input images increases, the overlapping area decreases, which will lead to performance degradation. 

\section{More Visualization Result}
\begin{figure*}[htb]
  
  \begin{center}
     \includegraphics[trim={0cm 1cm 2cm 1cm},clip,width=0.9\linewidth]{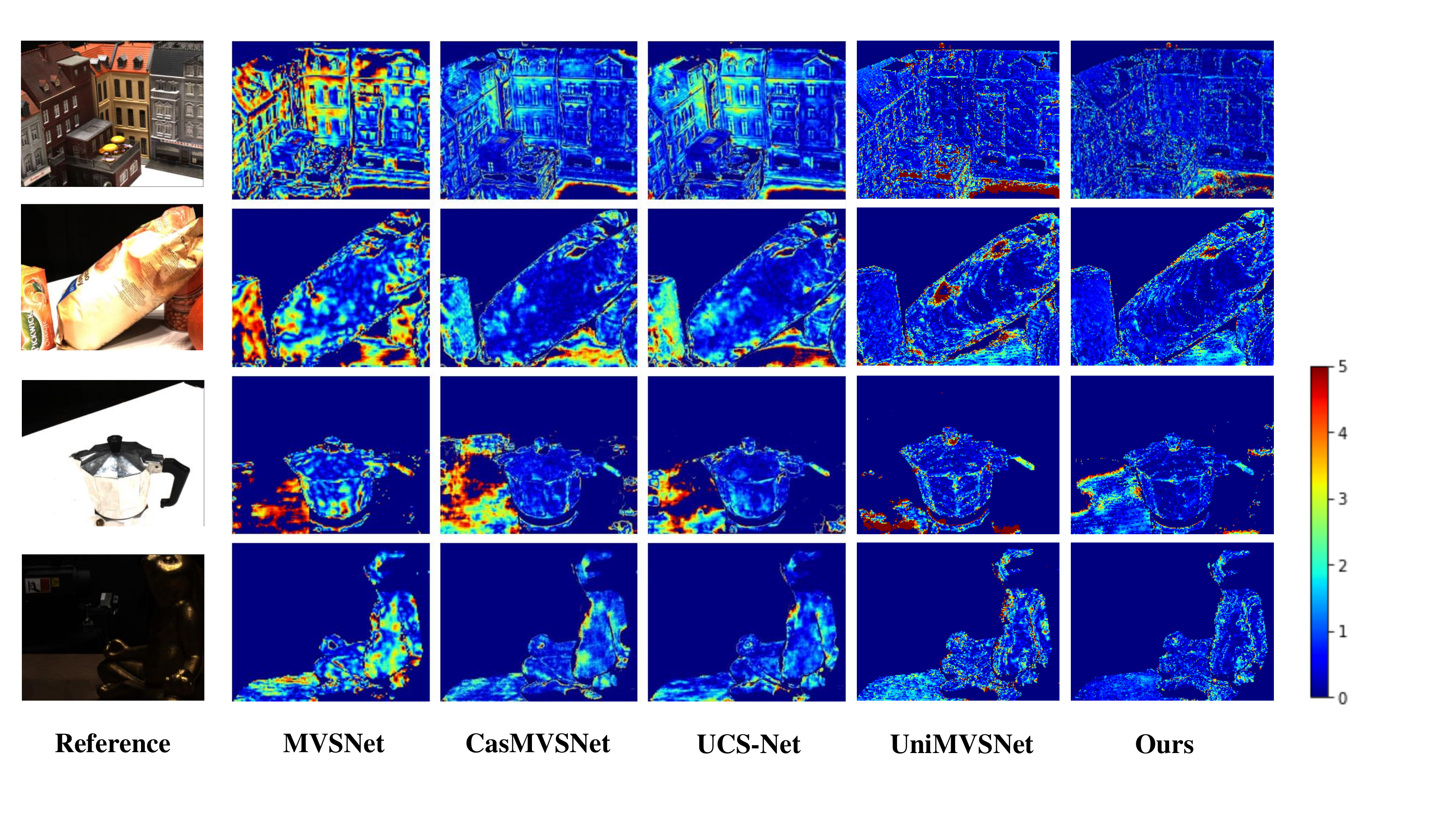}
  \end{center}
  \vspace{-0.6cm}
  \caption{{\bf Depth map metric.} }
  \label{fig:depthmap}
  \vspace{-0.2cm}
\end{figure*}

We show more qualitative results of the proposed model in this section. As shown in \cref{fig:depthmap}, the results of different methods are compared under the depth map metric. Our method still achieves the competitive results under 2D metrics. We show the reconstruction results on \textit{Tanks and Temples datasets} with gif files in the supplemental materials. Furthermore, \cref{fig:DTU} shows the reconstruction results of both in point cloud and mesh representation on \textit{DTU}. For better visualization of the results, we manually crop the mesh representation. Finally, the reconstruction results on \textit{Tanks and Temples}, and their corresponding precision and recall errors are shown in \cref{fig:TI} and \cref{fig:TA}

\begin{figure*}[htbp]
    \centering
    \begin{minipage}[t]{0.22\textwidth}
        \centering
        \includegraphics[width=1\textwidth]{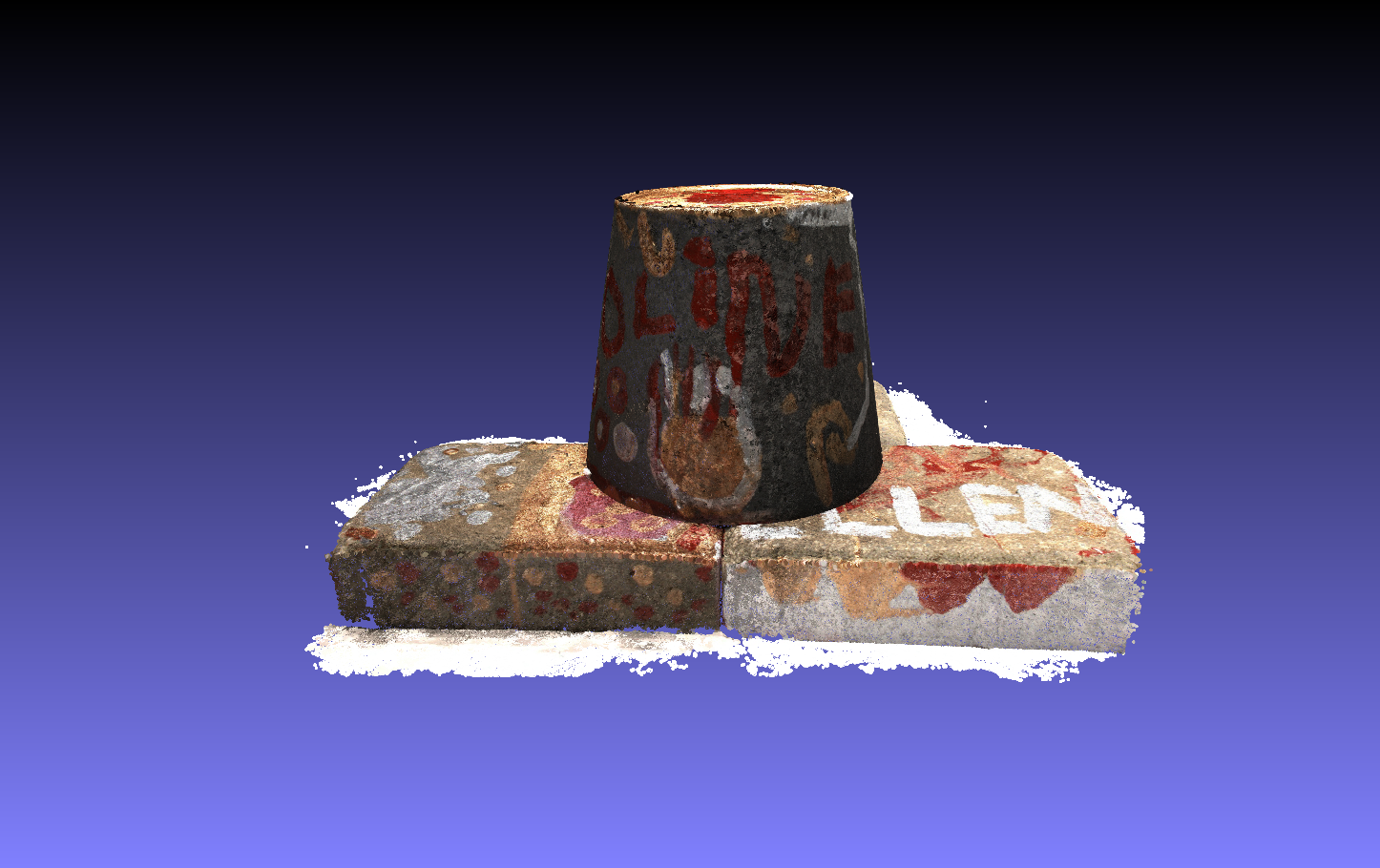}
    \end{minipage}
    \begin{minipage}[t]{0.216\textwidth}
        \centering
        \includegraphics[width=1\textwidth]{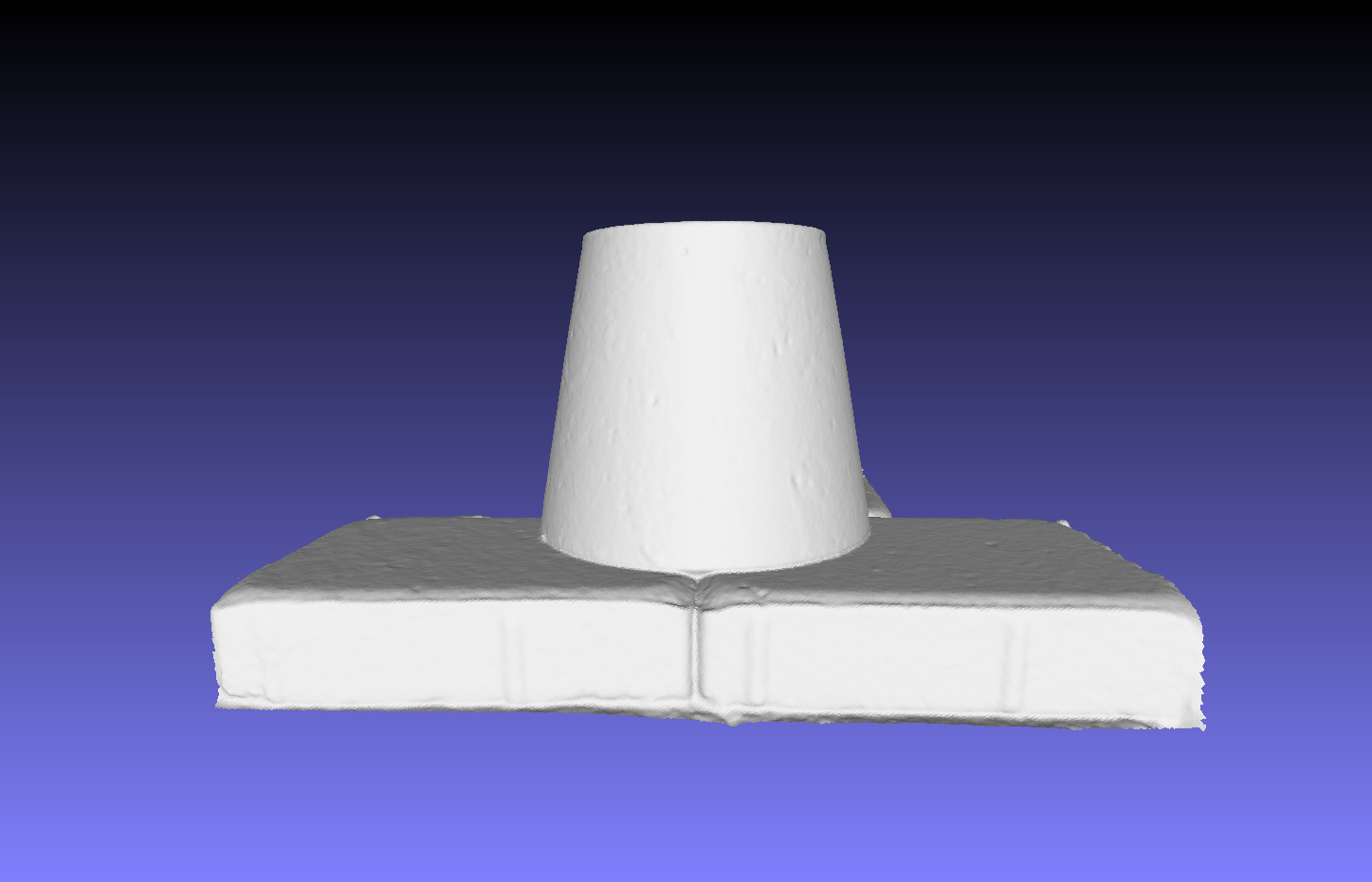}
    \end{minipage}
    \hspace{0.03\textwidth}
    \begin{minipage}[t]{0.22\textwidth}
        \centering
        \includegraphics[width=1\textwidth]{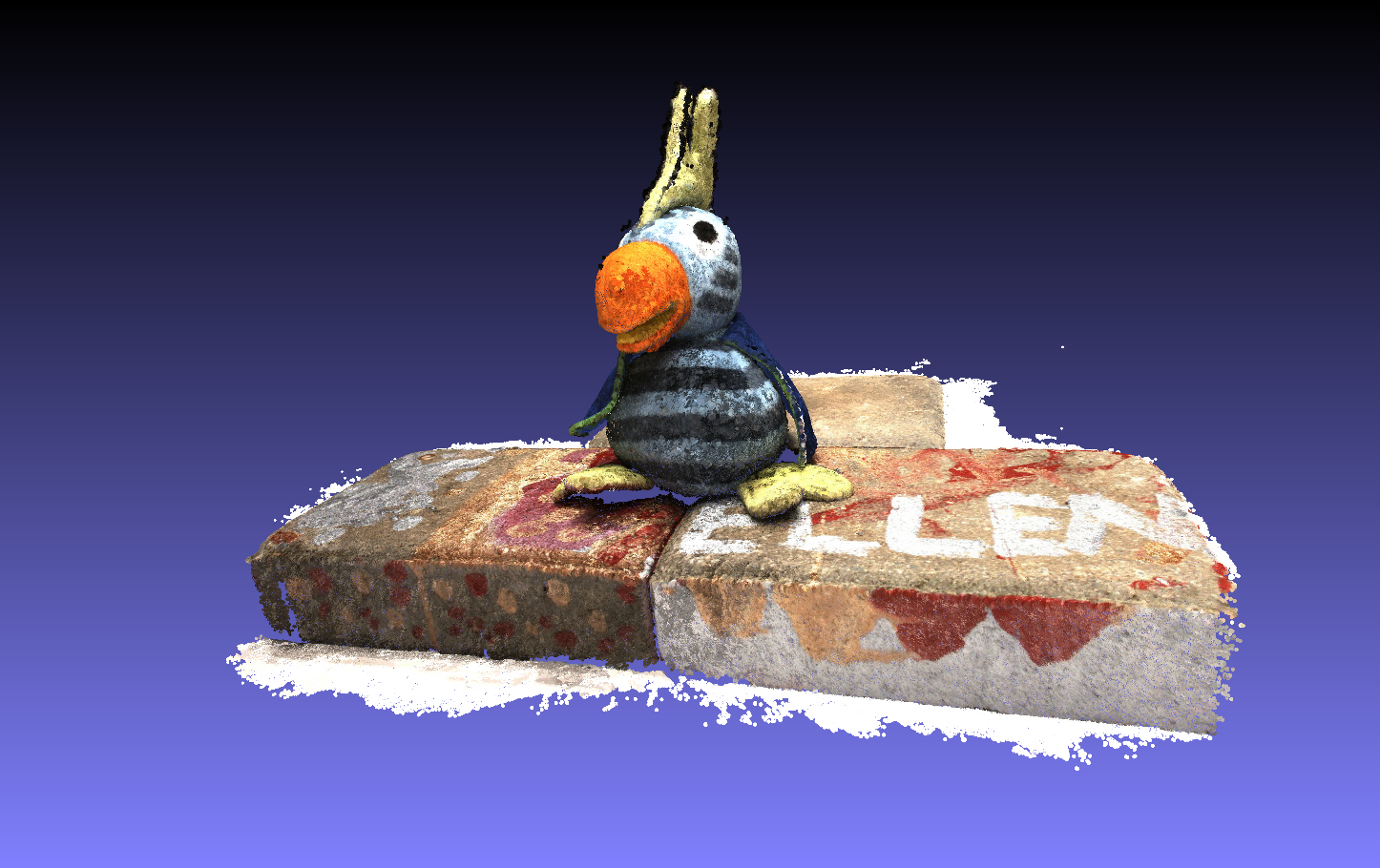}
    \end{minipage}
    \begin{minipage}[t]{0.216\textwidth}
        \centering
        \includegraphics[width=1\textwidth]{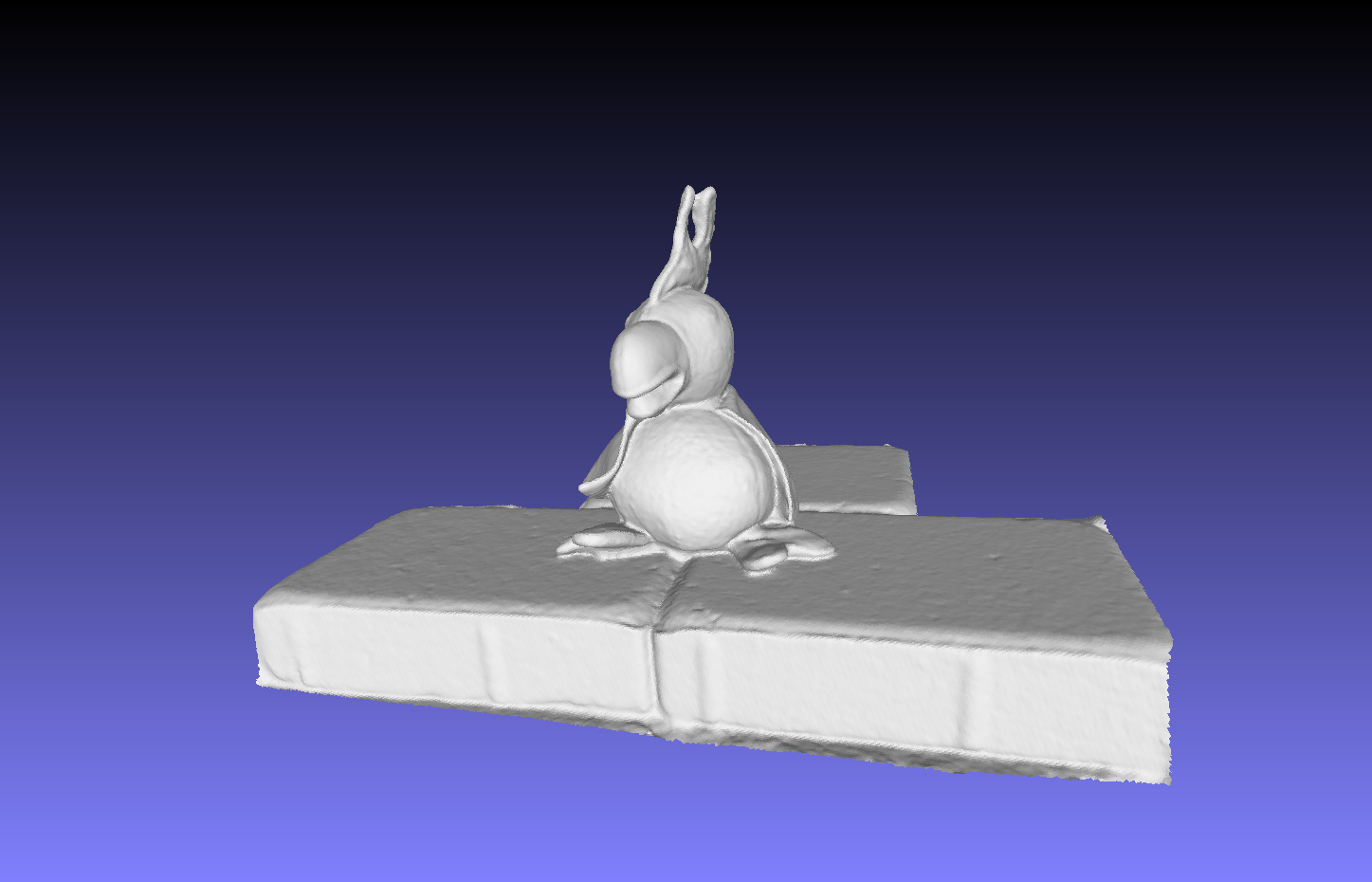}
    \end{minipage}
    
    \begin{minipage}[t]{0.22\textwidth}
        \centering
        \includegraphics[width=1\textwidth]{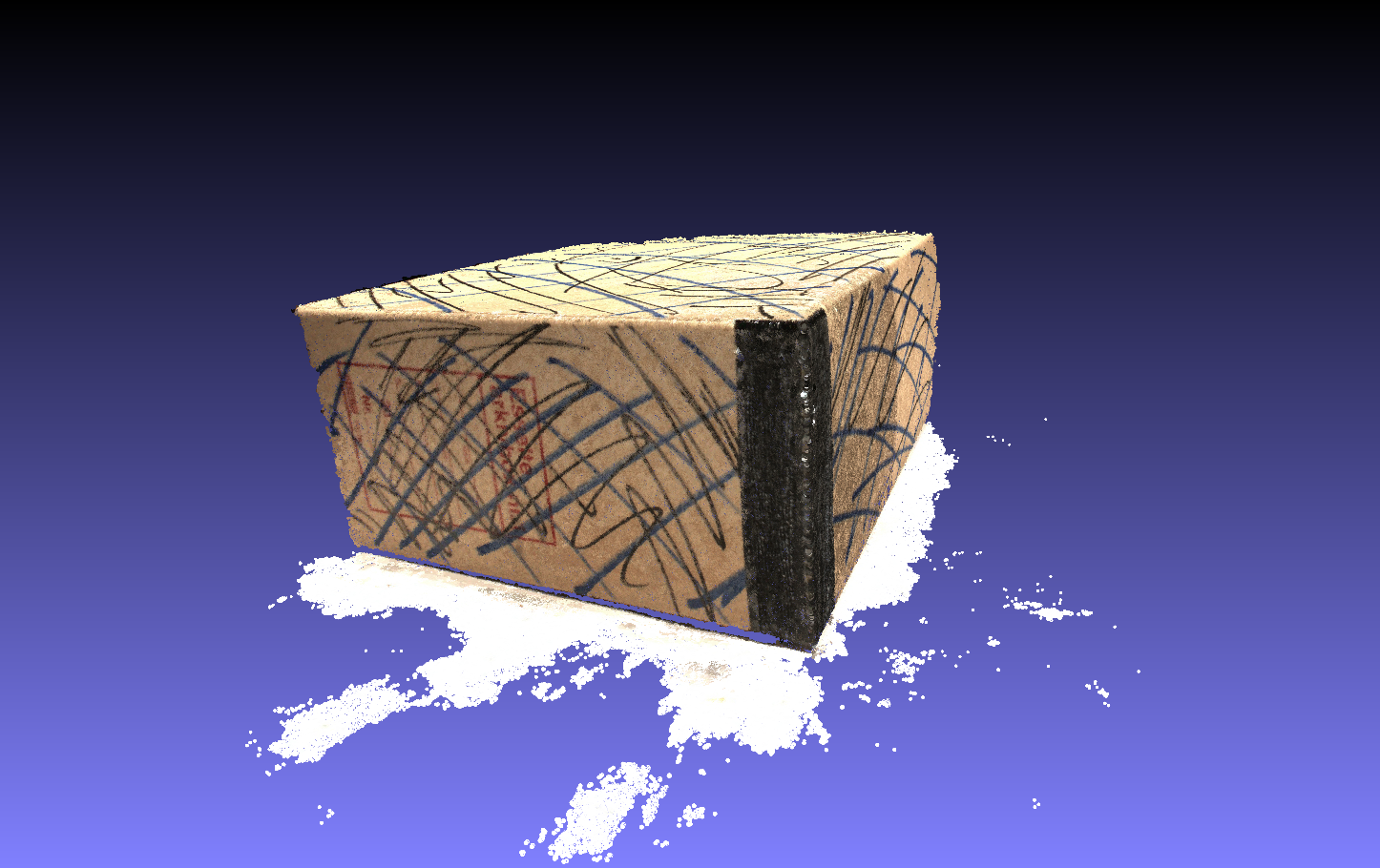}
    \end{minipage}
    \begin{minipage}[t]{0.216\textwidth}
        \centering
        \includegraphics[width=1\textwidth]{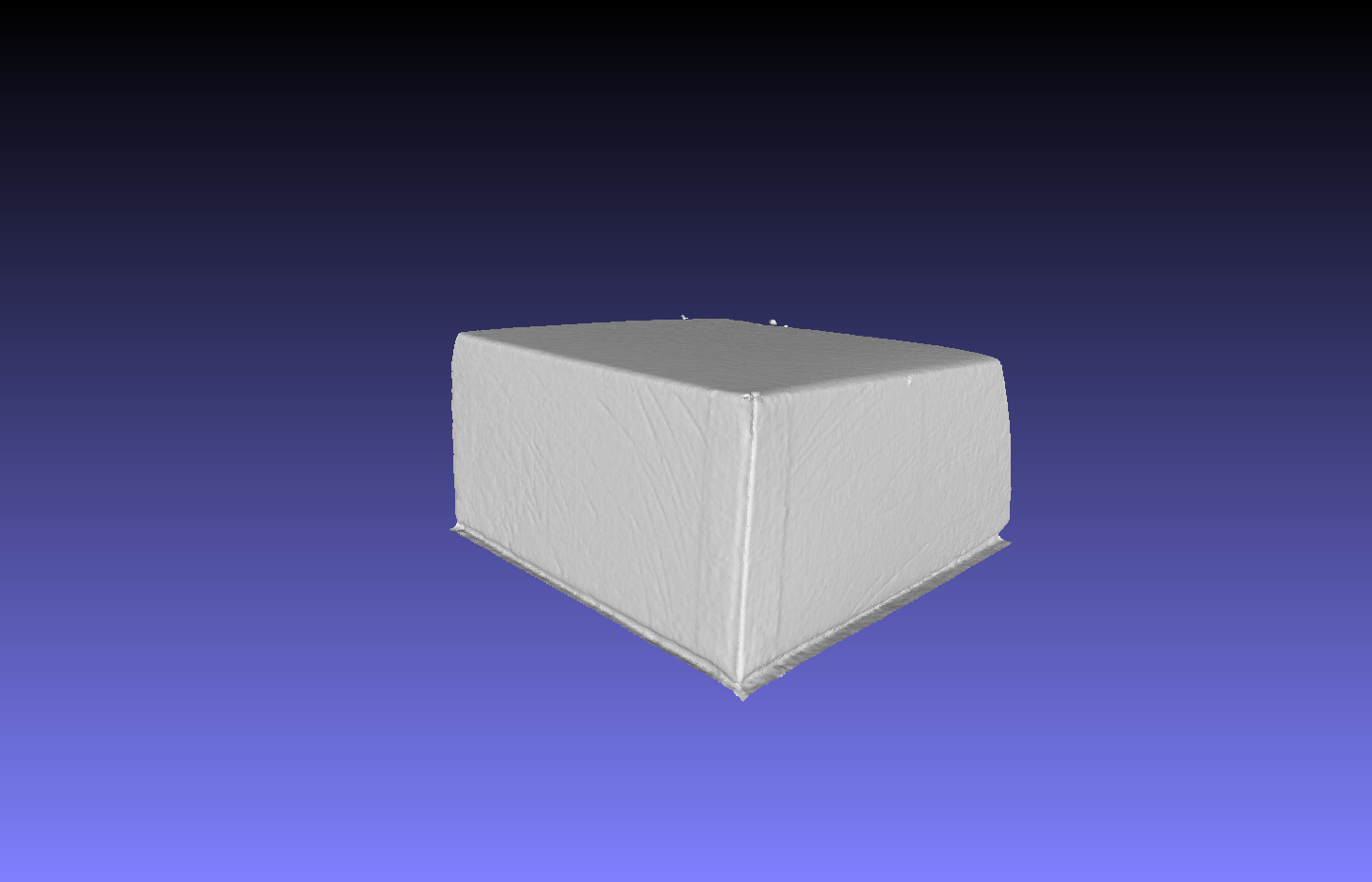}
    \end{minipage}
    \hspace{0.03\textwidth}
    \begin{minipage}[t]{0.22\textwidth}
        \centering
        \includegraphics[width=1\textwidth]{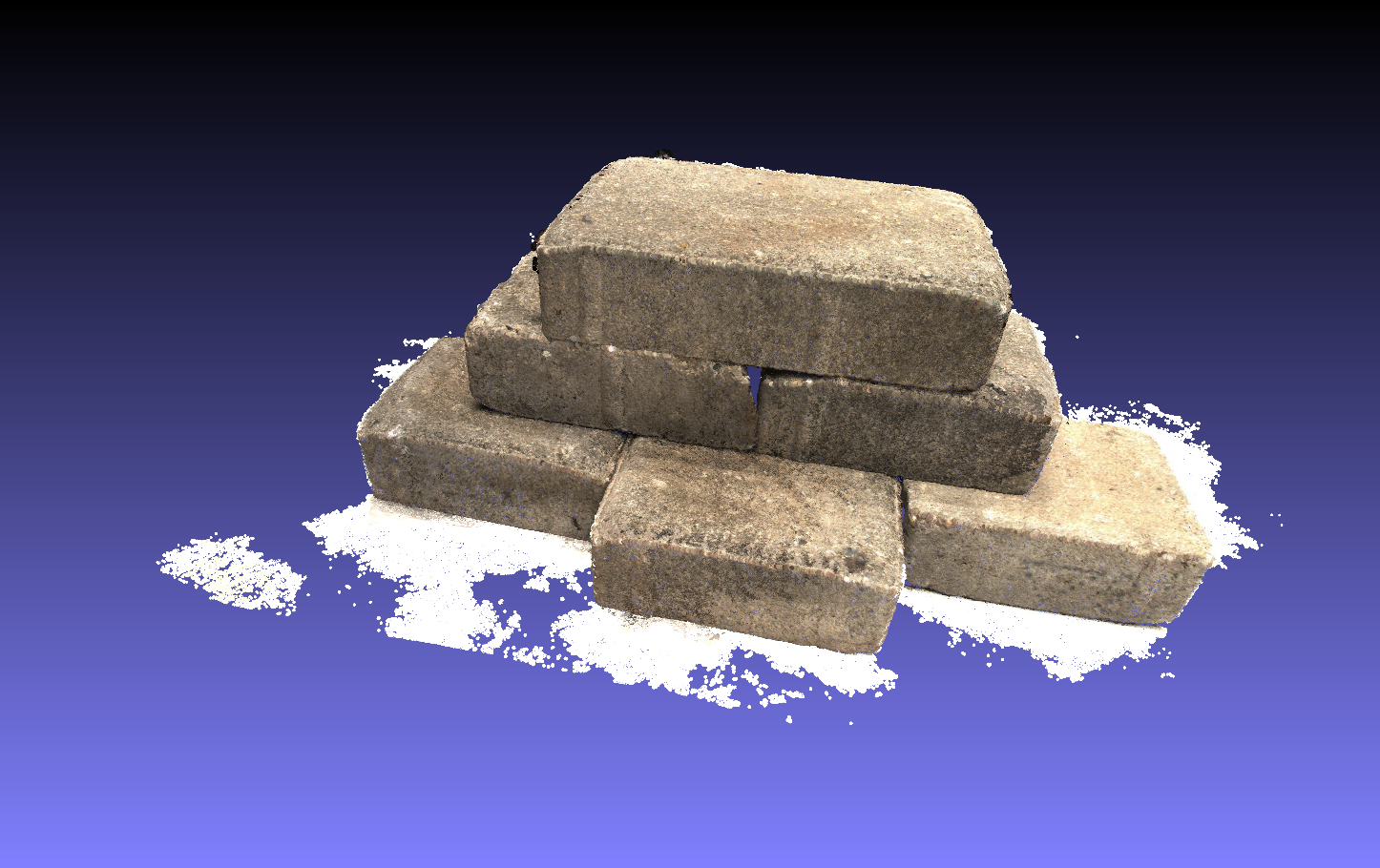}
    \end{minipage}
    \begin{minipage}[t]{0.216\textwidth}
        \centering
        \includegraphics[width=1\textwidth]{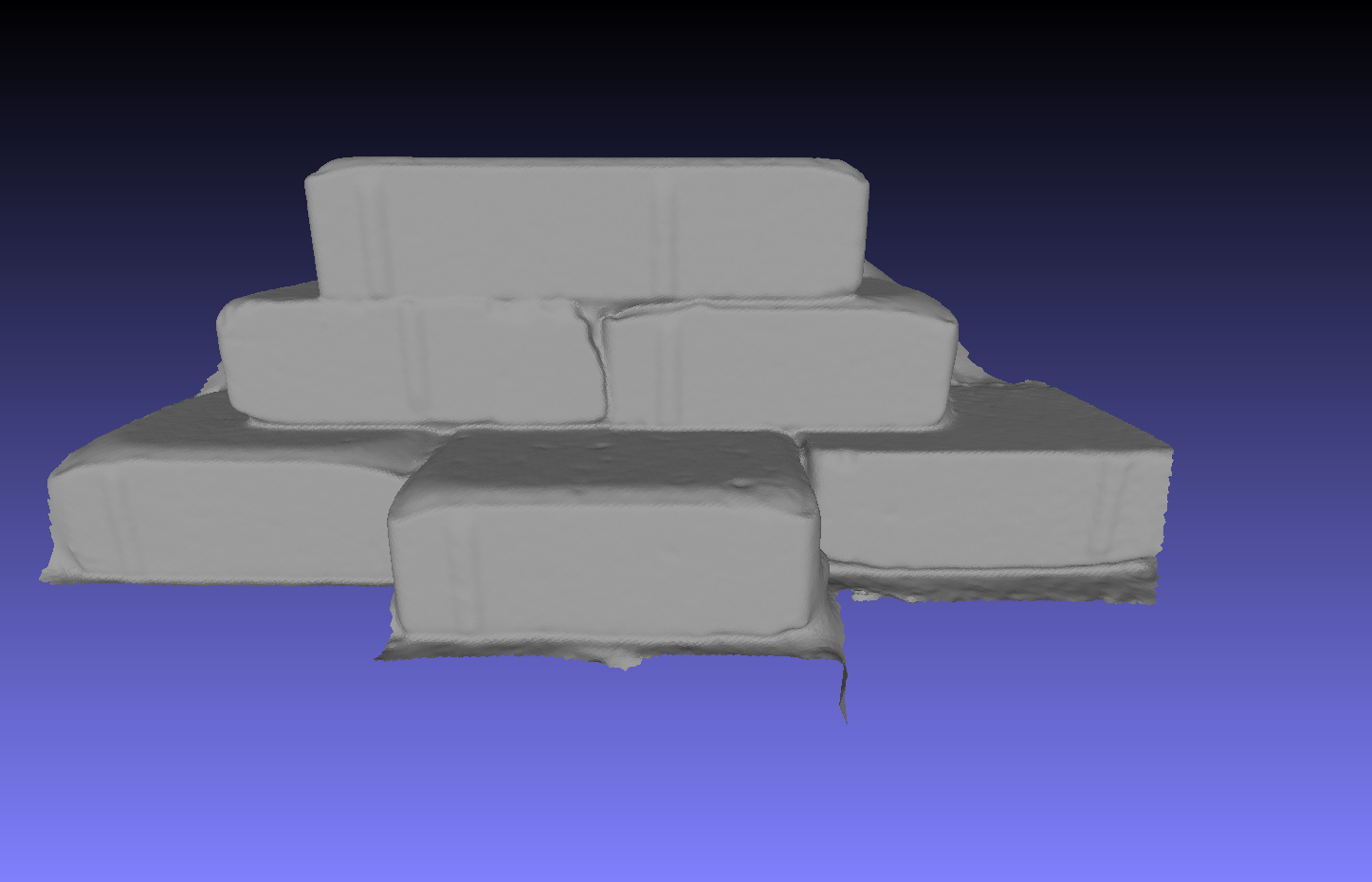}
    \end{minipage}
    
    \begin{minipage}[t]{0.22\textwidth}
        \centering
        \includegraphics[width=1\textwidth]{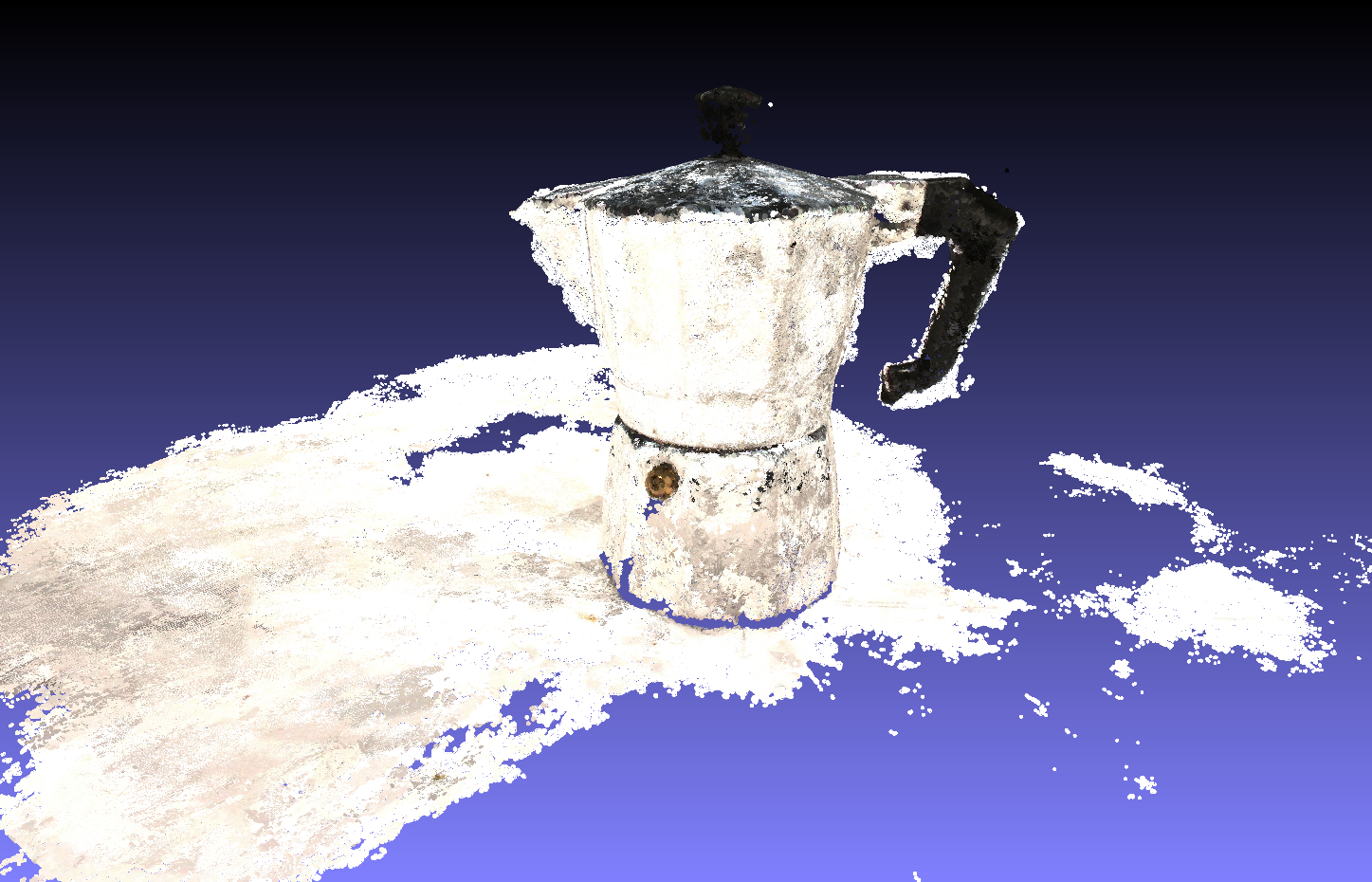}
    \end{minipage}
    \begin{minipage}[t]{0.216\textwidth}
        \centering
        \includegraphics[width=1\textwidth]{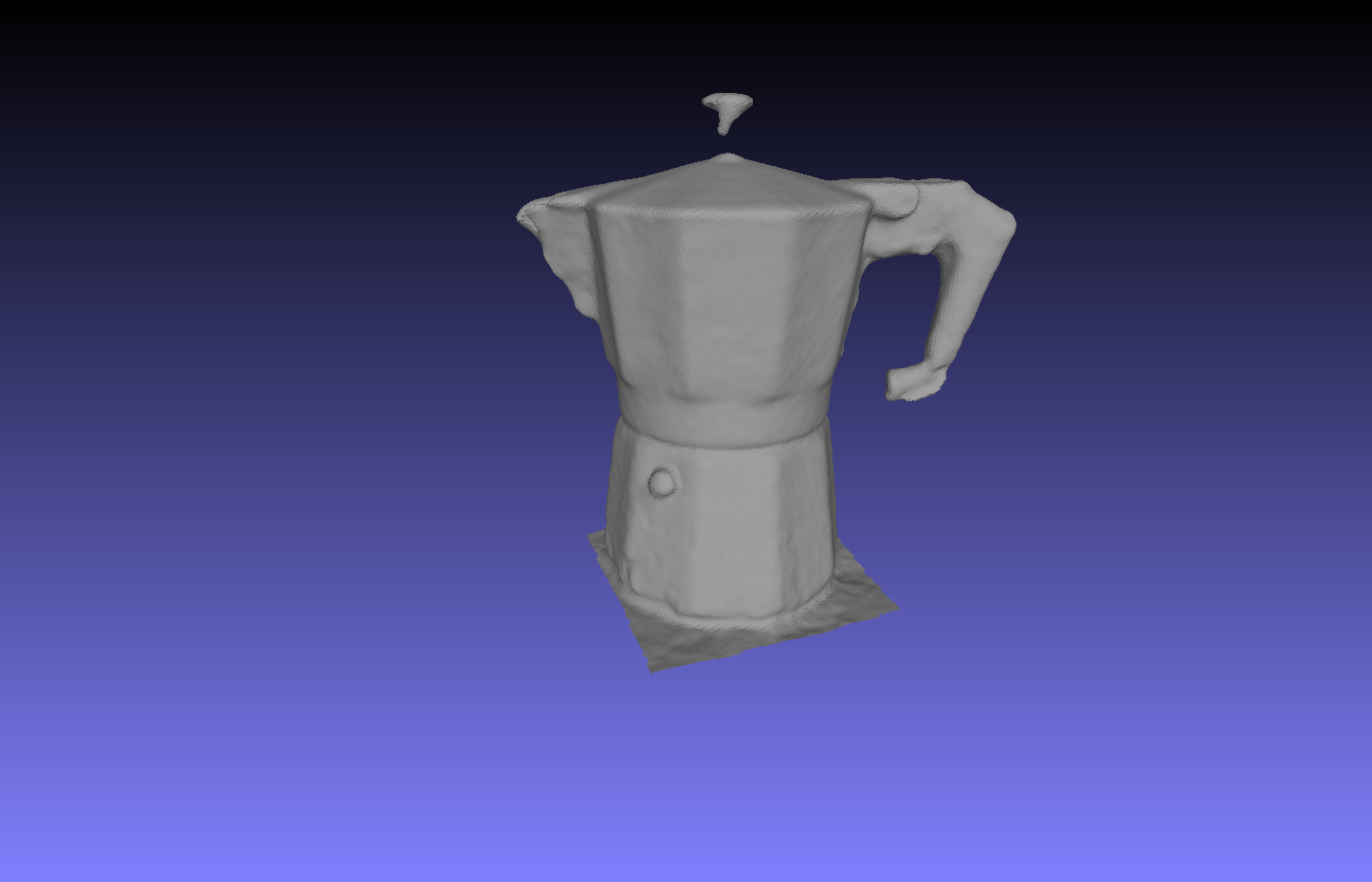}
    \end{minipage}
    \hspace{0.03\textwidth}
    \begin{minipage}[t]{0.22\textwidth}
        \centering
        \includegraphics[width=1\textwidth]{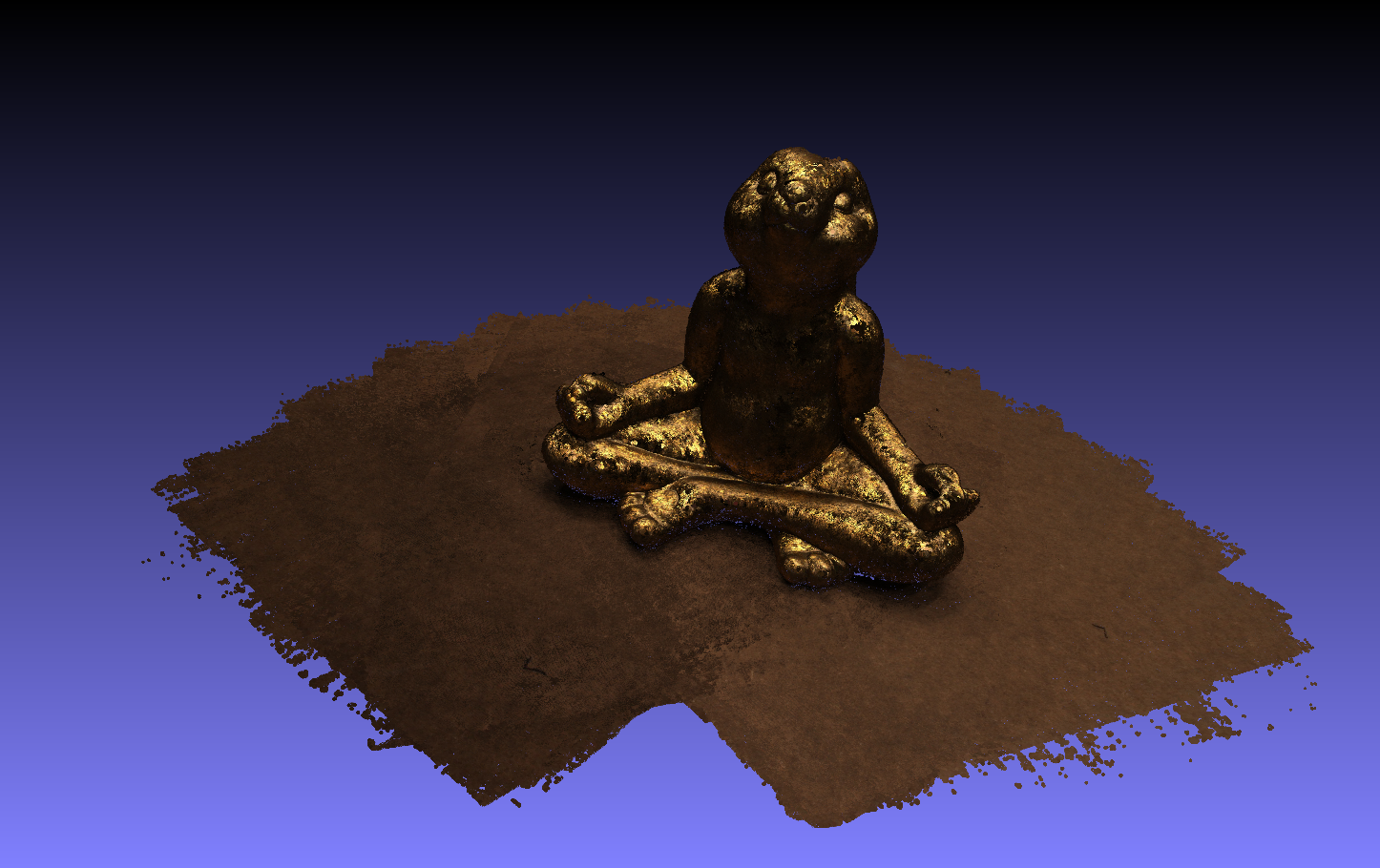}
    \end{minipage}
    \begin{minipage}[t]{0.216\textwidth}
        \centering
        \includegraphics[width=1\textwidth]{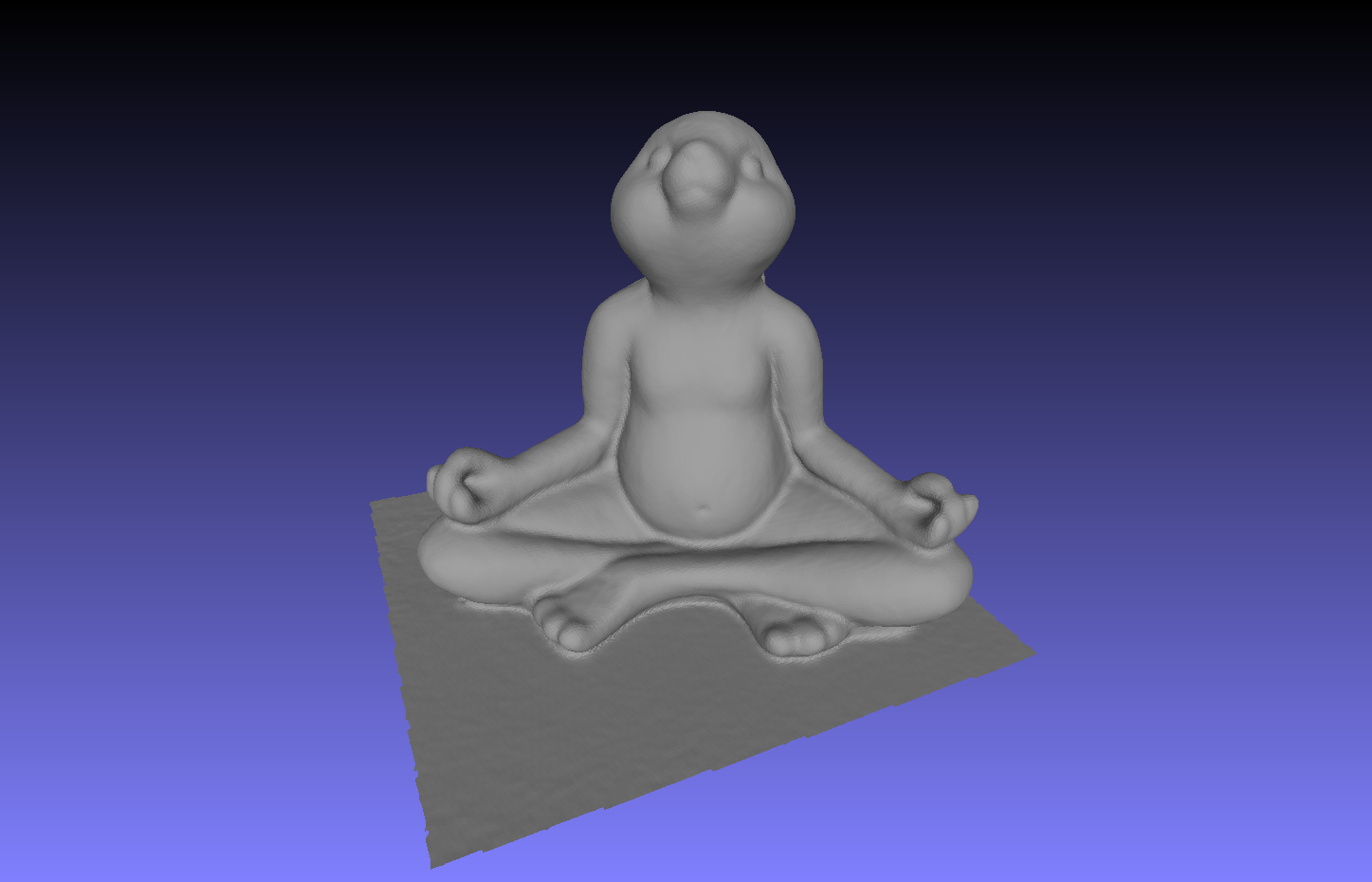}
    \end{minipage}
    
    \begin{minipage}[t]{0.22\textwidth}
        \centering
        \includegraphics[width=1\textwidth]{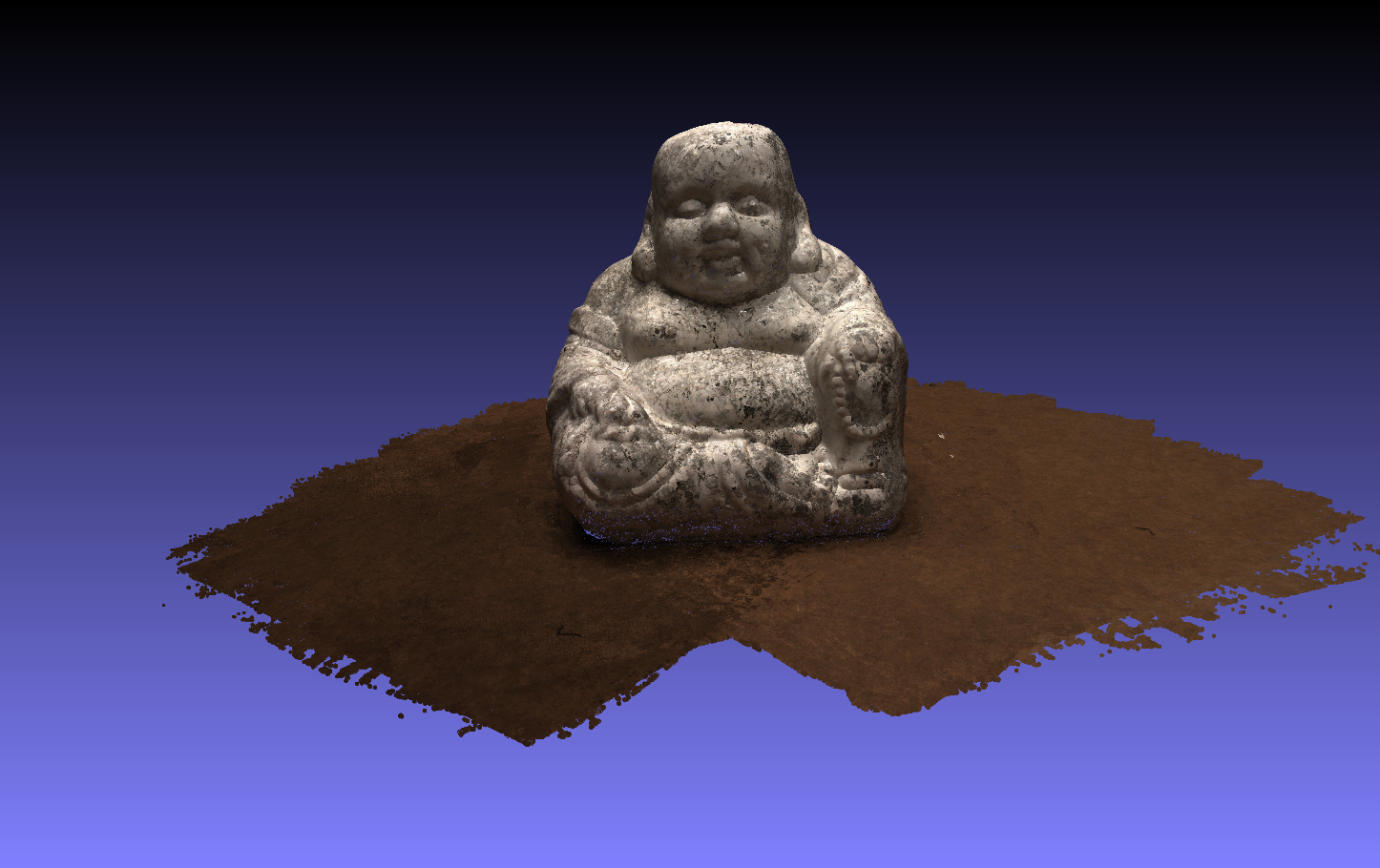}
    \end{minipage}
    \begin{minipage}[t]{0.216\textwidth}
        \centering
        \includegraphics[width=1\textwidth]{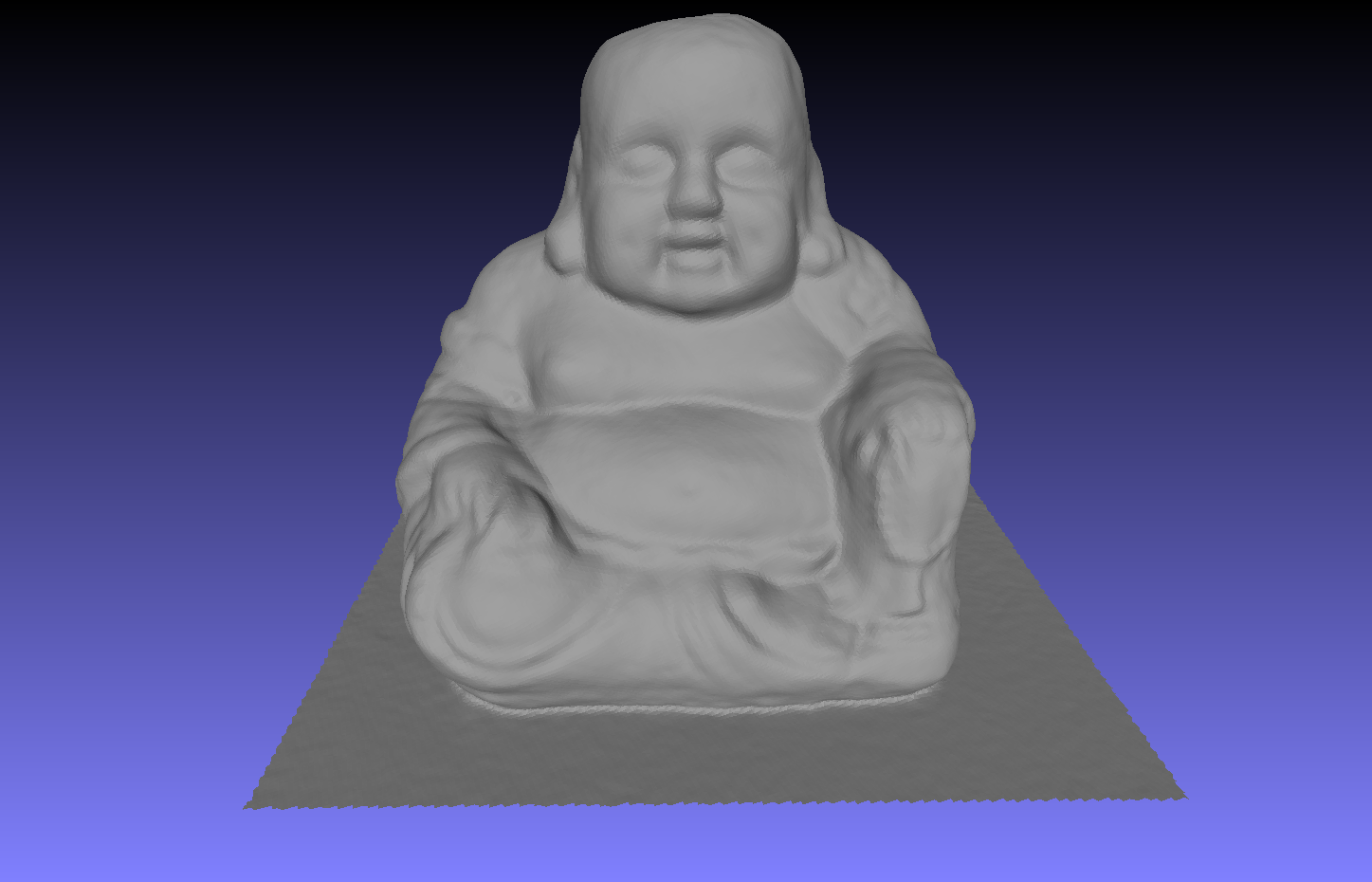}
    \end{minipage}
    \hspace{0.03\textwidth}
    \begin{minipage}[t]{0.22\textwidth}
        \centering
        \includegraphics[width=1\textwidth]{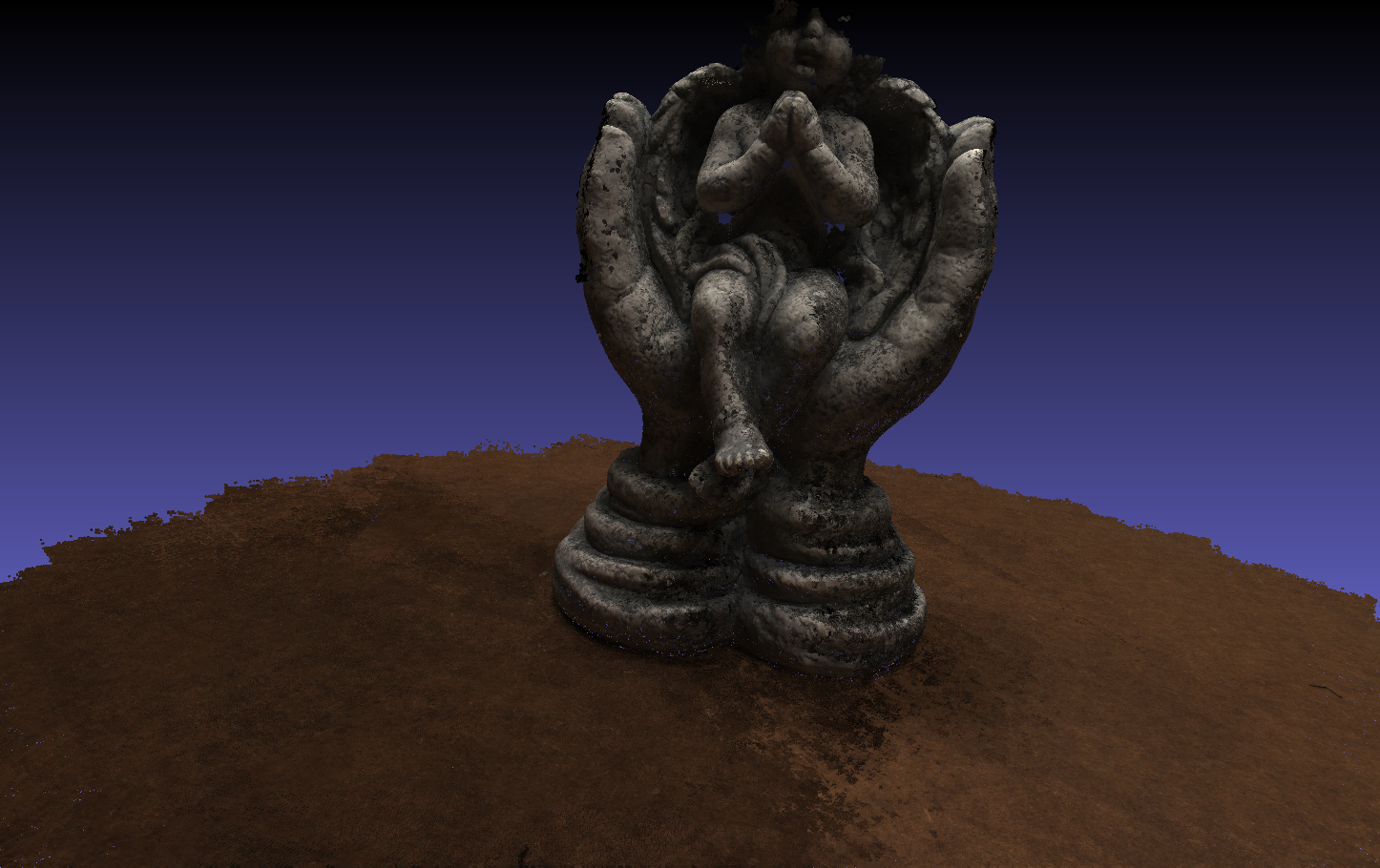}
    \end{minipage}
    \begin{minipage}[t]{0.216\textwidth}
        \centering
        \includegraphics[width=1\textwidth]{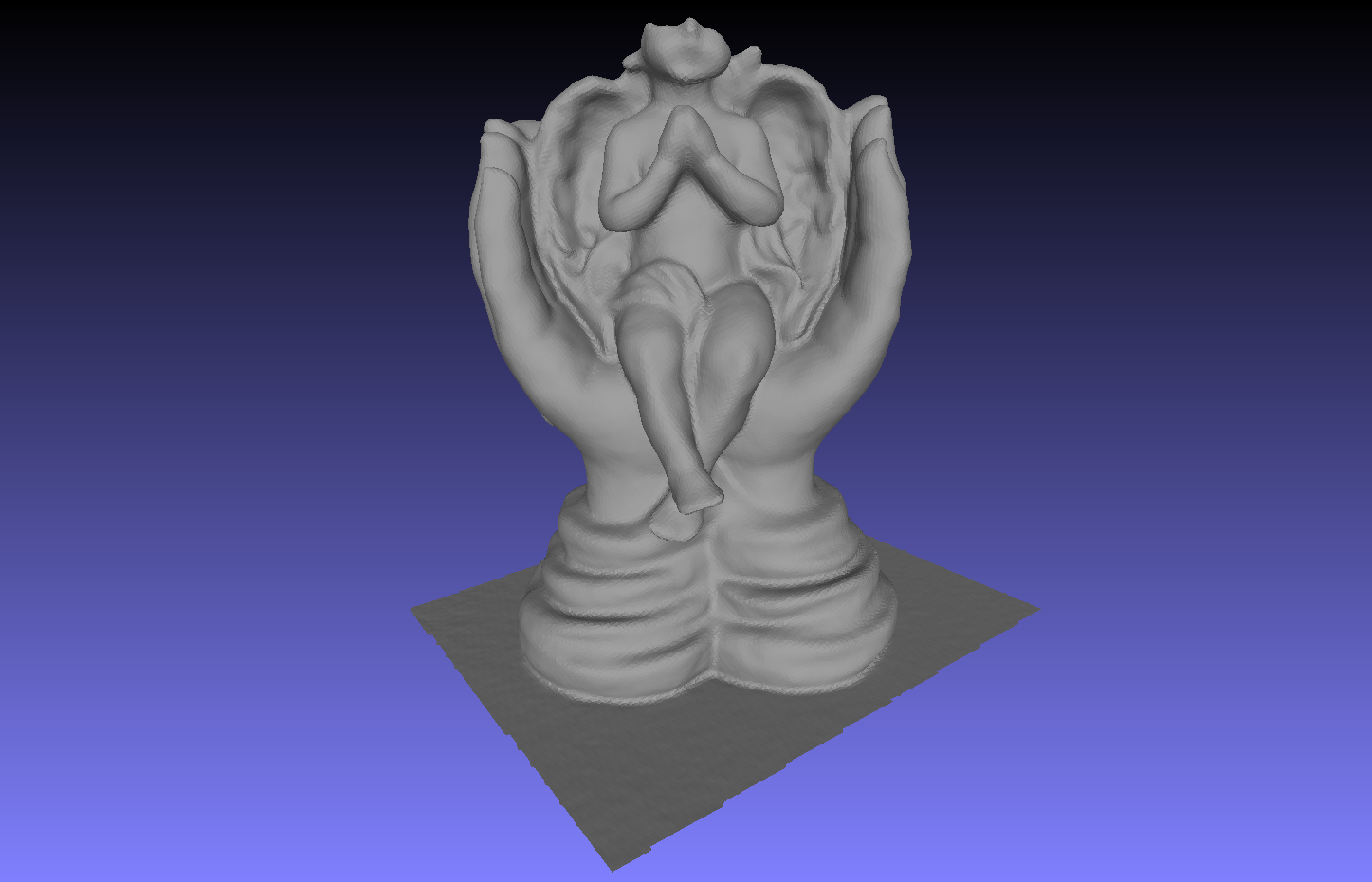}
    \end{minipage}
    
    \begin{minipage}[t]{0.22\textwidth}
        \centering
        \includegraphics[width=1\textwidth]{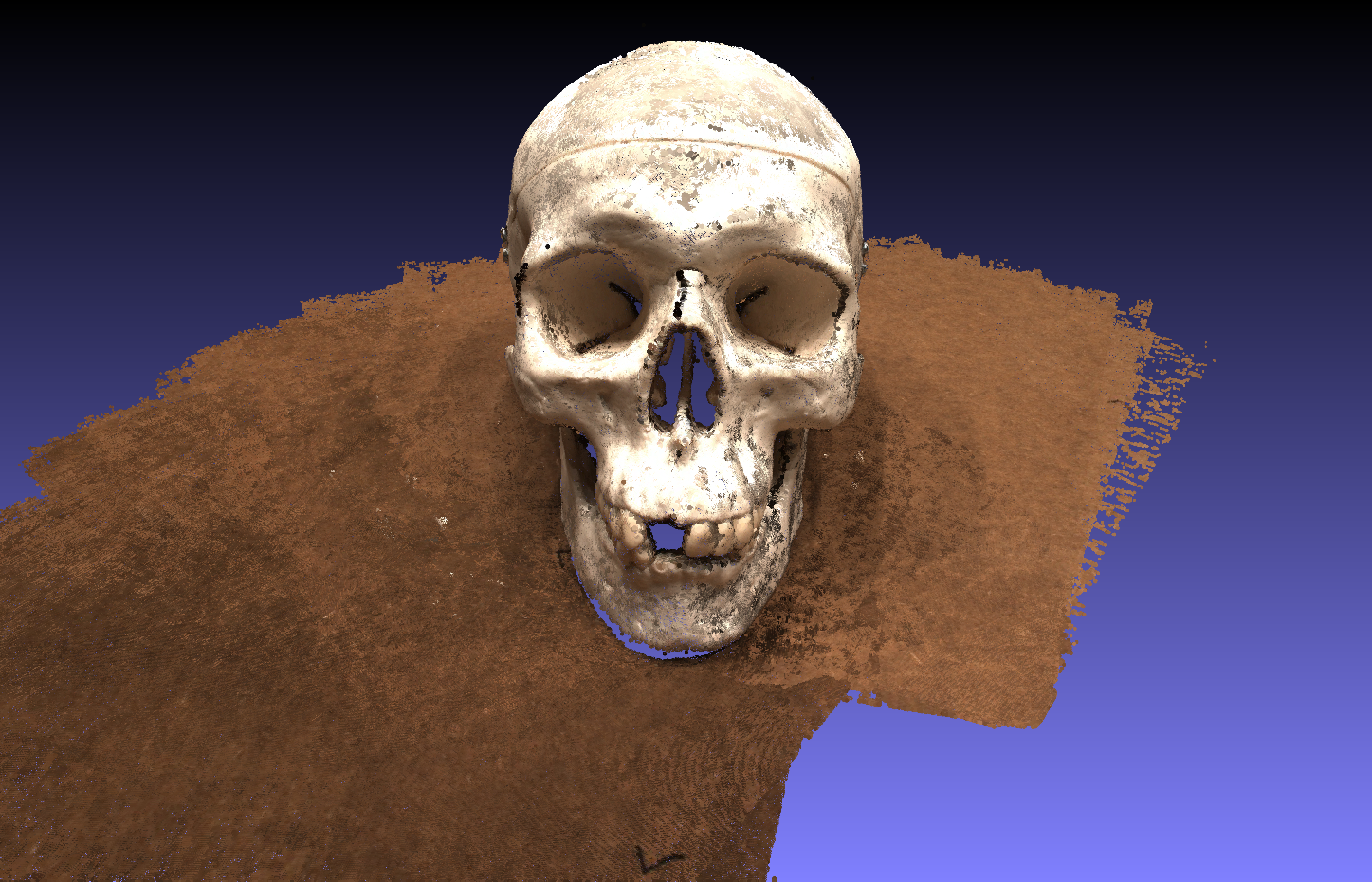}
    \end{minipage}
    \begin{minipage}[t]{0.216\textwidth}
        \centering
        \includegraphics[width=1\textwidth]{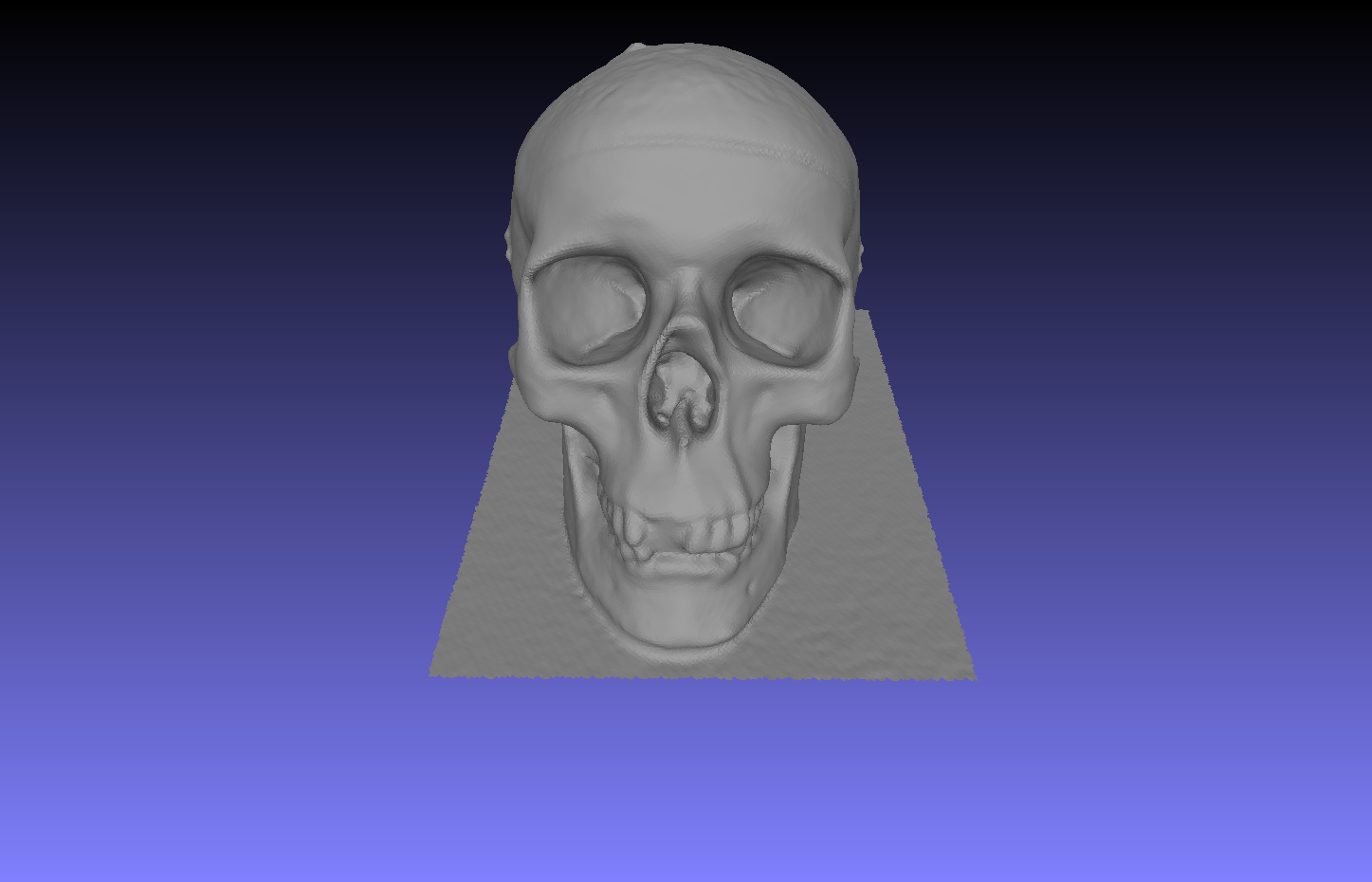}
    \end{minipage}
    \hspace{0.03\textwidth}
    \begin{minipage}[t]{0.22\textwidth}
        \centering
        \includegraphics[width=1\textwidth]{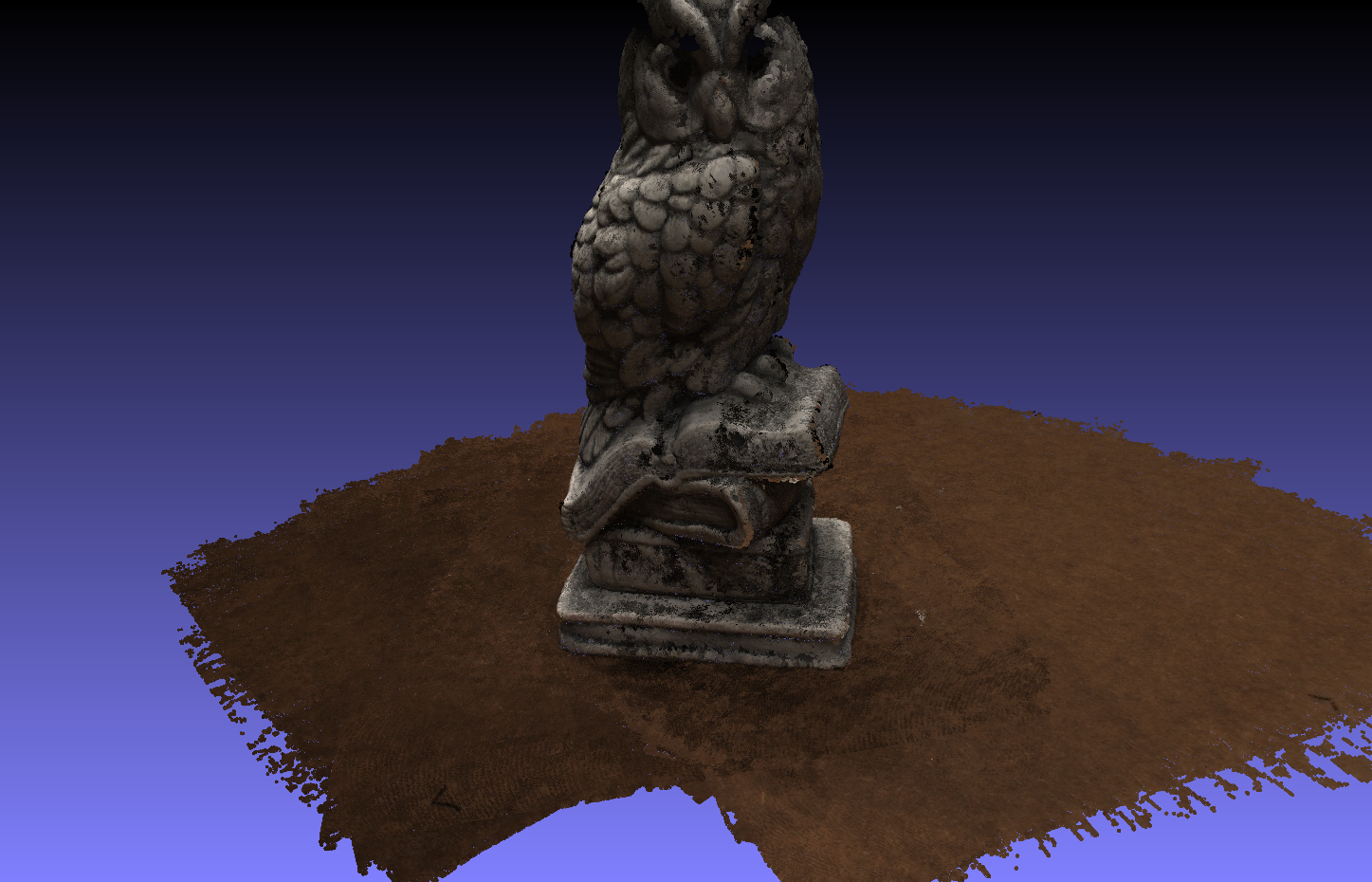}
    \end{minipage}
    \begin{minipage}[t]{0.216\textwidth}
        \centering
        \includegraphics[width=1\textwidth]{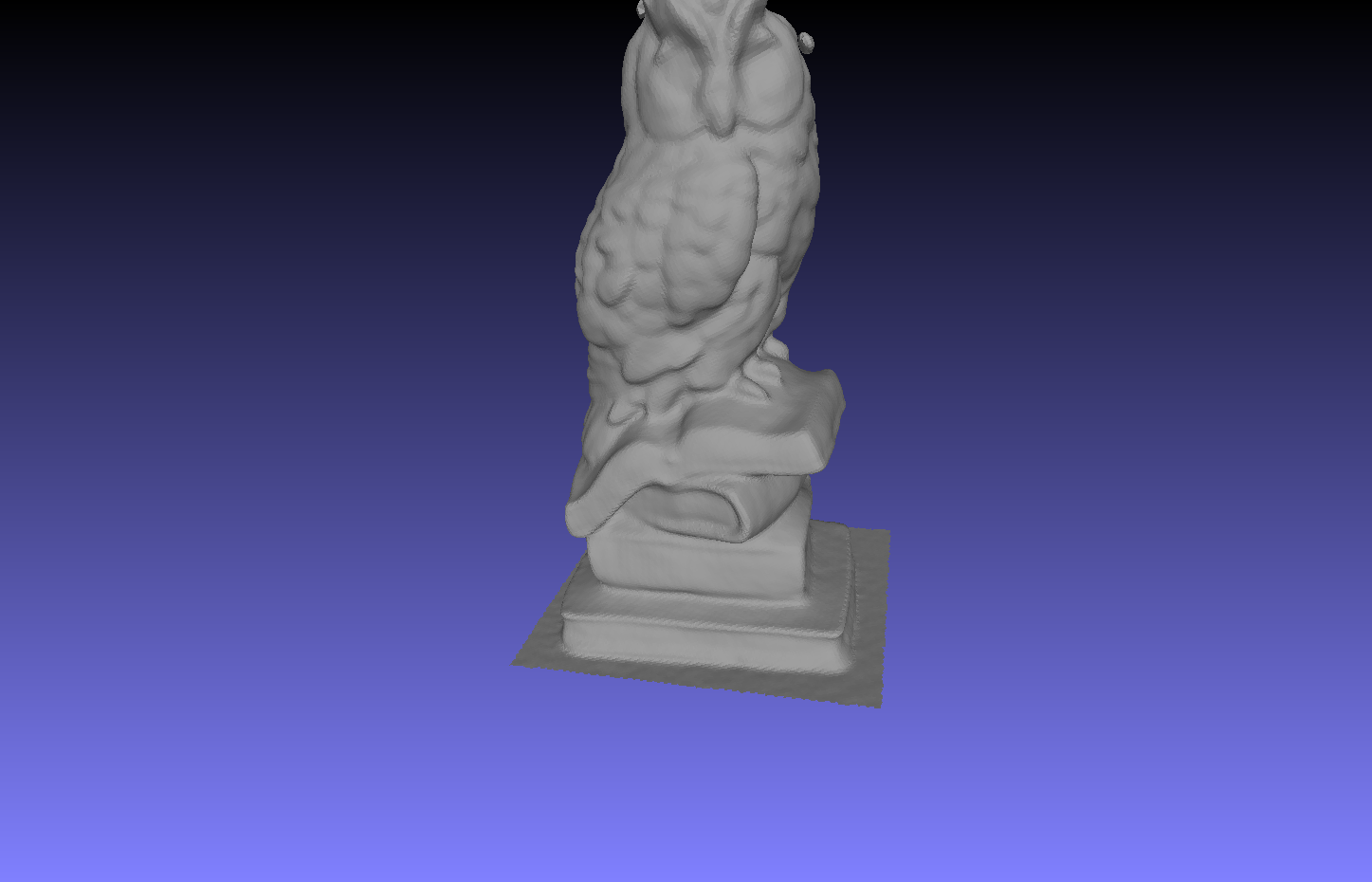}
    \end{minipage}
    \caption{{\bf More qualitative results on DTU.} }
    \label{fig:DTU}
    \vspace{-0.2cm}
    
\end{figure*}

\begin{figure*}[htb]
  \begin{center}
     \includegraphics[width=0.27\linewidth]{figs/Family00.png}
     \includegraphics[width=0.45\linewidth]{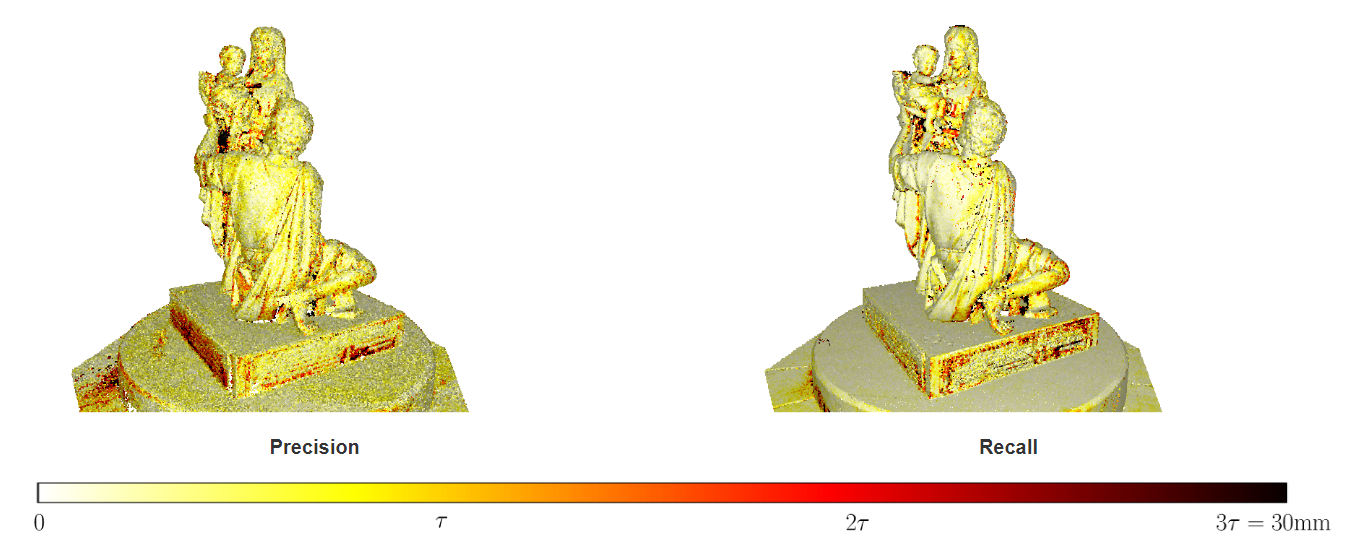}
     
     \includegraphics[width=0.27\linewidth]{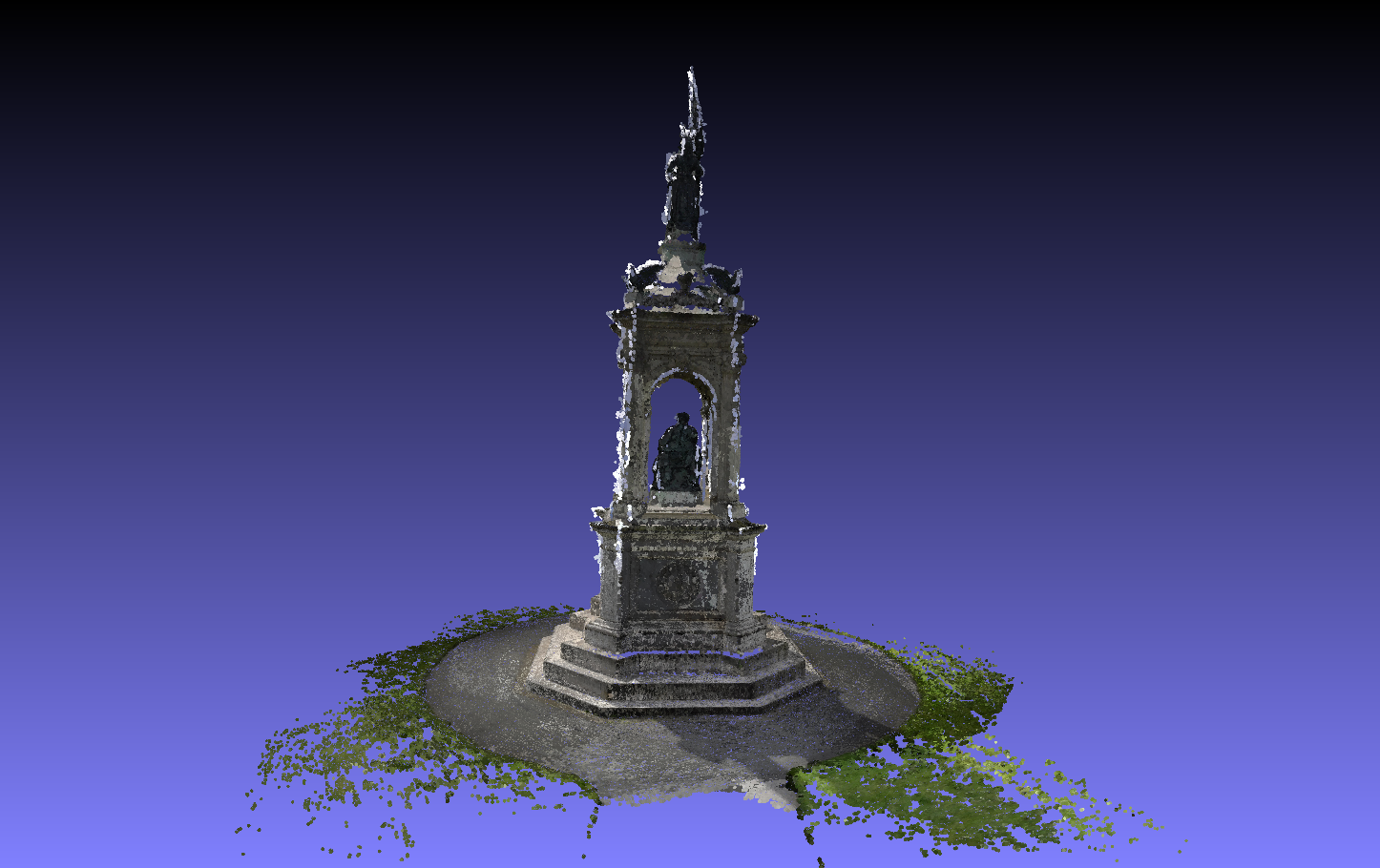}
     \includegraphics[width=0.45\linewidth]{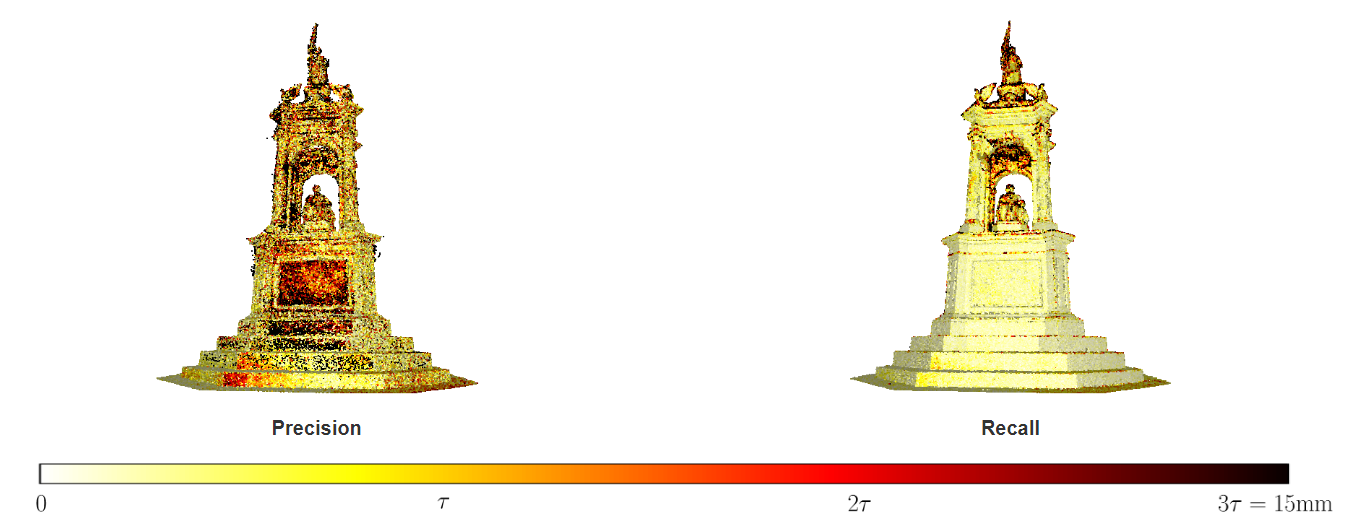}
     
     \includegraphics[width=0.27\linewidth]{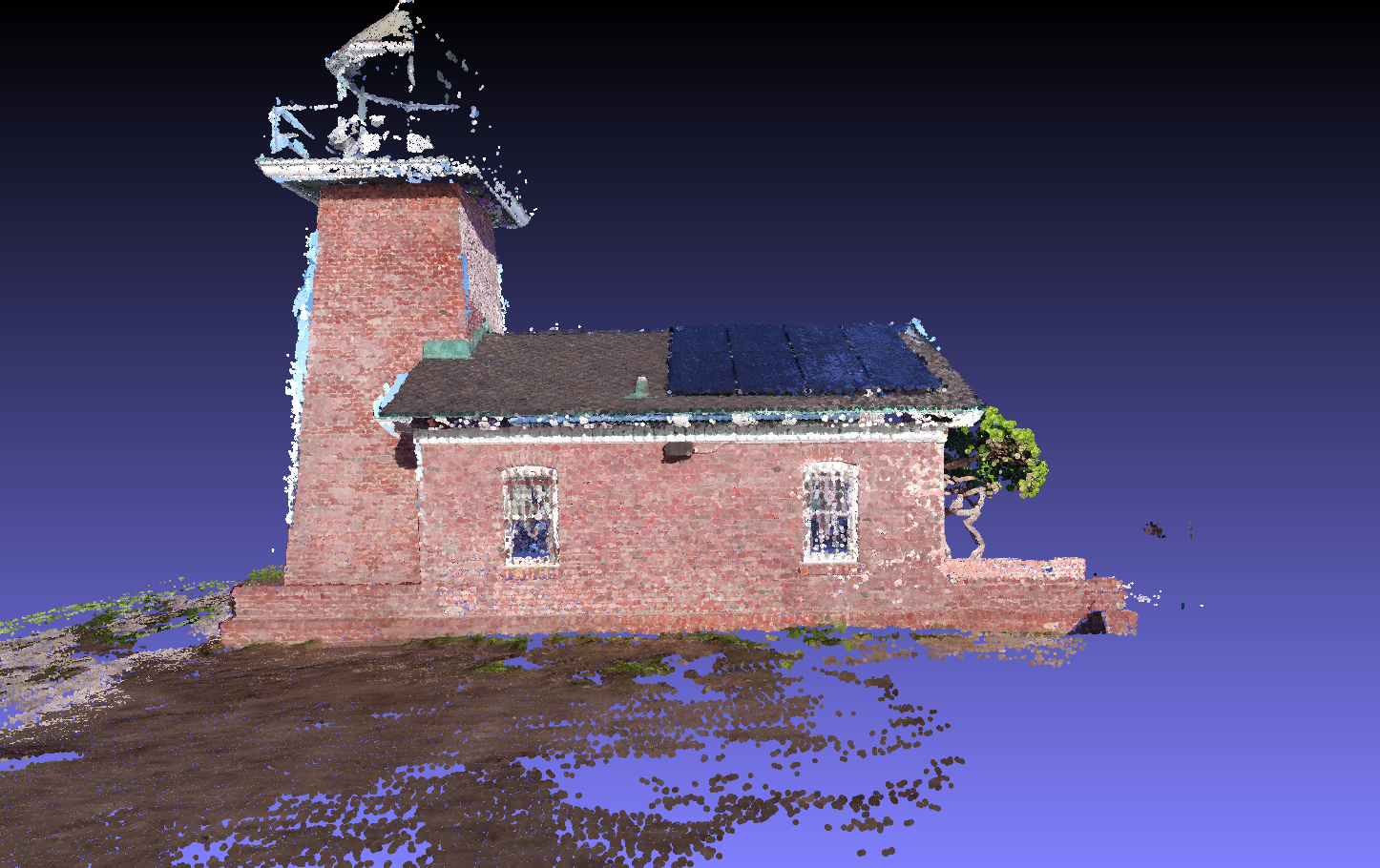}
     \includegraphics[width=0.45\linewidth]{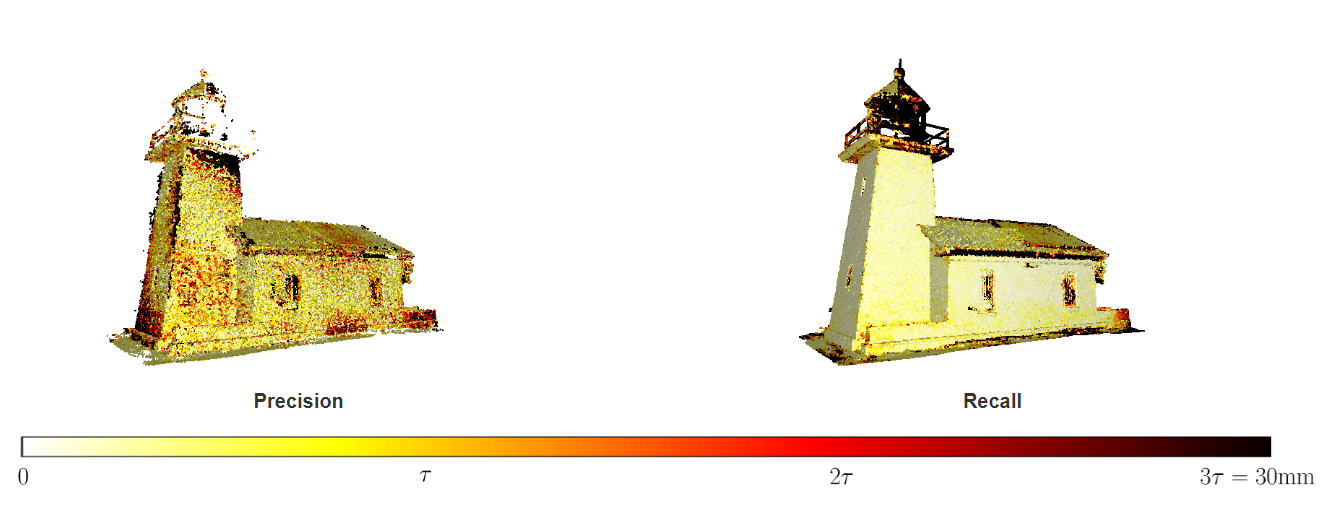}
     
     \includegraphics[width=0.27\linewidth]{figs/M6000.png}
     \includegraphics[width=0.45\linewidth]{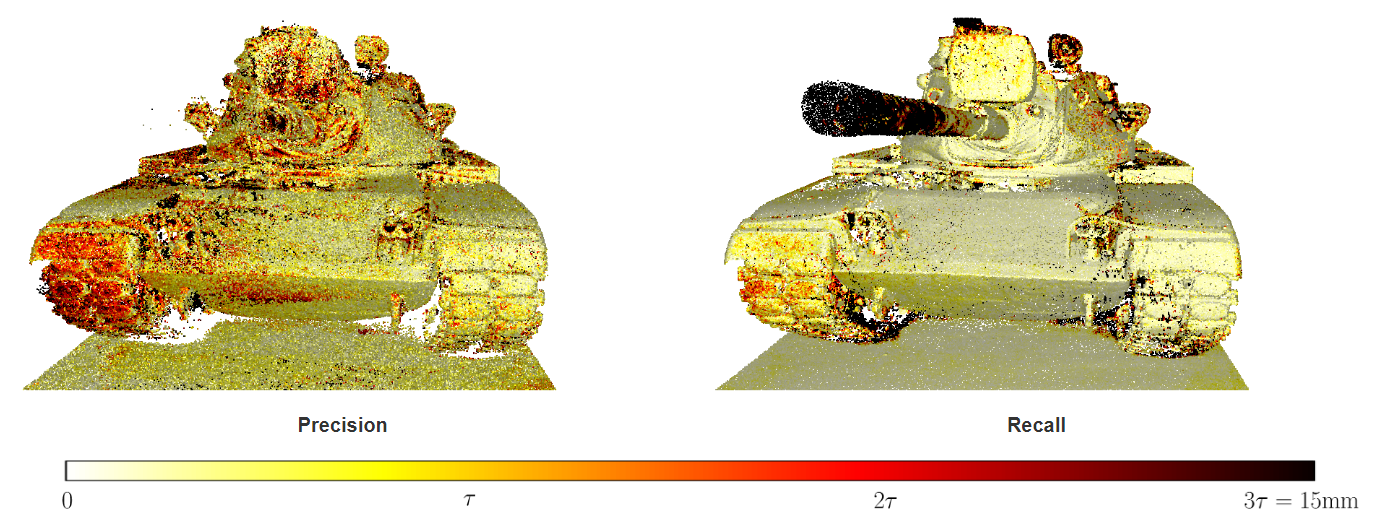}
     
     \includegraphics[width=0.27\linewidth]{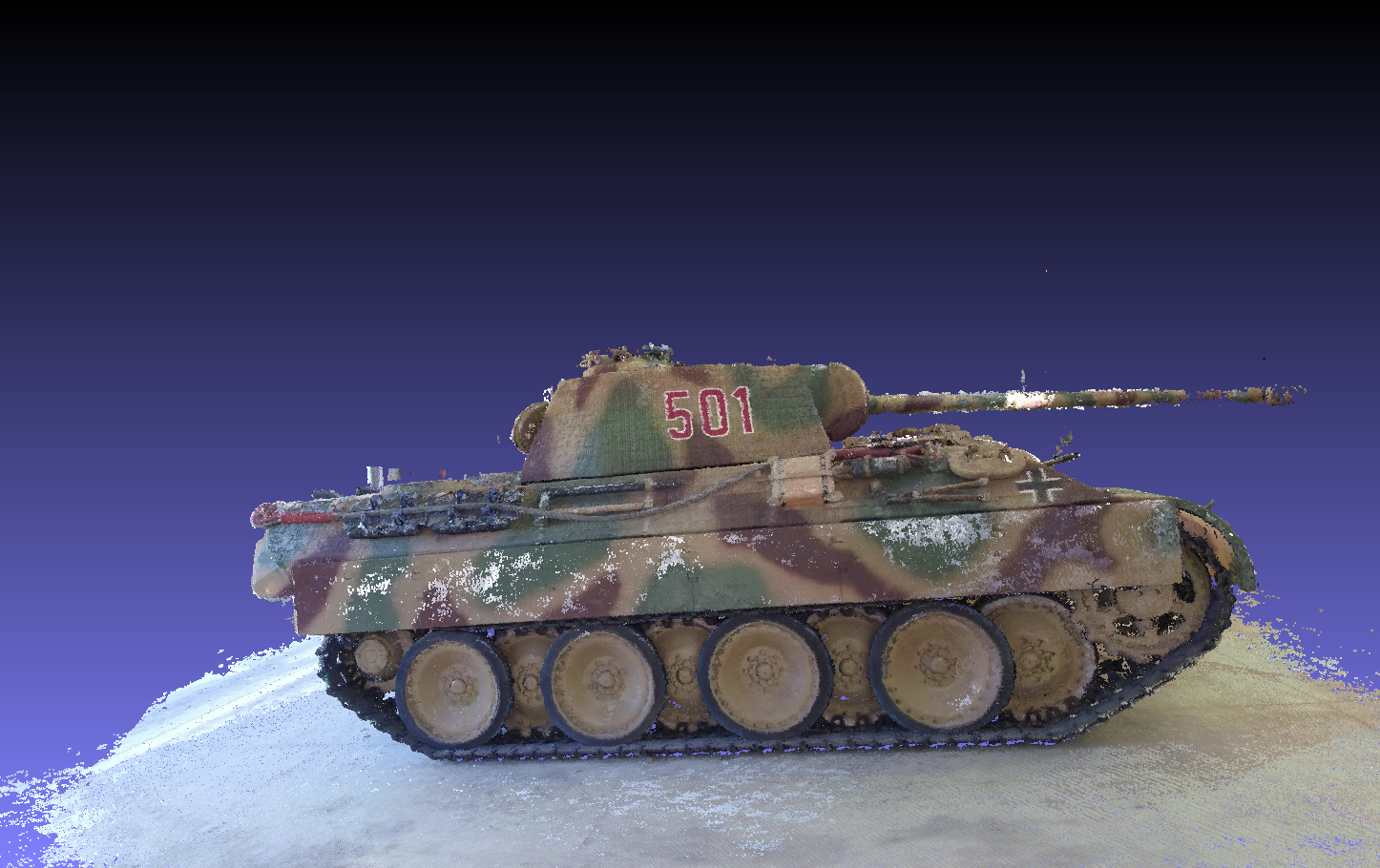}
     \includegraphics[width=0.45\linewidth]{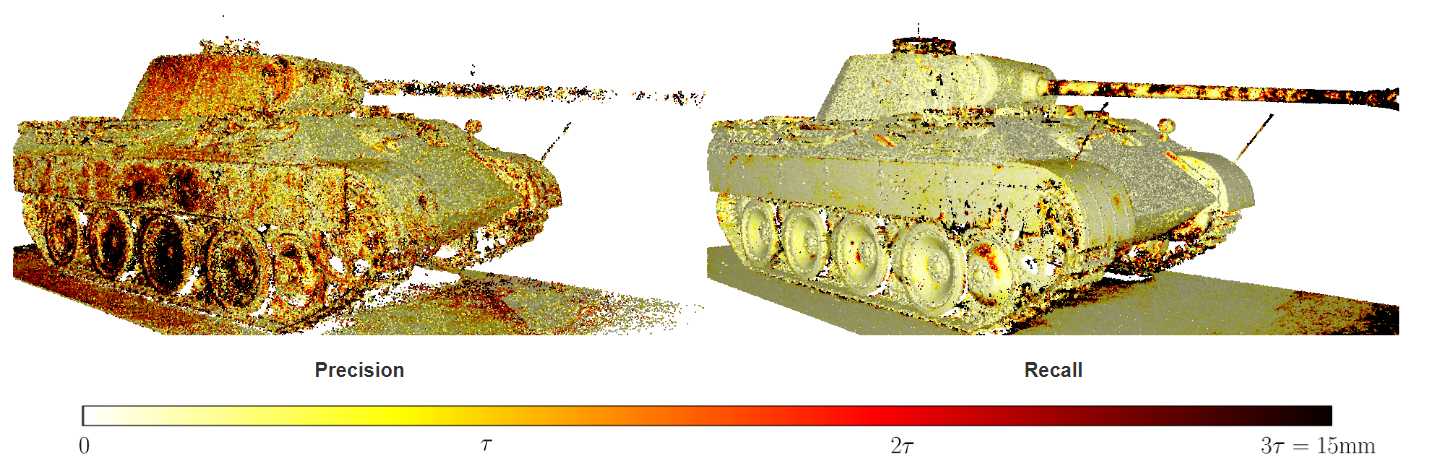}
     
     \includegraphics[width=0.27\linewidth]{figs/playground00.png}
     \includegraphics[width=0.45\linewidth]{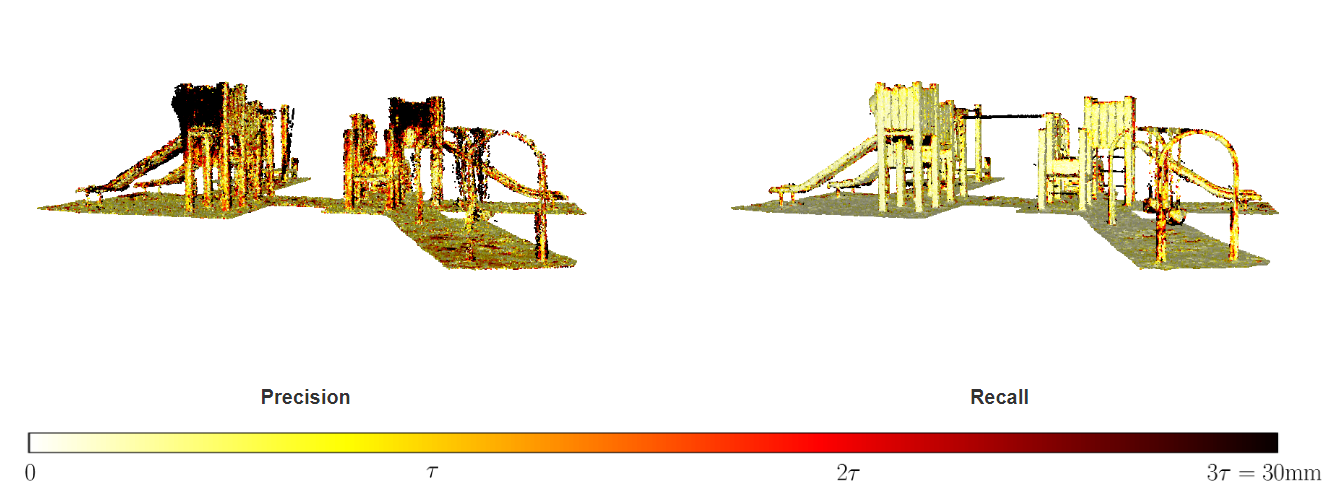}
     
     \includegraphics[width=0.27\linewidth]{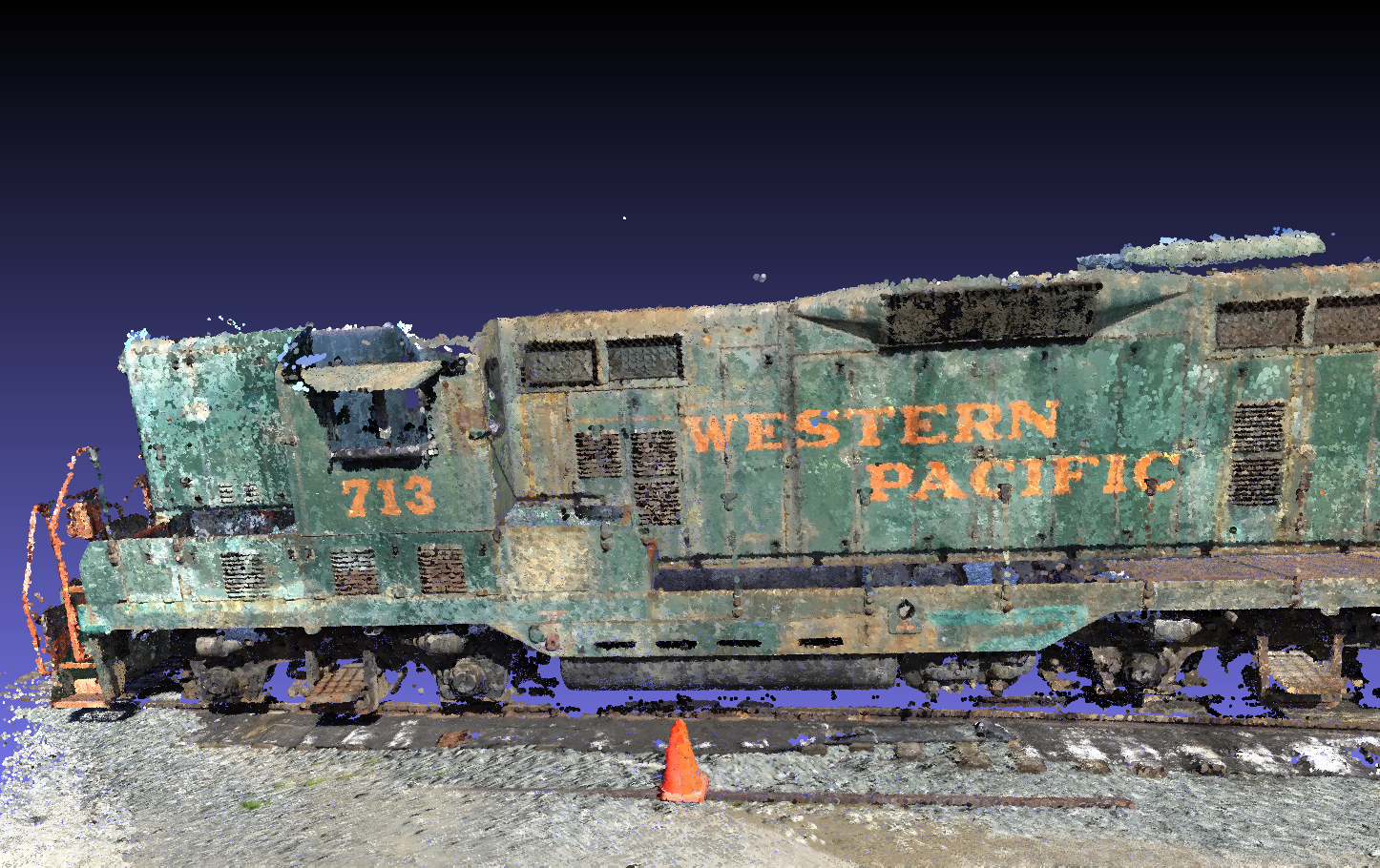}
     \includegraphics[width=0.45\linewidth]{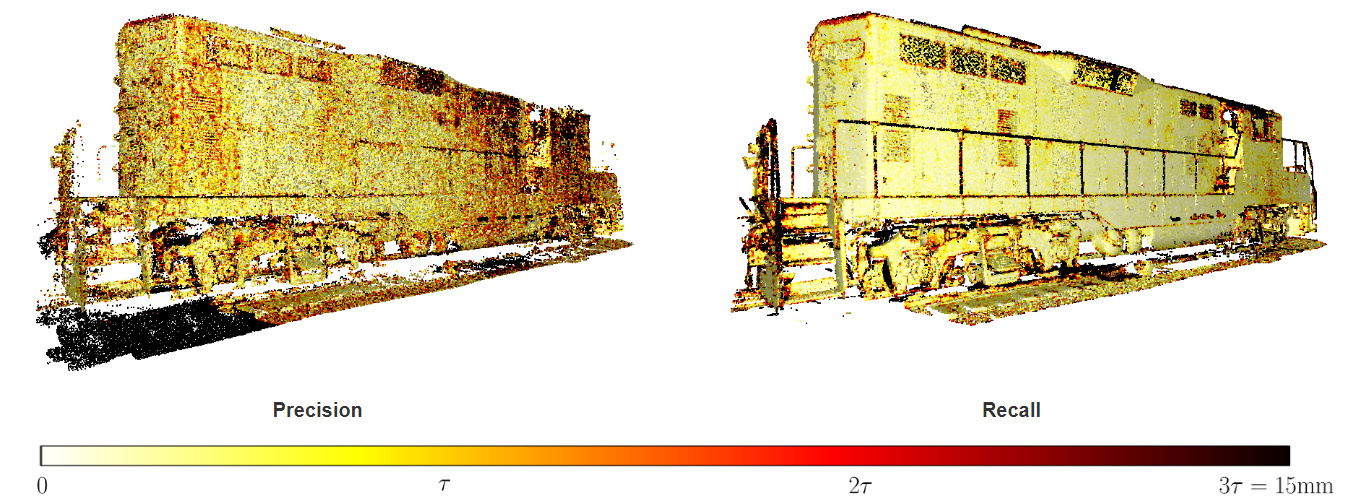}
     
     \caption{{\bf More qualitative results on Tanks and Temples Intermediate. } }
     \label{fig:TI}
     \vspace{-0.2cm}
  \end{center}
\end{figure*}

\begin{figure*}[htb]
  \begin{center}
     \includegraphics[width=0.28\linewidth]{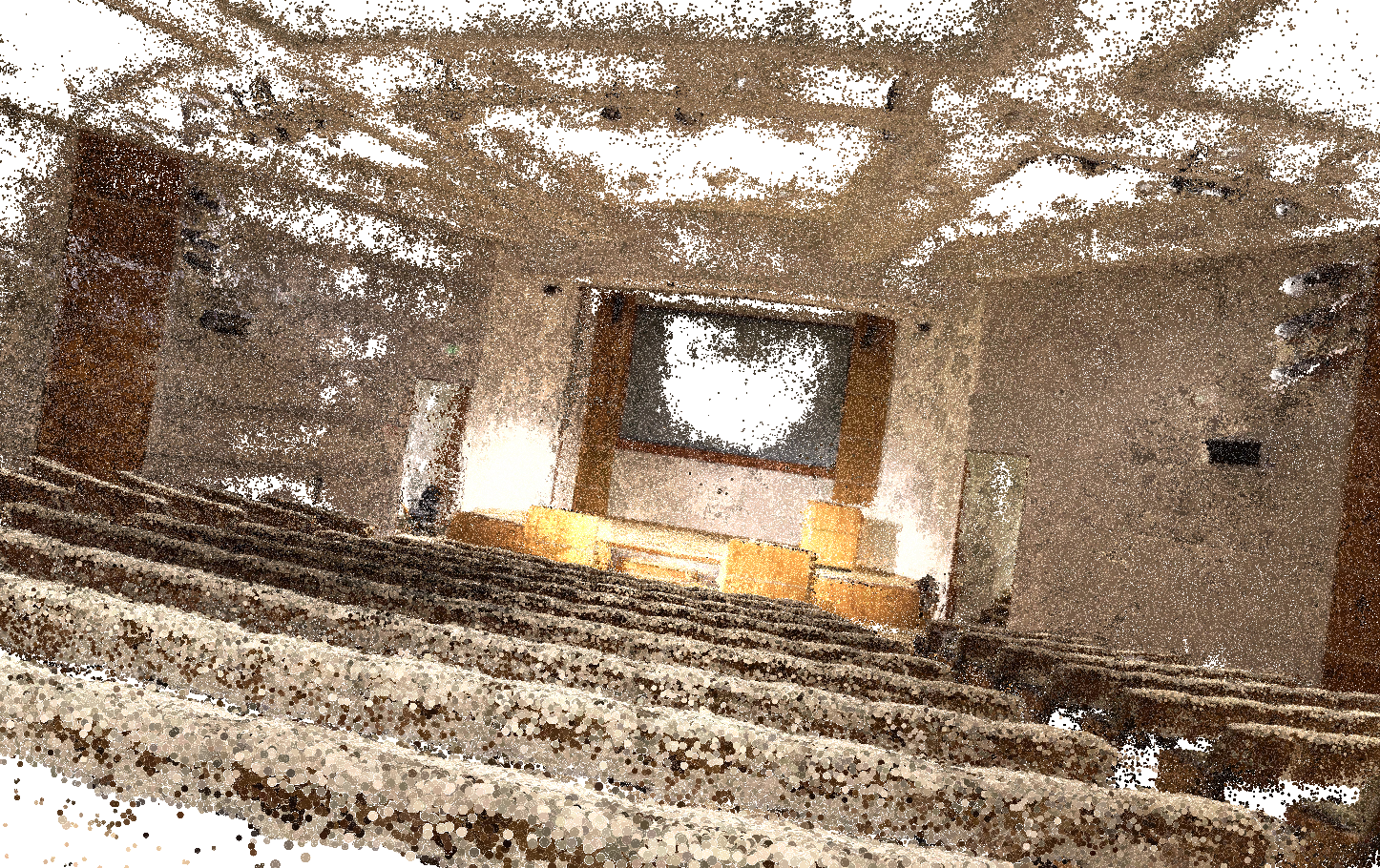}
     \includegraphics[width=0.46\linewidth]{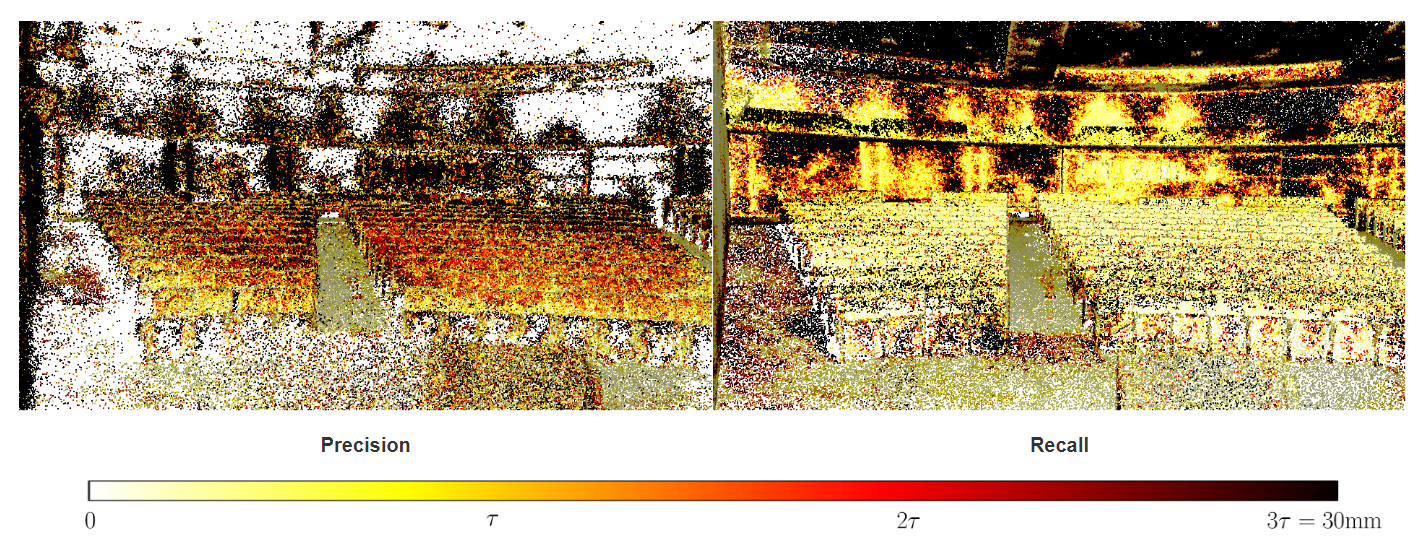}
     
     \includegraphics[width=0.28\linewidth]{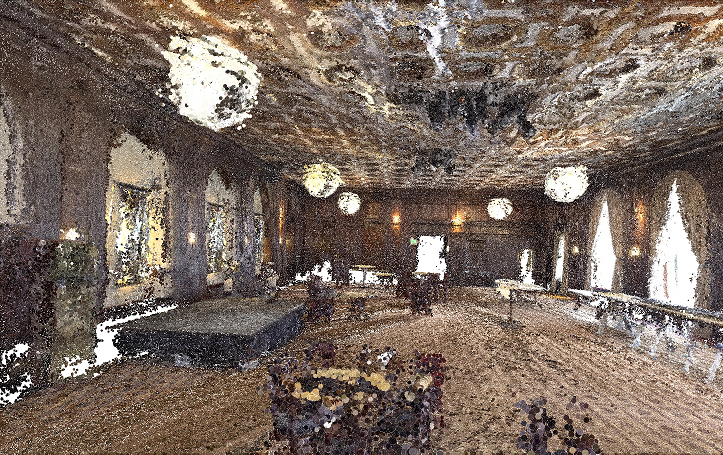}
     \includegraphics[width=0.46\linewidth]{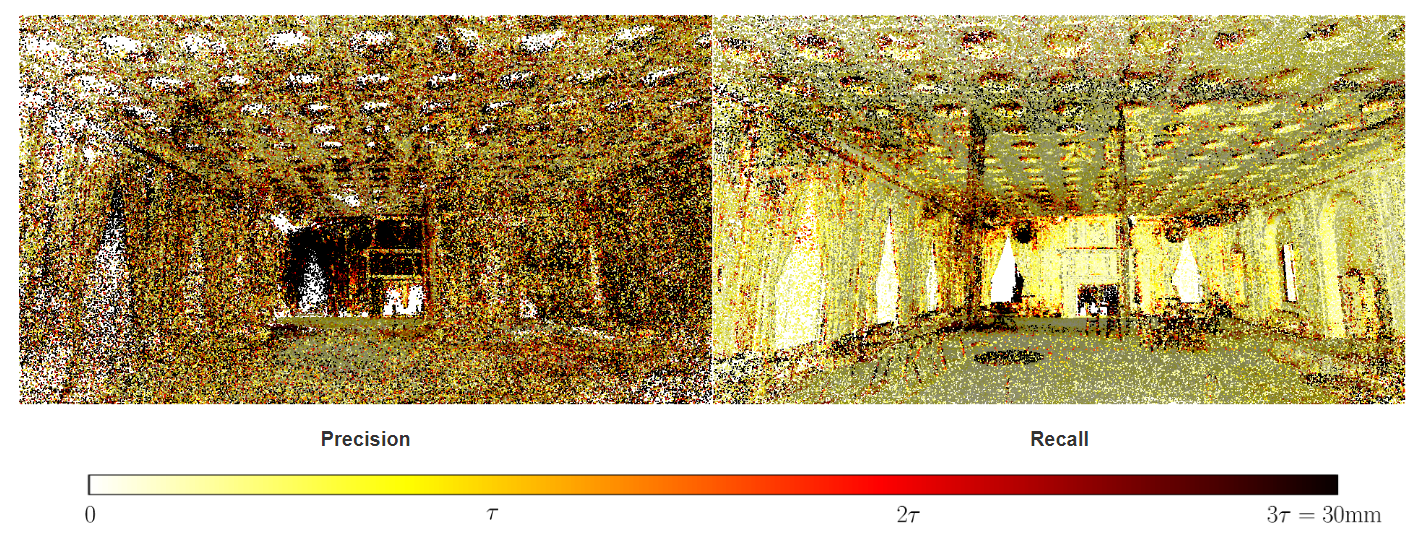}
     
     \includegraphics[width=0.28\linewidth]{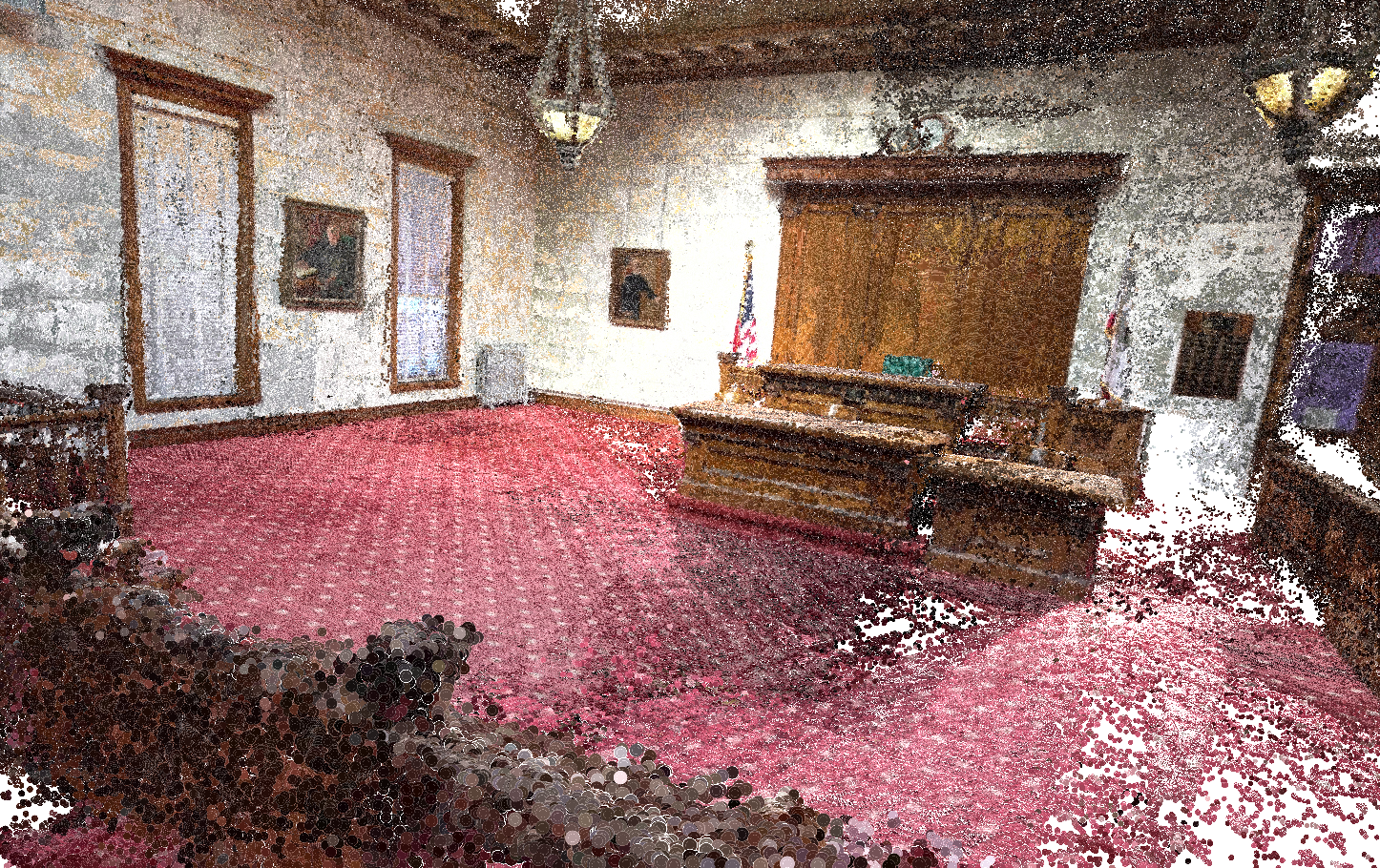}
     \includegraphics[width=0.46\linewidth]{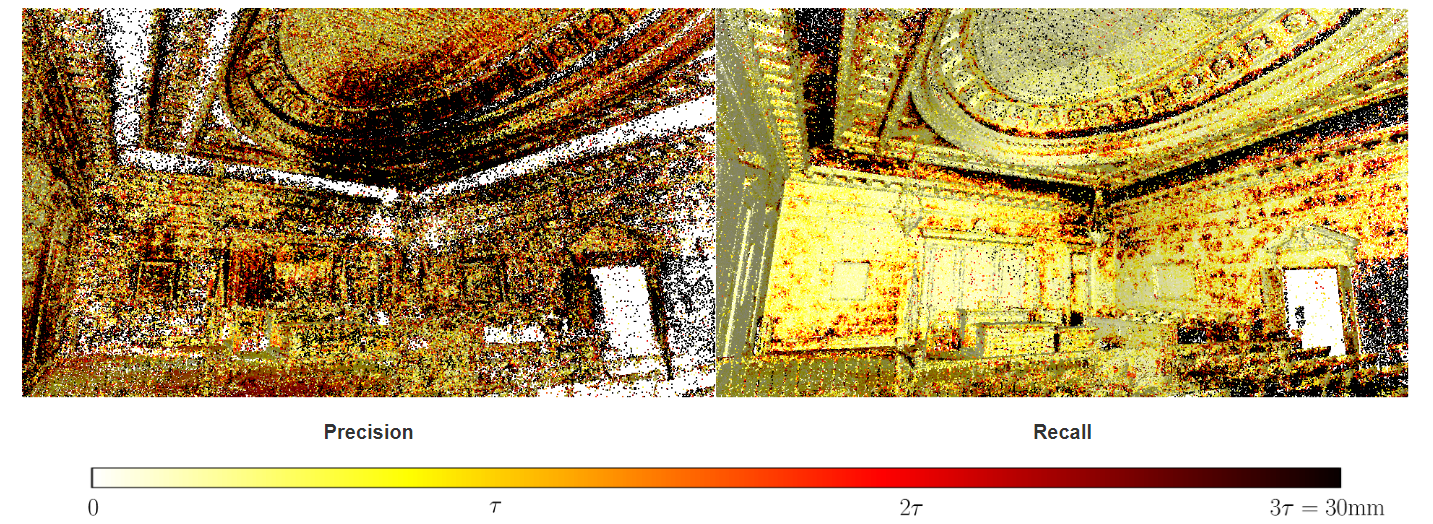}
     
     \includegraphics[width=0.28\linewidth]{figs/museum00.png}
     \includegraphics[width=0.46\linewidth]{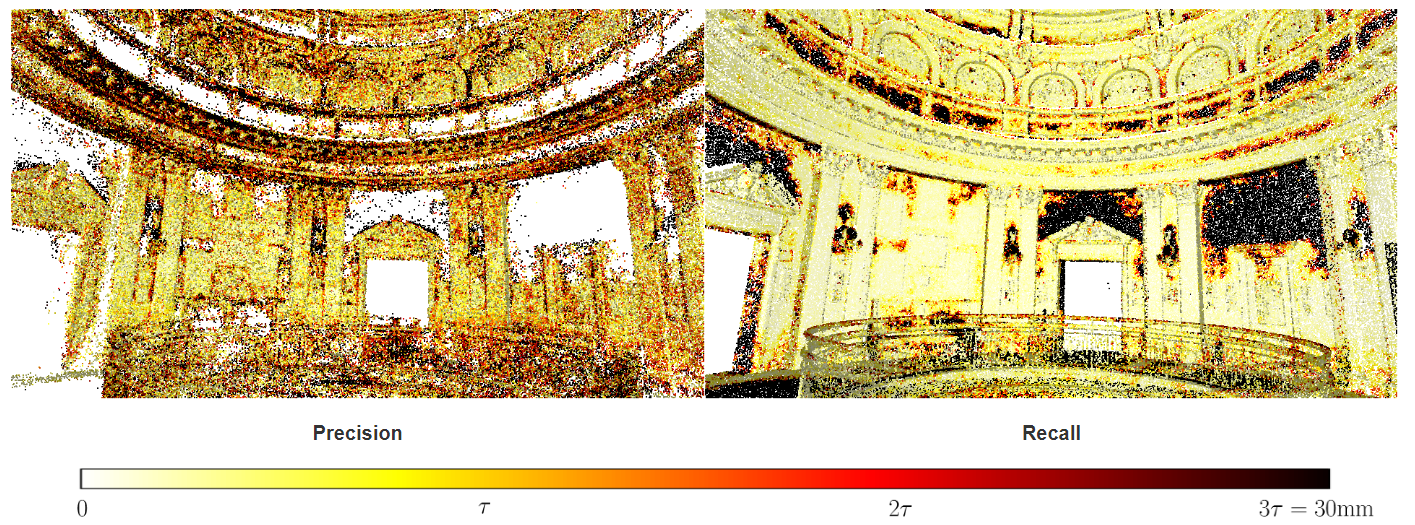}
     
     \includegraphics[width=0.28\linewidth]{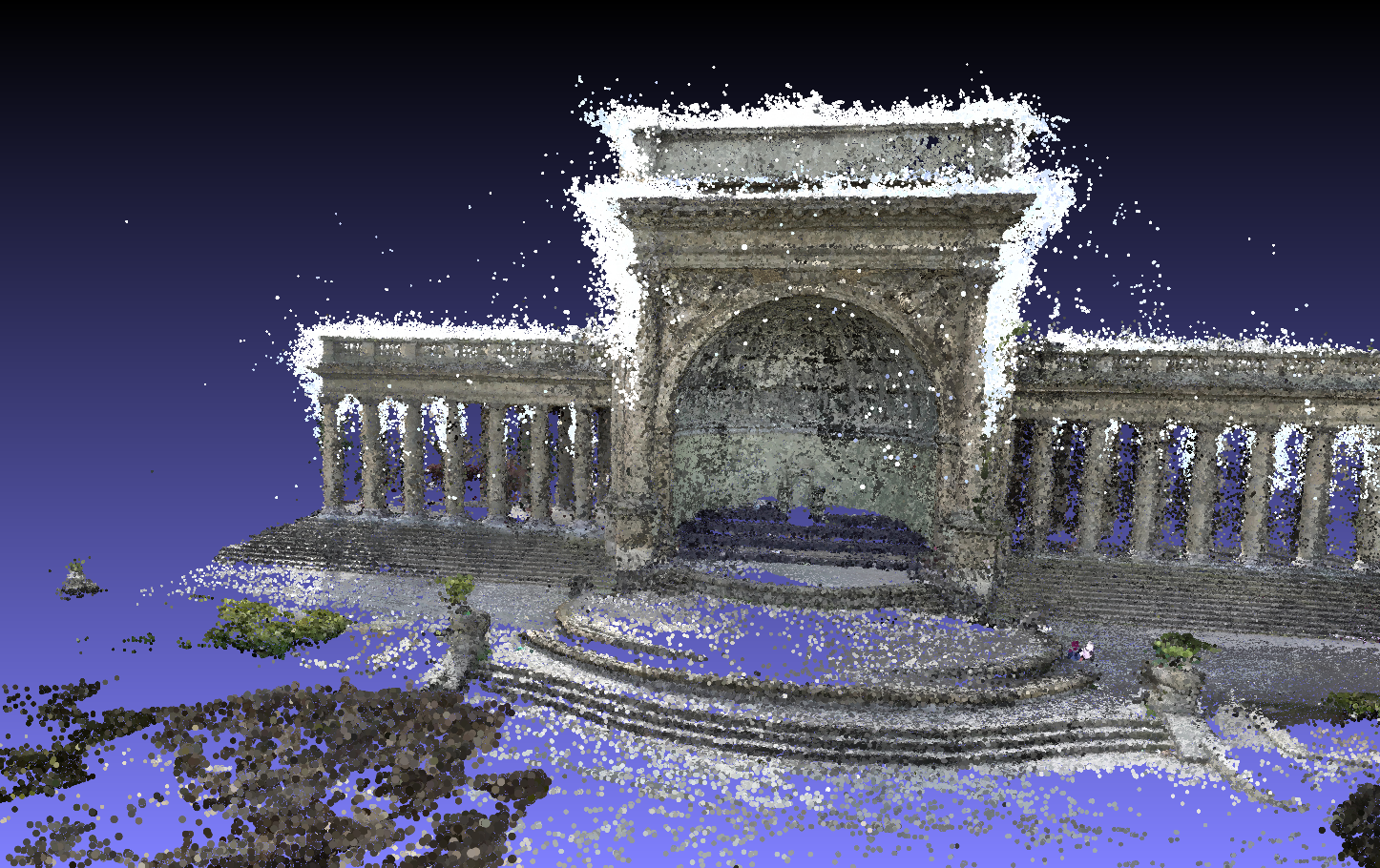}
     \includegraphics[width=0.46\linewidth]{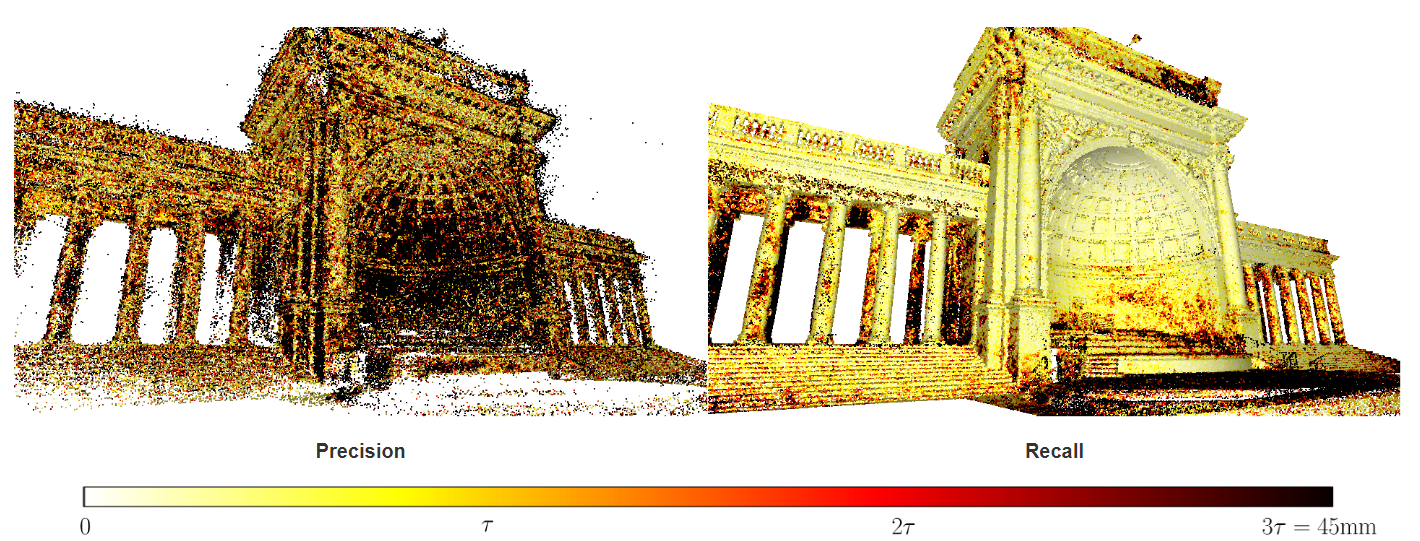}
     \caption{{\bf More qualitative results on Tanks and Temples Advanced. } }
     \label{fig:TA}
     
  \end{center}
\end{figure*}

\end{document}